\documentclass[
11pt, 
oneside, 
english, 
singlespacing, 
headsepline, 
]{MastersDoctoralThesis} 

\usepackage[utf8]{inputenc} 
\usepackage[T1]{fontenc} 

\usepackage{palatino} 

\usepackage[style=numeric,natbib=true]{biblatex} 

\usepackage[autostyle=true]{csquotes} 

\usepackage{times}
\usepackage{latexsym}
\usepackage{multirow}
\usepackage{graphicx}
\usepackage{amsmath,amsthm}
\usepackage{amsfonts}
\usepackage{bm}
\usepackage{xcolor, soul}

\newcommand\blue[1]{\textcolor{blue}{#1}}
\newcommand\orange[1]{\textcolor{orange}{#1}}
\newcommand\green[1]{\textcolor{green}{#1}}
\newcommand\red[1]{\textcolor{red}{#1}}

\hypersetup{%
  colorlinks = true,
  linkcolor  = black
}

\thesistitle{Learning Text Styles: A Study on Transfer, Attribution, and Verification} 
\supervisor{Dr. Roy Ka-Wei \textsc{Lee}} 
\examiner{} 
\degree{Doctor of Philosophy} 
\author{Zhiqiang \textsc{Hu}} 
\addresses{} 
\university	{SUTD}
\pillar{ISTD} 

\hypersetup{pdftitle=\ttitle} 
\hypersetup{pdfauthor=\authorname} 
\hypersetup{pdfkeywords=\keywordnames} 

\begin{document}

\frontmatter 

\pagestyle{plain} 


\begin{titlepage}
\begin{center}

\begin{figure}
\centering
\includegraphics[width=0.5\textwidth]{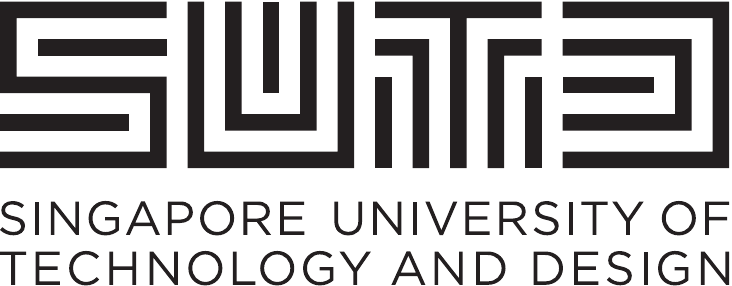}
\end{figure}

 \hfill\break\\[2.0cm]

{\huge \bfseries \ttitle}\\[3cm] 

Submitted by\\[1cm]
\authorname 

\vspace{4em}

Thesis Advisor\\[1cm]
\supname 

\vspace{4em}

\pillarname\\[1.5cm] 

\large{A thesis submitted to the Singapore University of Technology and Design in fulfillment of the requirement for the degree of \degreename}\\[1cm] 

{\large \the\year}\\[4cm] 

\vfill
\end{center}
\end{titlepage}


\begin{tec}
\addchaptertocentry{\tecname}
\begin{tabular}{ll}
	TEC Chair: & Prof. Soujanya Poria  \\\\
	Main Advisor: & Prof. Roy Ka-Wei Lee \\\\
	Internal TEC member 1: & Prof. Peng Song \\\\
	Internal TEC member 2: & Prof. Wei Lu \\\\
        \bottomrule 
\end{tabular}
\end{tec}
\vfill\eject

\begin{abstract}
\addchaptertocentry{\abstractname} 

This thesis advances the computational understanding and manipulation of text styles through three interconnected pillars: (1) Text Style Transfer (TST),  which alters stylistic properties (e.g., sentiment, formality) while preserving content; (2)Authorship Attribution (AA), identifying the author of a text via stylistic fingerprints; and (3) Authorship Verification (AV), determining whether two texts share the same authorship. We address critical challenges in these areas by leveraging parameter-efficient adaptation of large language models (LLMs), contrastive disentanglement of stylistic features, and instruction-based fine-tuning for explainable verification.

First, for TST, we conduct a comprehensive survey and reproducibility study of 19 state-of-the-art algorithms, establishing benchmarks across diverse datasets. Building on these insights, we introduce LLM-Adapters, a unified framework for parameter-efficient fine-tuning (PEFT) that enables cost-effective adaptation of LLMs for style-centric tasks. This culminates in Adapter-TST, a novel architecture that models multiple stylistic attributes (e.g., sentiment, tense) using lightweight neural adapters. Adapter-TST achieves superior performance in multi-attribute transfer and compositional editing while reducing computational costs by 80\% compared to full fine-tuning.

For AA, we propose ContrastDistAA, a contrastive learning framework that disentangles content and style features to address performance degradation under topic shifts. Our method advances both individual-level attribution and regional linguistic analysis, achieving state-of-the-art accuracy by isolating culturally influenced stylistic patterns.

In AV, we bridge the gap between LLM capabilities and forensic analysis with InstructAV, an instruction-tuned model that jointly optimizes verification accuracy and explainability. By aligning classification decisions with human-interpretable rationales, InstructAV outperforms ChatGPT and specialized baselines across multiple datasets, demonstrating the viability of LLMs for forensically rigorous tasks.

Collectively, this work establishes a paradigm for efficient, interpretable, and multi-faceted text style analysis. Our contributions span novel architectures (Adapter-TST, ContrastDistAA, InstructAV), reproducible benchmarks, and open-source tools (LLM-Adapters), enabling future research in style-aware NLP applications ranging from personalized text generation to forensic linguistics.

\end{abstract}


\begin{publications}
\addchaptertocentry{\publicationsname} 

\begin{itemize}
    \item \textbf{Text Style Transfer: A Review and Experimental Evaluation} \\ \textbf{Zhiqiang Hu}, Roy Ka-Wei Lee, Charu C Aggarwal, Aston Zhang.\\
    ACM SIGKDD Explorations Newsletter 2022

    \item \textbf{LLM-Adapters: An Adapter Family for Parameter-Efficient Fine-Tuning of Large Language Models.}\\
    \textbf{Zhiqiang Hu}, Lei Wang, Yihuai Lan, Wanyu Xu, Ee-Peng Lim, Lidong Bing, Xing Xu, Soujanya Poria, Roy Ka-Wei Lee.\\
    EMNLP 2023

    \item \textbf{Adapter-TST: A Parameter Efficient Method for Multiple-Attribute Text Style Transfer}\\
    \textbf{Zhiqiang Hu}, Roy Ka-Wei Lee, Nancy F Chen.\\
    EMNLP Findings 2023

    \item \textbf{Contrastive Disentanglement for Authorship Attribution.}\\
    \textbf{Zhiqiang Hu}, Thao Thanh Nguyen, Yujia Hu, Chia-Yu Hung, Ming Shan Hee, Chun Wei Seah, Roy Ka-Wei Lee.\\
    Companion Proceedings of the ACM on Web Conference 2024

    \item \textbf{InstructAV: Instruction Fine-tuning Large Language Models for Authorship Verification.}\\
    Yujia Hu, \textbf{Zhiqiang Hu*}, Chun Wei Seah, Roy Ka-Wei Lee\\
    COLM 2024
\end{itemize}

\end{publications}


\begin{acknowledgements}
\addchaptertocentry{\acknowledgementname} 
The four years I spent at the SUTD SocialAI Studio have been an unforgettable and precious experience. It was my first time living and studying outside of China for such a long period. Every day in Singapore was fulfilling and wonderful. To me, it was an invaluable opportunity to complete my PhD studies in Singapore, and of course, this would not have been possible without the help and support of many people.

First of all, I would like to thank my advisor, Prof. Roy Ka-Wei Lee. We met during my master’s studies in Chengdu—truly a case of "meeting late but to great effect." Later, we began doing research and publishing papers together, and Roy generously gave me my first opportunity to go abroad. With his support, I went to Canada for an internship, spending a year in a place colder than the Arctic. During that time, I was stranded in Canada due to the pandemic, and if it weren’t for James’s help, I might not have made it back. Many thanks to James, my buddy! With Roy’s continued support, I was able to apply for and join the PhD program at SUTD, beginning my four-year PhD journey. I’m truly grateful to Roy for his guidance and support throughout these four years. He provided not only constructive academic advice but also essential resources. This support played a vital role in my ability to complete my PhD within four years.

I would also like to thank my TEC members: Professors Song Peng, Lu Wei, and Soujanya Poria, for their insightful and valuable feedback during my Qualifying Exam and Preliminary Exam.

During my PhD studies, I was very fortunate to collaborate with many outstanding researchers. First, I would like to thank my most important collaborator, my senior Dr. Lei Wang, for guiding me academically during both my master’s and PhD stages, working with me on research projects, and together producing some impactful work. I also thank my collaborators in the text style-related projects: Dr. Charu C Aggarwal, Dr. Aston Zhang, Dr. Nancy F Chen, Chia-Yu Hung, Yujia Hu, and Thao Thanh Nguyen. For the LLM-related projects, I thank Yihuai Lan, Wanyu Xu, Wenhao Shi, and Dr. Yi Bin.

I’m grateful to all the members of SUTD SocialAI Studio — Mingshan Hee, Yuhao Wu, Daniel Chin, Nirmalendu Prakash, Shaun Toh, and Siqi Xiang — for their help in both life and study throughout my PhD. Many thanks to my colleagues at Alibaba DAMO NLP \& MultiModal team, and especially to Dr. Lidong Bing and Dr. Xin Li for their careful guidance, from which I benefited greatly. I also thank Dr. Hang Zhang, Dr. Wenxuan Zhang, Dr. Xuan Phi Nguyen, Guanzheng Chen, Sicong Leng, Zesen Cheng, Boqiang Zhang, Wenqi Zhang, Jiayan Guo, Ronghao Dang, Yuqian Yuan, Yuming Jiang, Jiashuo Sun, Kehan Li, Weiwen Xu, Yongxin Zhu, and Long Li. It has been my honor to collaborate with such outstanding researchers — thank you all for your guidance and support.

Most of all, I want to thank my parents for their unwavering support and encouragement throughout my academic journey. Their love is always selfless and unconditional. I hope they stay healthy and happy always.

Finally, I want to express my heartfelt gratitude to my girlfriend, Zongqian Li, for her endless love and support over the past four years. From a long-distance relationship within the same country to a cross-border one, and even through the COVID-19 pandemic, we have always firmly chosen each other. Thank you for your dedication and perseverance all along the way.

\end{acknowledgements}


\tableofcontents 

\listoffigures 

\listoftables 

\mainmatter 

\pagestyle{thesis} 


\chapter{Introduction} 

\label{introduction} 


\newcommand{\keyword}[1]{\textbf{#1}}
\newcommand{\tabhead}[1]{\textbf{#1}}
\newcommand{\code}[1]{\texttt{#1}}
\newcommand{\file}[1]{\texttt{\bfseries#1}}
\newcommand{\option}[1]{\texttt{\itshape#1}}


\section{Background}
The study of text style is not merely an academic curiosity; it addresses fundamental challenges and opens a plethora of opportunities in computational linguistics and natural language processing (NLP) \cite{fu2018style,hu2022text,jin-etal-2022-deep,zheng2023review,neal2017surveying,boenninghoff2019explainable,huang2024largelanguagemodelsidentify}. The stylistic properties of text influence our perception, interpretation, and interaction with written communication. Learning text style, therefore, has significant implications for various applications, from improving human-computer interaction to enhancing digital forensics.

Text style is an essential component of written communication that encapsulates the author’s unique voice, emotional undertone, and cultural background \cite{hu2022text,jin-etal-2022-deep}. The ability to understand and manipulate text style allows systems to generate content that resonates more deeply with human users. For instance, altering the sentiment of a customer service response to be more empathetic can drastically improve customer satisfaction \cite{hu2022current}. Similarly, adjusting the stylistic tone of educational content to be more engaging can enhance learner engagement and comprehension.

Text Style Transfer (TST) represents one of the primary applications wherein learning text style plays a pivotal role \cite{fu2018style,hu2022text,jin-etal-2022-deep}. TST algorithms enable the transformation of text from one style to another while preserving the original content's meaning. This functionality is particularly beneficial in areas such as:

\begin{enumerate}
    \item \textbf{Personalization}: Tailoring content to fit individual user preferences, such as adjusting formality in emails or modifying humor in social media posts.
    \item \textbf{Content Moderation}: Automatically converting offensive or inflammatory language into polite or neutral expressions can create safer digital environments.
    \item \textbf{Creative Writing}: Assisting authors and content creators by mimicking particular styles or generating stylistically diverse drafts to stimulate creative processes.
\end{enumerate}

Through these applications, TST technologies have found increasing relevance in both consumer-facing technologies and professional settings, leading to widespread adoption and continued research interest.

Authorship Attribution (AA) and Authorship Verification (AV) are critical areas where learning text style proves invaluable \cite{zheng2023review,neal2017surveying,boenninghoff2019explainable,huang2024largelanguagemodelsidentify}. AA techniques identify the author of a piece of text by analyzing linguistic and stylistic features, a capability crucial for:

\begin{enumerate}
    \item \textbf{Plagiarism Detection}: determining whether a text has been copied without proper attribution.
    \item \textbf{Digital Forensics}: identifying the authors of malicious texts such as anonymous threats, fake news, or propaganda.
    \item \textbf{Historical Analysis}: attributing historical texts to likely authors, thereby contributing to literary scholarship and the preservation of cultural heritage.
\end{enumerate}

AV extends these capabilities by verifying whether two texts share the same authorship, a task pivotal in legal investigations and academic integrity checks \cite{huang2024largelanguagemodelsidentify,promptav,wang2023planandsolvepromptingimprovingzeroshot}. The ability to distinguish between individual authorial styles through sophisticated models transforms these tasks from labor-intensive manual processes to automated, scalable solutions. 

Despite the remarkable advancements enabled by learning text style, significant challenges persist:

\begin{enumerate}
    \item \textbf{Semantic Preservation}: Ensuring that content remains unchanged while its style is modified is a non-trivial task that requires sophisticated models capable of disentangling content from style.
    \item \textbf{Resource Efficiency}: Fine-turning large language models to handle stylistic variations demands extensive computational resources and annotated data. Methods such as Parameter-Efficient Fine-Tuning (PEFT) offer promising solutions by minimizing resource requirements while maintaining high performance.
    \item \textbf{Explainability}: For both AA and AV, the interpretability of the model’s decisions is crucial. Transparent methodologies like \textsf{InstructAV} that align classification decisions with understandable explanations address the need for accountability and trust in automated systems.
\end{enumerate}

Furthermore, as social nuances and cultural contexts continue to evolve, models that handle text style must adapt dynamically to new stylistic trends and linguistic innovations. This evolving landscape presents opportunities for continual research and development to refine the precision and applicability of text-style technologies.

Paying attention to learning text style is not solely about advancing NLP methodologies but also about enhancing the way we interact with technology and each other through written language. The integration of stylistic understanding in computational systems leads to more natural, engaging, and effective communication. This field holds the promise of transforming numerous industries, from digital content creation to cybersecurity, making it a cornerstone of future developments in computational linguistics and artificial intelligence.

\section{Motivation}
The investigation into text style across the domains of TST, AA, and AV is driven by the pressing need to enhance the capabilities of computational systems to understand, generate, and authenticate text with human-like proficiency \cite{hu2022text,jin-etal-2022-deep,zheng2023review,neal2017surveying}. This motivation is rooted in the significant societal, technological, and academic challenges that these fields address, each contributing uniquely to the broader landscape of NLP.

In an increasingly digital world where personalized communication is paramount, TST stands as a central area of interest due to its transformative potential. The motivation behind studying TST is its application in creating more adaptable and human-centric communication systems. By enabling the dynamic alteration of text styles while preserving meaning, TST facilitates a range of applications, from tailoring content for different audiences to enhancing user engagement in digital interfaces. The need for a comprehensive study is underscored by the growing number of TST algorithms, leading to extensive survey and experimental evaluations that establish foundational benchmarks and taxonomies.

The explosion of LLMs like ChatGPT \cite{qin-chatgpt-2023} and GPT-4 \cite{openai-gpt4-2023} has offered unprecedented opportunities while also imposing significant computational demands. The proposed LLM-Adapters and Adapter-TST address this challenge through the development of parameter-efficient fine-tuning techniques. By leveraging adapter-based methods, these studies provide solutions that make the deployment of sophisticated NLP models more accessible and viable. Their motivation lies in reducing resource consumption while maintaining high performance, thus democratizing advanced NLP capabilities and supporting diverse applications, from simple text classification to complex TST tasks as elaborated in the Adapter-TST framework.

In the realm of AA, understanding text style serves a critical function in identifying authorship through stylistic elements. The motivation for \textsf{ContrastDistAA} stems from the limitations of existing methods in disentangling style from topical content, which is crucial in datasets exhibiting significant topic variations. Addressing this gap, the introduction of \textsf{ContrastDistAA} employs contrastive learning and mutual information to improve accuracy and extend AA’s applicability to regional linguistic analysis, enhancing both forensic linguistics and cultural studies.

AV extends these capabilities by providing tools to verify text authorship, a task essential in educational, forensic, and legal contexts. The motivation for developing \textsf{InstructAV} arises from the necessity to enhance accuracy and interpretability in AV tasks. By aligning the decision-making process of LLMs with transparent explanations, this research ensures that progress in AV respects the demands for accountability and trust, thereby strengthening digital security measures.

Across all the proposed approached, the overarching motivation is to confront the prevalent challenges in the current landscape of computational linguistics. The necessity for semantic preservation and resource efficiency forms a common thread, emphasizing the need for techniques that maintain content integrity while optimizing computational resources. Furthermore, as language and stylistic conventions evolve, these studies collectively advocate for adaptive models capable of accommodating new linguistic paradigms.

\section{Overview of Dissertation Works}
In the subsequent section, we present an overview of the dissertation, dividing it into three primary parts, each encompassing a detailed description of the key research concepts.
\subsection{Parameter-efficient Text Style Transfer}
We conduct a comprehensive survey and  reproducibility study of 19
state-of-the-art approaches. TST has garnered significant attention due to its potential applications in content personalization, sentiment adaptation, and more. However, adapting large language models for multi-attribute text style transfer presents a formidable challenge, primarily due to the extensive computational resources and labeled data required for fine-tuning such models on specific downstream tasks. This constraint underscores the necessity for parameter-efficient fine-tuning methods, which can achieve desired performance with significantly less computational cost. We first introduce LLM-Adapters, a unified framework for PEFT that enables cost-effective adaptation of LLMs for style-centric tasks. Subsequently, we propose to use adapter-based PEFT method as a viable solution to perform multi-attribute TST. In our proposed method, \textsf{Adapter-TST}, we freeze the pre-trained model's original parameters and incorporate neural adapters to capture various stylistic attributes like sentiment, tense, and voice. By using BART or T5 as the backbone model, our approach efficiently manages multiple attributes through the insertion of modular adapters, enabling effective compositional editing. Our experiments demonstrate that \textsf{Adapter-TST} not only outperforms current state-of-the-art baselines in traditional and multiple-attribute transfer tasks, but it also achieves these results with substantially less computational effort.

\subsection{Contrastive Disentaglement for Authorship Attribution}
Authorship attribution aims to discern the authorship of a text by analyzing distinctive writing styles. Despite recent advancements, existing AA methods struggle in scenarios characterized by significant topic shifts, primarily due to their inability to disentangle topical content from stylistic elements inherent in an author's writing. Moreover, the field has largely focused on individual-level AA, neglecting the exploration of regional-level AA that could unveil linguistic patterns influenced by cultural and geographical factors. This dissertation seeks to address these limitations by introducing \textsf{ContrastDistAA}, a framework leveraging contrastive learning and mutual information maximization techniques to clearly disentangle content and stylistic features within latent representations used for AA. Our comprehensive experiments demonstrate that \textsf{ContrastDistAA} not only outperforms current state-of-the-art models but also broadens the applicability of AA to include regional linguistic analysis, thus enriching its contributions to computational linguistics.

\subsection{Instruction Fine-tuning LLM for Authorship Verification}

Authorship verification tasks, requiring the determination of whether two texts share the same authorship, represent a particular challenge even for advanced large language models like ChatGPT. The primary motivators for enhancing AV performance lie in both improving accuracy and increasing the transparency and interpretability of the models' decisions. Existing LLMs often fall short in these areas, necessitating novel approaches. We introduce \textsf{InstructAV}, a method that marries LLM functionality with PEFT to uplift both accuracy and explainability in AV tasks. The crux of our method lies in aligning AV classification decisions with transparent, understandable explanations, marking a significant advancement in the field. Through rigorous experimentation across diverse datasets, \textsf{InstructAV} consistently demonstrates state-of-the-art performance, achieving high classification accuracy while concurrently enhancing the reliability of explanatory outputs.

\section{Dissertation Structure}

We first discuss related works that are connected to the topics of this thesis in
Chapter \ref{related_work}. The rest of this thesis consists of three parts — Part I: Parameter-efficient Text Style Transfer, Part II: Contrastive Disentaglement for Authorship Attribution, and Part III: Instruction Fine-tuning LLM for Authorship Verification.

Part I contains Chapter \ref{tst_overview}, \ref{llm_adapters}, and \ref{adapter_tst}, focusing on parameter-efficient text style transfer. In Chapter \ref{tst_overview}, we delve into a comprehensive survey and experimental evaluation of TST methodologies. This chapter provides a structured taxonomy of existing TST models and reviews the current evaluation methodologies. By conducting a large-scale reproducibility study, it benchmarks 19 state-of-the-art TST algorithms on publicly available datasets. The findings of this chapter set the stage for understanding the strengths and limitations of existing TST approaches, highlighting areas for further exploration.

Chapter \ref{llm_adapters} serves as a bridging work for the parameter-efficient fine-tuning methods discussed in this part. This chapter introduces the LLM-Adapters framework, which integrates various adapter-based PEFT methods into LLMs. The chapter discusses the framework's impact on task-specific performance while requiring only minimal additional parameters. It evaluates the effectiveness of different adapter configurations across multiple reasoning tasks and paves the way for its application in text style transfer.

In Chapter \ref{adapter_tst}, we propose the \textsf{Adapter-TST}, a novel framework deploying parameter-efficient fine-tuning for multi-attribute text style transfer. Building upon the foundational insights from the previous chapters, this chapter explores how adapter-based methods can control multiple stylistic attributes simultaneously. Empirical evaluations demonstrate the efficiency and superior performance of \textsf{Adapter-TST} compared to state-of-the-art baselines, showcasing its capacity for effective compositional editing with limited computational resources.

Part II contains Chapter \ref{contrastiveaa}, where we introduce \textsf{ContrastDistAA}, an innovative framework for enhancing authorship attribution. This chapter addresses the challenges of disentangling stylistic elements from topical content, especially in scenarios with significant topic shifts. Utilizing contrastive learning and mutual information maximization, the proposed method advances both individual and regional-level AA tasks. The chapter elaborates on the substantial contributions to computational linguistics by extending the applicability of AA to incorporate regional linguistic analysis.

Part III contains Chapter \ref{instractav}, focusing on the novel approach of \textsf{InstructAV} for authorship verification. It explores how instruction fine-tuning in LLMs can elevate the accuracy and explainability of AV tasks. By aligning classification decisions with clear explanations, \textsf{InstructAV} provides significant advancements in AV methodologies. Comprehensive experiments underscore the framework's state-of-the-art performance, highlighting enhanced explanatory reliability alongside high classification accuracy.

Finally, in Chapter \ref{conclusion}, we summarize the thesis’s contributions and suggest potential future directions.

\section{Contributions}

The technical contributions of this thesis are summarized as follows:

\begin{itemize}
    \item We have developed a user-friendly framework, LLM-Adapters, which seamlessly integrates diverse adapters into LLMs. This empowers researchers to implement adapter-based PEFT methods across a wide range of tasks. Furthermore, we have conducted a comprehensive empirical study examining various PEFT methods within different open-source LLMs.
    
    \item We present \textsf{Adapter-TST}, a parameter-efficient framework capable of executing multiple-attribute text style transfer tasks while requiring significantly reduced computational resources. The \textsf{Adapter-TST} comprises two TST configurations: \textit{parallel}, which facilitates multiple-output TST, and \textit{stacking}, which enables compositional editing. Extensive experimentation on real-world datasets demonstrates that \textsf{Adapter-TST} surpasses state-of-the-art baselines in transferring and generating high-quality text, as indicated by both automatic and human evaluations.
    
    \item Our proposed \textsf{ContrastDistAA} framework signifies a methodological advancement in authorship attribution, utilizing contrastive learning and mutual information maximization to effectively distinguish between content and style in textual data. This approach achieves unparalleled accuracy in AA tasks, especially in contexts affected by topic shifts. Through extensive experimentation, we have benchmarked \textsf{ContrastDistAA} against current state-of-the-art AA methods across various scenarios. The results clearly exhibit \textsf{ContrastDistAA}'s superior performance in both individual and regional-level AA tasks, highlighting its effectiveness and adaptability.
    
    \item We introduce the \textsf{InstructAV} framework for authorship verification  tasks, which accurately determines whether two texts are authored by the same individual and provides substantial linguistic explanations for the AV outcomes. We have curated three instruction-tuning datasets, each supplemented with reliable linguistic explanations pertinent to AV tasks. Evaluation results from both automated methods and human reviewers confirm the efficacy of \textsf{InstructAV} in delivering precise AV predictions and dependable linguistic explanations.
\end{itemize}
\chapter{Related Work} 

\label{related_work} 

\section{Text Style Transfer}

TST is an emerging research topic that has garnered attention from computational linguists and computer science researchers. The recent comprehensive survey \cite{hu2022text,jin-etal-2022-deep} summarizes the existing TST approaches. 

While the majority of existing studies have focused on performing TST on single attributes such as sentiment~\cite{li2018delete,Luo19DualRL,fu2018style} or formality~\cite{rao2018dear}, recent studies have also explored multiple-attribute TST tasks, where TST models are designed to control and transfer text in multiple target stylistic attributes. \citet{lampleSSDRB19} attempts style transfer with multiple attributes by conditioning on the average embedding of each target attribute and using a combination of denoising autoencoder (DAE) and back-translation techniques. \citet{goyal-etal-2021-multi} propose an approach to initialize an encoder-decoder setup with a transformer-based language model that is pre-trained on a generic corpus and enhances its capability of re-writing to multiple target style dimensions by utilizing multiple style-aware language models as discriminators. In this study, we contribute to this limited multiple-attribute TST literature by proposing an alternative approach to generate multiple stylistic outputs and perform compositional editing efficiently. 

Due to the lack of parallel training data, most existing TST methods are designed to train with non-parallel style-labeled sentences as input. A popular line of TST approaches aims to disentangle the text's content and style in the latent space to perform TST~\cite{shen2017style,zhao2018adversarially,fu2018style,chen2018adversarial,logeswaran2018content,yin2019utilizing,lai2019multiple,john2019disentangled}. Another common approach is to leverage PLMs. For instance, \citet{syed2020adapting} fine-tune a denoising autoencoder (DAE) for the stylized re-writing task by initializing the encoder and decoder with a pre-trained language model trained on Masked Language Modeling (MLM) objectives \cite{devlin-etal-2019-bert}. \citet{wang2019harnessing} fine-tune GPT-2 model \cite{radford2019language} using the text formality transfer rules harnessed from analyzing the GYAFC parallel dataset \cite{rao2018dear}. The fine-tuned GPT-2 model was subsequently used to transfer the formality of text (e.g., informal to formal text). However, fine-tuning PLMs for multiple-attribute TST remains challenging as a significant amount of computational resources is required to perform the task; multiple PLMs need to be fine-tuned for the different attributes to perform multiple-attribute TST. In this study, we overcome this limitation by proposing \textsf{Adapter-TST}, which is a parameter-efficient framework that leverages on PLMs but requires significantly lesser computational resources to perform multiple-attribute TST.

\section{Parameter-Effecient Fine-Tuning}

\begin{figure}[t] 
	\centering
	\includegraphics[scale = 0.12]{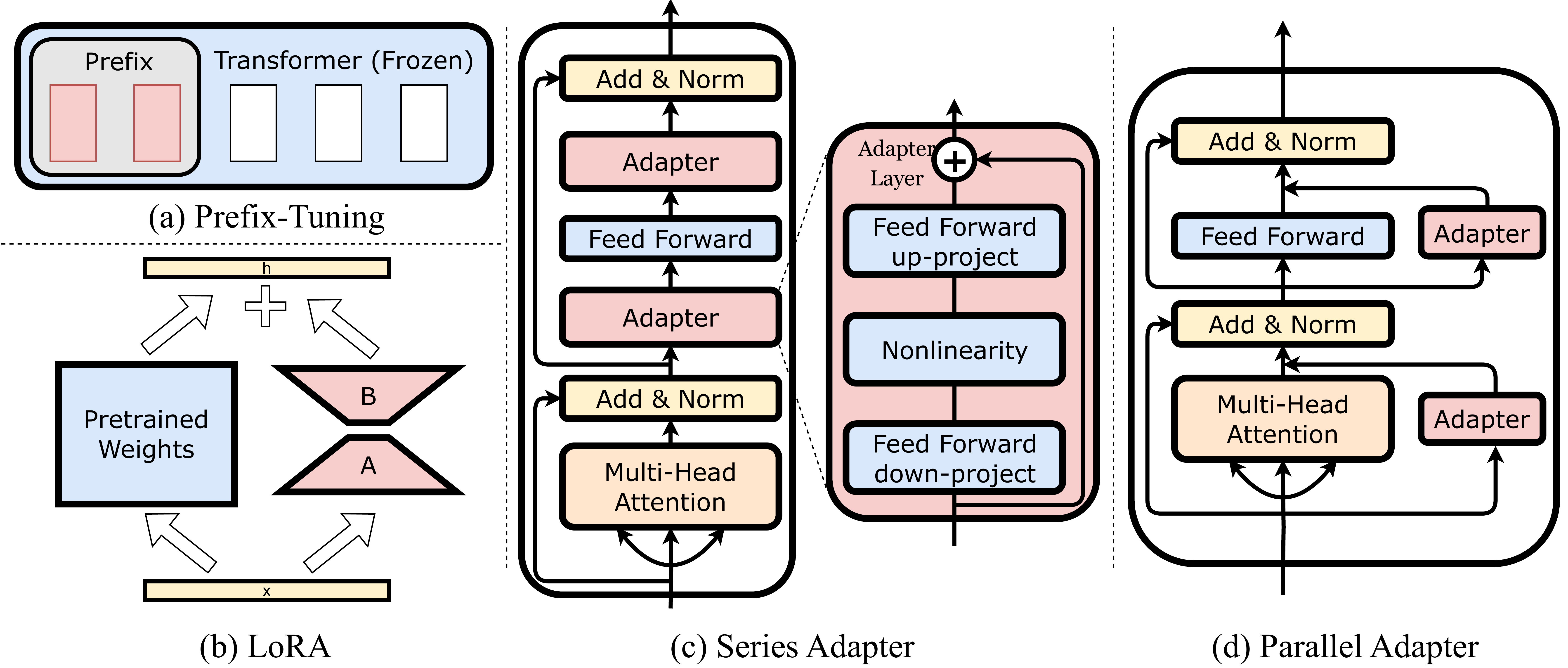} 
	\caption{A detailed illustration of the model architectures of three different adapters: (a) Prefix-Tuning, (b) LoRA, (c) Series Adapter, and (d) Parallel Adapter.}
        \label{fig:peft_overview}
\end{figure}

\begin{table}[t]
\centering
\caption{The PEFT methods are categorized based on the four common basic methods. "Prompt" represents prompt-based learning methods, "Repara" denotes reparametrization-based methods, "Series" is Series Adapter, while "Parallel" represents Parallel Adapter.
}
\begin{tabular}{l|cccc}
\hline
Method & Prompt & Repara & Series & Parallel  \\\hline
Prompt Tuning \cite{prompt_tuning}  &$\surd$ & & & \\
Prefix-Tuning  \cite{prefix}    & $\surd$& & & \\
Spot   \cite{spot}            & $\surd$& & & \\
IPT    \cite{ipt}        &$\surd$ & & & \\
LoRA     \cite{lora}       & &$\surd$ & & \\
KronA   \cite{krona}       & &$\surd$ & & \\
Adapters \cite{adapters}  & & &$\surd$ &        \\
AdaMix \cite{adamix}  & & &$\surd$ & \\
SparseAdapter \cite{sparseadapter}       & & &$\surd$ & \\
LeTS  \cite{lets}    & & & $\surd$ & \\
Parallel Adapter \cite{parallel_adapter} & & & & $\surd$ \\
MAM Adapter  \cite{mam}    &$\surd$ &$\surd$ &$\surd$ & \\
UniPELT   \cite{unipelt}      &$\surd$ & $\surd$ &$\surd$ & \\
Compacter  \cite{compacter}     & &$\surd$ &$\surd$ & \\
S4-model  \cite{s4_model}     &$\surd$ &$\surd$ & & \\
\hline
\end{tabular}

\label{tab:peft_category}
\end{table}

In this section, we provide a brief overview of four PEFT methods: prompt-based learning, reparametrization-based methods, series adapters, and parallel adapters. \cite{prefix,lora,adapters,parallel_adapter}

\paragraph{Prompt-based learning.} As shown in Figure~\ref{fig:peft_overview}(a), prompt-based learning transforms the discrete optimization problem of finding the optimal hard prompt into a continuous (soft) prompt. To achieve this, \citet{prompt_tuning} proposed the concept of prompt tuning, where a trainable tensor is added as a prefix to the input embeddings. Another approach called Prefix-Tuning\cite{prefix} independently explored the addition of soft prompts to the hidden states of all layers. Intrinsic Prompt Tuning~\cite{ipt} employs an autoencoder to compress and decompress the soft prompt.
We take learnable vectors incorporated into the attention layer as an example of prompt-based learning, which can be formulated as follows:
\begin{align}
 H_{o}^{} = \text{Attn}(H_{i}W_{Q},[P_{K};H_{i}W_{K}], [P_{V};H_{i}W_{V}]),
\end{align}
where $H_i\in\mathrm{R}^{T\times d}$ and $H_o\in\mathrm{R}^{T\times d}$ are the input and output of the attention layer respectively. Note that $T$ is the maximum input length and $d$ is the vector dimension. $P_{K}\in\mathrm{R}^{L\times d}$ and $P_{V}\in\mathrm{R}^{L\times d}$ are the learnable vectors for PEFT. 
$L$ is the number of learnable tokens, which is discussed in the experiment section in detail.
$Q, K, V$ denote the query, key, value vectors of th attention module, respectively.

\paragraph{Reparametrization-based method.} This type of methods aim to transform network weights using a low-rank technique. This approach effectively reduces the number of trainable parameters while preserving the ability to handle high-dimensional matrices. Intrinsic SAID~\cite{intrinsic_said} investigates the intrinsic dimensionality of fine-tuning within a low-rank subspace. LoRA~\cite{lora} introduces a simple approach to update the parameters of a weight matrix by decomposing it into a product of two low-rank matrices. KronA~\cite{krona} improves upon the matrix factorization aspect of LoRA by utilizing the Kronecker product in its technique.
We take LoRA as an example of  Reparametrization-based learning, which can be formulated below:
\begin{align}
H_{o}=H_{i}W_{0}+H_{i}\Delta W=H_{i}W_{0}+ H_{i}BA,
\end{align}
where $W_0\in\mathrm{R}^{d\times d}$ can be any pre-trained weight matrix, including weights in the MLP or Attention layer. $B\in\mathrm{R}^{r\times d}$ and $A\in\mathrm{R}^{r\times d}$ are lower-rank matrix intended for covering $\Delta W$. $r \ll d$ is an important hyper-parameter for LoRA.

\paragraph{Series Adapter.} Series adapters involve incorporating additional learnable modules in a sequential manner within a specific sublayer. In their study, \citet{adapters} proposed integrating fully-connected networks after the attention and FFN layers in the Transformer model~\cite{vaswani2017attention}. Another finding by \citet{madx} revealed that achieving comparable performance is possible by inserting the adapter solely after the self-attention layer, instead of using two adapters per transformer block. AdaMix~\citep{adamix} introduces a method that utilizes multiple series adapters in a mixture-of-experts (MoE) fashion. Compacter~\citep{compacter} utilizes the Kronecker product, low-rank matrices, and parameter sharing across layers to generate adapter weights. This technique aims to reduce the computational complexity associated with the adapters while maintaining their performance. 
Series Adapter can be formulated as follows:
\begin{equation}
     H_{o} \gets  H_{o}+f( H_{o}W_{down})W_{up},
\end{equation}
where the output $H_{o}$ of a specific layer, such as the MLP layer, is first down-projected by $W_{down}\in\mathrm{R}^{d\times r}$ to a lower dimension $r$, and then up-projected back by $W_{up}\in\mathrm{R}^{r\times d}$ to the original dimension $d$. $f$ is a non-linear function. We discuss the choice of $r$ in the experiment Section.

\paragraph{Parallel Adapter.} Parallel adapters \cite{parallel_adapter}  aim to incorporate additional learnable modules in parallel with distinct sublayers within the backbone model. 
The parallel adapter can be formulated below:
\begin{equation}
     H_{o} \gets  H_{o}+f( H_{i}W_{down})W_{up},
\end{equation}
where $H_{i}$ ($H_{o}$) is the input (output) of a specific layer.

Expanding on this concept, the Multi-head Parallel Adapter takes it a step further by using parallel adapters to modify the outputs of head attention. On the other hand, the Scaled Parallel Adapter is a variant that applies the composition and insertion format of LoRA~\cite{lora} to adapters. Another approach, called Ladder Side-Tuning~\cite{ladder_side_tuning}, involves training a lightweight ladder side network. This network accepts intermediate activations from the backbone networks through shortcut connections (ladders).

\section{Authorship Attribution}
Authorship Attribution has garnered extensive attention in computational linguistics, with surveys by Zheng et al. \cite{zheng2023review} and Tyo et al. \cite{tyo2022state} offering detailed reviews of key developments and methodologies in the field. Initially, AA research focused on heuristic and statistical methods, employing basic stylometric features such as word lengths, sentence lengths, and function words to identify authors \cite{neal2017surveying, ding2017learning}. These features formed the basis for early attempts to link writing styles with author identities using classical machine learning algorithms \cite{boenninghoff2019similarity, boenninghoff2019explainable, theophilo2019needle}.

The advent of deep learning technologies marked a pivotal shift in AA research, introducing the capability to uncover more nuanced and complex patterns in writing styles. Studies by Zhang et al. \cite{zhang2015character}, Hitschler et al. \cite{hitschler2017authorship}, and Ruder et al. \cite{ruder2016character} explored the application of Convolutional Neural Networks (CNNs) in AA, reporting enhanced outcomes but also highlighting new challenges, such as the artificial ease introduced by automated bots and the over-reliance on specific textual features like hashtags and mentions. Shrestha et al. \cite{shrestha2017convolutional} further investigated CNNs for attributing authorship of social media messages, noting a decline in classification accuracy upon the removal of automated bot profiles from their dataset. In a significant advancement, Hu et al. \cite{hu2020deepstyle} integrated diverse embeddings—character, word, n-gram, and Part-Of-Speech (POS) tags—within a CNN framework, demonstrating improved AA accuracy for social media texts.

The integration of pre-trained language models, such as BERT \cite{devlin2019bert}, has significantly revolutionized the field of AA, enabling researchers to achieve unparalleled state-of-the-art results. By fine-tuning these models for specific AA tasks, studies by Rivera-Soto et al. \cite{rivera-soto-etal-2021-learning}, Manolache et al. \cite{manolache2021transferring}, Reimers et al. \cite{reimers2019sentence}, and El Boukkouri et al. \cite{el-boukkouri-etal-2020-characterbert} have demonstrated remarkable improvements in identifying authors based on their writing styles. Despite these advancements, a critical gap persists: these models often underperform in topic-shift scenarios, as observed by Altakrori et al. \cite{altakrori-etal-2021-topic-confusion}. Such scenarios, where the testing phase introduces topics absent from the training data, present significant challenges in distinguishing between an author's style and the content of the text.

Our proposed \textsf{ContrastDistAA} method seeks to bridge this gap by leveraging contrastive learning and mutual information techniques. This innovative approach is designed to disentangle content (i.e., topic) from linguistic style within the latent space, specifically addressing the limitations of pre-trained models in topic-shift scenarios. By focusing on the separation of these elements, \textsf{ContrastDistAA} aims to enhance the robustness and accuracy of AA, paving the way for more reliable authorship determination across a broader range of textual contexts.

\section{Contrastive Learning and Disentangled Representation Learning}

\subsection{Contrastive Learning}
Contrastive Learning has become a fundamental technique in developing embedding spaces that effectively cluster similar data points while distinguishing between dissimilar ones. This method has shown remarkable versatility, being applicable in both supervised and unsupervised learning scenarios to highlight the unique features of data instances. Its success is particularly notable in the domain of computer vision, where innovations like the data augmentation framework by Chen et al. \cite{chen2020simple}, and the Momentum Contrast (MoCo) mechanism for enhanced representation learning by He et al. \cite{9157636}, have set new benchmarks for performance.

Transitioning to the field of Natural Language Processing (NLP), contrastive learning has played a crucial role in improving sentence representation. This is evidenced by the work of Giorgi et al. \cite{giorgi2021declutr} and Gao et al. \cite{gao2022simcse}, who leveraged contrastive loss to refine textual embeddings, demonstrating the adaptability of this technique beyond visual data. Wu et al. \cite{wu2020clear} further explored sentence-level strategies, proposing methods to create vector subspaces that remain invariant to noise, a critical aspect for ensuring the fidelity of NLP models. The definition and utilization of negative samples have also seen innovative approaches, as seen in the work of Wang et al. \cite{wang2021cline}, who introduced a method for generating semantically rich negative examples to bolster model robustness and representation learning. Moreover, addressing the challenge of selecting hard negatives, Robinson et al. \cite{robinson2021contrastive} developed a user-controlled sampling technique, underscoring the technique's potential to enhance performance in downstream tasks across multiple modalities.

These advancements in contrastive learning, especially in the realm of NLP, provide a rich foundation for our work. By integrating these principles, our \textsf{ContrastDistAA} approach aims to leverage the strength of contrastive learning for disentangling content and stylistic features in text, addressing the specific challenges of AA in the face of topic variability and stylistic diversity.

\subsection{Disentangled Representation Learning}
Disentangled Representation Learning, which aims to isolate distinct attributes of data into separate variables, plays a pivotal role in advancing the state-of-the-art across various domains. This methodology facilitates a granular understanding and manipulation of data by segregating different aspects of the information it contains.

In computer vision, techniques like CycleGAN demonstrate the power of disentangled representations by using latent embeddings to achieve image translation without paired examples, showcasing its utility in scenarios with limited direct correspondences \cite{zhu2020unpaired}. Similarly, in speech processing, the separation of voice style from content through mutual information minimization exemplifies how disentangled representations can enhance clarity and control in model outputs \cite{yuan2021improving}.

The field of NLP has also seen significant contributions, with models such as ADNet employing motivational and adversarial losses to separate style and meaning within text embeddings \cite{romanov2019adversarial}. Addressing the challenge of learning from scarce parallel corpus data, Fu et al. \cite{fu2017style} introduced a multi-decoder model capable of distinguishing content representations from stylistic elements, facilitating effective style transfer. Further innovations include utilizing auxiliary multi-task and adversarial objectives for improved performance in text transfer tasks on non-parallel corpora \cite{john-etal-2019-disentangled}, and incorporating denoising objectives to balance generation quality with reconstruction accuracy \cite{shen2020educating}.

Inspired by these advancements, our \textsf{ContrastDistAA} framework adopts disentangled representation learning to meticulously separate content and stylistic features in textual data for AA. This approach allows for more precise authorship attribution by minimizing the confounding effects of topic variation, thereby enhancing the accuracy and applicability of AA across diverse textual domains.

\section{Authorship Verification}

In the last two decades, AV has evolved significantly, transitioning from traditional methods focusing on linguistic features like spelling and style to machine learning techniques~\citep{boenninghoff2019explainable}. However, traditional machine learning, such as support vector machines, showed limited effectiveness~\citep{konstantinou2022different}. Recent advancements involve contextual embeddings from language models like BERT, T5, and MPNET~\citep{bert,t5,mpnet}, and further studies have explored graph convolutional networks and BiLSTM with attention mechanisms~\citep{graphav,bilstmav}.

\textcolor{black}{Moreover, \citet{huang2024largelanguagemodelsidentify} explored different representations of authorship to verify their effectiveness in encoding writing styles. They concluded that authorship representations might be expected to remain robust against certain types of data shifts.}

Previous neural network-based methods, such as BERT and MPNET, have been crucial in advancing classification tasks, including AV. However, these models inherently lack mechanisms to elucidate their decision-making processes, a critical gap as the demand for explainability in AI grows \cite{promptav}. Addressing this, \citet{promptav} introduced PromptAV, an innovative technique that leverages the capabilities of large language models (LLMs) for AV, employing step-by-step stylometric explanation prompts to enhance interpretability. Their findings demonstrate that PromptAV significantly outperforms traditional approaches like chain of thought (CoT) prompting~\citep{wei2022chain} and PS+ prompting~\citep{wang2023planandsolvepromptingimprovingzeroshot} in both accuracy and interpretability, marking a noteworthy advancement in the application of LLMs to AV tasks.
\textcolor{black}{~\citet{huang2024largelanguagemodelsidentify} utilized LLMs with the Linguistically Informed Prompting (LIP) technique for authorship verification, revealing that even without domain-specific fine-tuning, the LIP method guides LLMs to satisfactory performance.}

Despite the progress, these methods' reliance on a few-shot demonstration model poses challenges in ensuring the consistency and relevance of their explanations. This limitation underscores a critical need for a more robust solution that can deliver both precise classification and meaningful explanations across a broader range of scenarios.
To address this need, we introduce the \textsf{InstructAV} framework, which adopts an instruction fine-tuning approach to significantly enhance the classification accuracy and explanation quality in AV tasks. By refining the model's ability to generate relevant and consistent explanations, \textsf{InstructAV} not only builds on the foundation established by PromptAV but also addresses its primary limitations, offering a comprehensive solution that advances the field of explainable AV.
\chapter{Text Style Transfer Overview and Experimental Evaluation} 

\label{tst_overview} 

\section{Style Definition and Task Formulation}
Before we dive into the details of this survey, we first introduce the basic terminology and concepts used throughout this survey. We will also describe the TST task formulation and summarize the notions commonly used in TST techniques.

\subsection{Definition of Style}
\label{sec:definition}
We discuss the definitions of text style from two perspectives: (a) the definitions of text style discussed in linguistic studies and (b) the definitions adopted in text style transfer literature.

\textbf{Linguistic View.} Style is an intuitive concept that pertains to the manner in which certain semantics are conveyed \cite{mcdonald-pustejovsky-1985-computational}. In this context, the semantics (or `\textit{content}' as referred to in this survey) of a text denotes the subject matter or argument the author intends to communicate. Text style constitutes the literary element that characterizes the author's use of language. This includes aspects such as word choice, sentence structure, figurative language, and sentence arrangement, all of which collectively contribute to establishing the tone, imagery, and semantics of the text. Style reflects the manner in which the author depicts events, objects, and ideas, thereby providing additional information that the reader can interpret and engage with. It is impractical to enumerate all possible styles, as each speaker possesses a unique set of linguistic techniques, often adapted to specific audiences, to fulfill their interpersonal objectives \cite{hovy1987generating}.

\textbf{Style in Text Style Transfer.}
In contrast to linguistic studies that offer a theoretical, rule-based definition of text style, research in TST adopts a more data-driven approach. TST studies typically define "style" as text style attributes or labels that are derived from style-specific corpora. These corpora generally contain attributes that neural machine learning techniques can model with relative ease. For instance, sentiment transfer tasks (where \textit{positive} and \textit{negative} serve as style attributes) and formality transfer tasks (where \textit{formal} and \textit{informal} serve as style attributes) are among the most prevalent benchmarks for assessing TST performance. A more comprehensive list of evaluation tasks and available corpora will be provided in Section \ref{sec:evaluation}. It is, however, debatable whether the sentiment of a text should be considered as an aspect of its style. The primary objective of TST is to alter the style of a text while maintaining its original semantics. One could argue that sentiment transfer inherently modifies the semantics of a sentence. Nevertheless, from the perspective of current deep learning methods for TST, most existing models discussed in this survey are applicable to any datasets labeled with varying stylistic attributes, including sentiment, formality, gender, and political orientation.

\subsection{Task Formulation}
The TST task aims to change the stylistic properties of any given text while preserving its style-independent content. 
The input includes a set of attributes $\mathcal{A}$ with text for each attribute in the corpus $\bm{X}$.  
For example, for the formality transfer task, there are two attributes: formal and informal. The task consists of taking the sentence $\bm{x}$ with the source attribute $s$ (e.g., formal) and generating the sentence $\bm{x'}$ with the target attribute $t$ (e.g., informal) while preserving the style-independent content. The style corpus $\bm{X}$ can be parallel or non-parallel. In the parallel setting, for each sentence with the source attribute $s$, a counterpart sentence with the same style-independent content with the target attribute $t$ is contained in $\bm{X}$. For the non-parallel setting, there is no alignment information among sentences with different attributes.

\section{A Taxonomy of Text Style Transfer Methods}
\label{sec:taxonomy}
This section first proposes a taxonomy to organize the most notable and promising advances in TST research in recent years. Then we discuss each category of TST models in detail. 

\subsection{Categories of Text Style Transfer Models}
\begin{figure}[t]
    \centering
    \includegraphics[width=\textwidth]{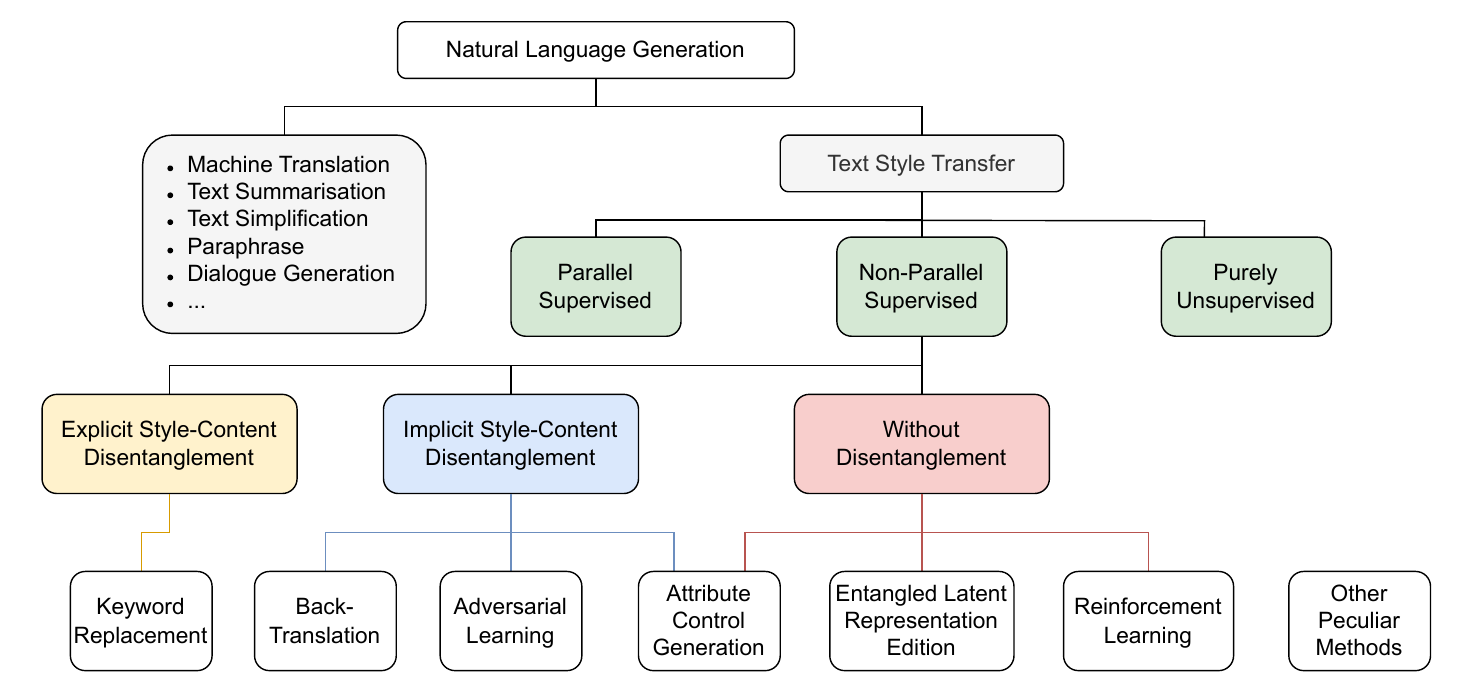}
    \caption{A taxonomy of text style transfer methods}
    \label{fig:taxonomy}
\end{figure}

To provide a comprehensive overview of this field, we classify existing TST models based on three criteria: (1) data settings, (2) strategies, and (3) techniques. Figure \ref{fig:taxonomy} presents a summary taxonomy of TST methods. We consider TST as one of several tasks within natural language generation. The techniques developed for TST can be broadly categorized according to the data settings employed during training, namely parallel supervised, non-parallel supervised, and purely unsupervised settings. Given that recent advancements predominantly focus on designing non-parallel supervised TST techniques, our survey is specifically scoped to explore these techniques in greater detail. In particular, we examine the broad strategies utilized by non-parallel supervised TST techniques, along with various purely unsupervised TST approaches.

\subsection{Data Settings}
\label{sec:tax_datasetting}
We generally classify existing TST studies into three categories based on the data settings used for model training. 

\begin{itemize}
    \item \textbf{Parallel Supervised.} In this data setting, the TST models are trained with known pairs of text with different styles. Commonly, NMT methods such as {\em sequence-to-sequence} (Seq2Seq) models \cite{xu2012paraphrasing,cho2014learning,sutskever2014sequence,DBLP:journals/corr/BahdanauCB14} are applied to transfer the style of text. For example, Jhamtani et al. \cite{jhamtani2017shakespearizing} trained a Seq2Seq model with a pointer network on a parallel corpus and applied the model to translate modern English phrases to Shakespearean English. Details of the techniques applied on parallel datasets will be discussed in Section \ref{sec:techniques}
    \item \textbf{Non-Parallel Supervised.} TST models in the non-parallel supervised setting aim to transfer the style of text without any knowledge of matching text pairs in different styles. Most of the existing TST studies fall into this category due to the scarcity of parallel datasets in real-world TST applications.
    \item \textbf{Purely Unsupervised.} In both parallel and non-parallel supervised data settings, the style labels are available to enable supervised training of the TST models. A more challenging setting is purely unsupervised where only an unlabeled text corpus is available, and the TST models need to be trained in an unsupervised fashion to perform text style transfer without any knowledge of style labels.
\end{itemize}

\subsection{Strategies}
\label{sec:strategy}
In order to perform TST in the \textit{non-parallel supervised} setting, existing studies have proposed to disentangle the style and content in text, which is a strategy commonly used in NST \cite{gatys2015neural}. 

Within the TST research community, the possibility of disentangling style from content has been a topic of extensive discussion and remains an unresolved research question. As outlined in Section \ref{sec:definition}, the semantics or content of a text refers to the subject matter or argument the author intends to convey, whereas text style pertains to the literary elements through which the text is communicated. These literary elements include word choice, syntactic structures, figurative language, and sentence arrangement. Despite the precise definitions of content and style, the feasibility of disentangling them within a text remains uncertain. Intuitively, an author's stylistic choices may be influenced by the content being communicated. For instance, when considering the sentiment of texts related to food reviews, some foods may be universally perceived as tastier than others—such as pizza, which is generally favored by many individuals. Consequently, a food review discussing pizza is likely to be associated with a positive sentiment. Such confounding factors make it challenging to disentangle style from content without additional assumptions or domain-specific knowledge.

Lample et al. \cite{lampleSSDRB19} also contend that style-content disentanglement presents significant challenges, demonstrating that the adversarial method proposed by Fu et al. \cite{fu2018style} did not effectively separate style and content. However, it is important to note that Lample et al. \cite{lampleSSDRB19} conducted their evaluation solely on the model developed by Fu et al. \cite{fu2018style}. The question of whether style and content can be successfully disentangled, and the methods by which this might be achieved, remains an open area of research. We encourage researchers within the TST community to further explore methodologies for more effectively disentangling style and content, and to establish whether such disentanglement is achievable from various perspectives.

In this chapter, we discuss three types of style-content disentanglement strategies and categorize the existing non-parallel supervised TST methods into one of these three strategies:

\begin{itemize}
    \item \textbf{Explicit Style-Content Disentanglement.} This strategy employs a direct text replacement approach within TST models to generate texts in a target style. For instance, Li et al. \cite{li2018delete} initially identified segments of text related to the original style and subsequently replaced them with phrases corresponding to the target style. The modified text, containing the newly substituted phrases, was then processed through a Seq2Seq model to produce a fluent text sequence in the target style. The specific techniques employed to disentangle content from style will be thoroughly examined in Section \ref{sec:techniques}.
    \item \textbf{Implicit Style-Content Disentanglement.} To implicitly disentangle style from content in text, TST models initially focus on learning the latent representations of the content and style within a given text sequence. Subsequently, the latent representation of the original text’s content is amalgamated with the latent representation of the target style to generate new text employing the target style. Various techniques, including back-translation, adversarial learning, and controllable generation \cite{shen2017style,zhao2018adversarially,fu2018style,prabhumoye2018style,hu2017toward}, have been proposed to achieve the disentanglement of latent representations of content and style.
    \item \textbf{Without Style-Content Disentanglement.} Recent research indicates that assessing the quality of text style and content disentanglement poses difficulties and that such disentanglement may be unnecessary for effective TST \cite{lampleSSDRB19}. Consequently, subsequent studies have focused on executing TST without separating a text's style and content. Techniques such as adversarial learning, controllable generation, reinforcement learning, probabilistic modeling, and pseudo-parallel corpus construction \cite{lampleSSDRB19, dai2019style, li2019domain, luo2019dual, He2020A} have been utilized to perform TST without requiring the disentanglement of text content and style.
\end{itemize}

\subsection{Techniques}
\label{sec:techniques}

Table~\ref{tbl:taxonomy} enumerates the various techniques commonly employed for TST. These techniques are organized according to the previously outlined taxonomy, and each technique will be reviewed in detail in the subsequent subsections. Furthermore, we provide references to the literature that have utilized these techniques.

\begin{table}[h]
\centering
\small
\caption{Publications Based on Different Text Style Transfer Techniques}
\label{tbl:taxonomy}
\begin{tabular}{|c|p{1.2cm}|c|p{2.5cm}|}
\hline
\textbf{Data Setting} & \textbf{\begin{tabular}[c]{@{}p{1.2cm}@{}} Strategy \end{tabular}} & \textbf{Technique} & \multicolumn{1}{c|}{\textbf{Literature}} \\ \hline
Parallel Supervised & - & Sequence-to-Sequence  & \cite{jhamtani2017shakespearizing,carlson2018evaluating,shang2019semi,wang2019harnessing,jin2019imat,nikolov2018large,liao2018quase,xu2019formality,zhang2020parallel}\\ \hline
{Non-Parallel} Supervised & Explicit  & Explicit Style Keyword Replacement & \cite{li2018delete,xu2018unpaired,zhang2018learning,sudhakar2019transforming,wu2019hierarchical} \\ \cline{2-4} & {Implicit}  & Back-Translation & \cite{prabhumoye2018style,zhang2019machine} \\ \cline{3-4} 
 & & Adversarial Learning  &  \cite{shen2017style,zhao2018adversarially,fu2018style,chen2018adversarial,logeswaran2018content,yin2019utilizing,zhao2018language,yang2018unsupervised,lai2019multiple,john2019disentangled,park2019paraphrase}\\ \cline{3-4} 
 &  & Attribute Control Generation & \cite{hu2017toward,tian2018Structured} \\ \cline{3-4}  & & Other Peculiar Methods  &  \cite{DBLP:conf/acl/ChengMSMZLC20} \\ \cline{2-4} 
 & {Without} & Attribute Control Generation &  \cite{lampleSSDRB19,dai2019style,zhang2018shaped,jain2019unsupervised,zhou2020exploring} \\ \cline{3-4} 
 & & Entangled Latent Representation Edition & \cite{mueller2017sequence,xu2019variational,wang2019controllable,liu2020revision}\\ \cline{3-4} 
 & & Reinforcement Learning & \cite{luo2019dual,gong2019reinforcement} \\ \cline{3-4} 
 & & Other Peculiar Methods  &  \cite{He2020A,DBLP:conf/acl/ChengMSMZLC20,syed2020adapting} \\ \hline
 Purely Unsupervised & - & Purely Unsupervised  & \cite{radford2017learning,xu2019variational, shen2020educating,DBLP:conf/iclr/DathathriMLHFMY20} \\ \hline
\end{tabular}
\end{table}

\paragraph{Sequence-to-Sequence Model with Parallel Data.} 
The Seq2Seq model, which is based on the encoder-decoder architecture \cite{xu2012paraphrasing, cho2014learning, sutskever2014sequence, DBLP:journals/corr/BahdanauCB14}, serves as a fundamental component in numerous natural language generation tasks \cite{gatt2018survey}, including TST. Typically, a Seq2Seq model is trained using a parallel corpus, where text in the original style is processed by the encoder, and the decoder generates the corresponding text in the target style. Several Seq2Seq models for TST have been developed and trained on parallel datasets. Jhamtani et al. \cite{jhamtani2017shakespearizing} expanded upon the work of Xu et al. \cite{xu2012paraphrasing} by incorporating a pointer network \cite{vinyals2015pointer} into the Seq2Seq model, enabling the selective copying of word tokens from the input text to facilitate the transfer from modern English to Shakespearean English. This direct copying mechanism was inspired by the observation that Shakespearean and modern English share significant vocabulary overlap, and certain rare words, such as infrequent nouns, are challenging for a Seq2Seq model to generate. Carlson \cite{carlson2018evaluating} proposed a Seq2Seq model with attention mechanisms and evaluated its performance using a parallel Bible-prose-style corpus.

\paragraph{Explicit Style Keyword Replacement}
\label{sec:tax-explicit}

Text style attributes, such as sentiment, are frequently characterized by distinctive keywords and phrases. For instance, words like "nice" and "good" often suggest positive sentiment, whereas words like "bad" and "nasty" imply negative sentiment. Consequently, a straightforward approach for transferring a text's sentiment is to substitute keywords and phrases that are linked to specific sentiments. Drawing on this intuitive concept, researchers have devised TST methods that explicitly disentangle content and style by replacing keywords associated with a particular style \cite{li2018delete, xu2018unpaired, zhang2018learning, sudhakar2019transforming, wu2019hierarchical}.

\paragraph{Adversarial Learning}

To address the limitations associated with explicit style keyword replacement methods, researchers have explored an alternative approach to implicitly disentangle text content and style information for performing TST. Adversarial learning is a widely employed technique in many implicit content-style disentanglement TST methods \cite{shen2017style, zhao2018adversarially, fu2018style, chen2018adversarial, logeswaran2018content, yin2019utilizing, zhao2018language, lai2019multiple, john2019disentangled, park2019paraphrase, yang2018unsupervised}. Generally, these TST methods utilize adversarial learning for two primary objectives: (a) to generate text outputs that are indistinguishable from real data \cite{shen2017style, logeswaran2018content, tian2018Structured, chen2018adversarial, zhao2018language, yin2019utilizing, park2019paraphrase}; and (b) to eliminate style attributes from the latent representation of the text \cite{zhao2018adversarially, fu2018style, yang2018unsupervised, lai2019multiple, john2019disentangled}. A seminal framework for the latter objective was proposed by Fu et al. \cite{fu2018style}.

\paragraph{Back-Translation} 

Back-translation has been utilized in neural machine translation to create artificial corpora \cite{sennrich2015improving}. This technique has also been employed to generate pseudo-parallel datasets for TST \cite{zhang2020parallel}. In this context, this section examines how back-translation has been investigated to derive content representations devoid of style for TST \cite{prabhumoye2018style, zhang2019machine}. This approach is inspired by the findings of Rabinovich et al. \cite{rabinovich-etal-2017-personalized}, who determined that certain author traits, such as gender, are obfuscated in both human and machine translations. However, it is important to consider two caveats: (1) Rabinovich et al. demonstrated only the obfuscation of gender information in translation processes, without exploring other text attributes such as sentiment, tense, and politeness; (2) they did not conduct quantitative experiments to measure the extent of lost gender information.

\paragraph{Attribute Control Generation}

Attribute-controlled generation is becoming an increasingly prevalent technique in TST models \cite{hu2017toward, tian2018Structured, lampleSSDRB19, dai2019style, zhang2018shaped, jain2019unsupervised, zhou2020exploring}. This approach typically involves learning an attribute code, denoted as $\bm{a}$, to guide text generation across different styles. There are two primary strategies for implementing attribute-controlled generation in TST: (a) implicitly disentangling the latent representation $\bm{z}$ to ensure it contains only content information while concurrently learning the style attribute code; (b) operating without disentangling or constraining the latent representation $\bm{z}$, thus allowing it to hold more than just content information while learning the style attribute code. In both strategies, classifier-guided loss is employed to ensure that the generator $G$ produces a sentence $\bm{x'}$ with the desired style attribute. Specifically, the loss function aims to minimize the following:

\begin{equation}
    L_{\text{Cla}}(\Theta_G,t) = - \mathbb{E}_{p(\bm{x})}[\log D(\bm{x'})],
\end{equation}
Here, $D$ represents a style classifier that has been pre-trained on real data $\bm{x}$. In a manner similar to adversarial training applied to generated sentences, $L_{\text{Cla}}$ can be optimized using the Gumbel-softmax distribution \cite{DBLP:conf/iclr/JangGP17} or the policy gradient algorithm \cite{williams1992simple}. It is important to note that both adversarial training on generated sentences and classifier-guided loss share the objective of ensuring that the transformed sentence $\bm{x'}$ embodies the target attribute $t$. Despite this common goal, the two methodologies employ distinct loss functions.

Given that the attribute-controlled generation technique also relies heavily on the style classifier, it shares several limitations with methods based on adversarial learning. Specifically, the accuracy of the style classifier constrains the ability to effectively learn a robust style attribute code for performing TST. Nonetheless, should the style attribute code be successfully learned, it holds the potential to improve other NLG tasks. For instance, a pre-trained style attribute code can be utilized within machine translation to not only translate languages but also generate text in specific styles.

\paragraph{Entangled Latent Representation Editing}
\begin{figure}[htb]
    \centering
    \includegraphics[scale = 0.6]{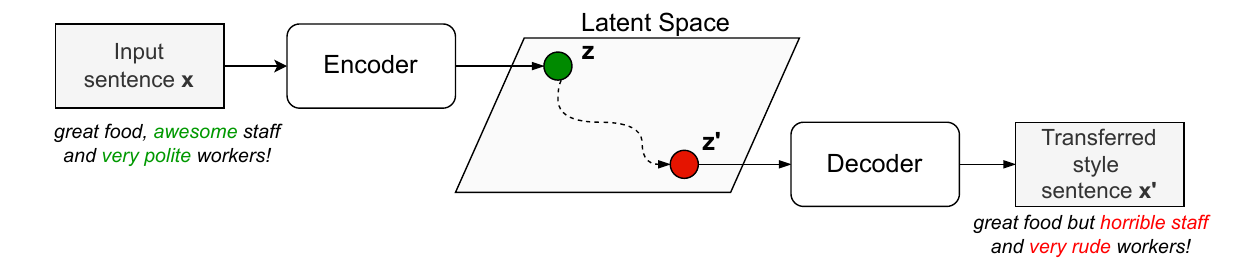}
    \caption{Common framework for editing text's latent representation for TST.}
    \label{fig:edit}
\end{figure}

Another approach to performing TST without disentangling content and style involves directly editing the latent representations learned by autoencoder-based models. Figure \ref{fig:edit} illustrates a common framework employed in research focused on modifying latent representations for TST. Typically, the latent representation $\bm{z}$ developed through an autoencoder is altered using various techniques. In this framework, a style classifier is jointly trained with the autoencoder, where the training process iteratively updates the latent representation $\bm{z}$ within a constrained space to maximize the prediction confidence score associated with the target attribute label of the style classifier. Specifically, each update is derived from the gradient of the style classifier loss with respect to $\bm{z}$. The modified latent representation $z'$ is subsequently inputted into the decoder to generate text in the desired style.

\paragraph{Reinforcement Learning} 

\begin{figure}[htb]
    \centering
    \includegraphics[scale = 0.6]{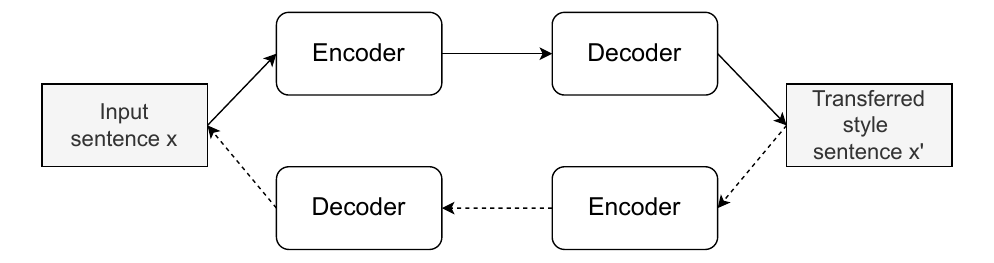}
    \caption{Dual reinforcement learning model for TST proposed in \protect\cite{luo2019dual}.}
    \label{fig:reinforce}
\end{figure}

Reinforcement learning has been applied as a technique for TST. The fundamental concept of reinforcement learning-based TST is to employ specially designed reward functions to guide the transfer process, rather than relying on the various loss functions used in other TST methods. The policy gradient algorithm \cite{williams1992simple} is utilized to maximize the expected reward of the transferred text, thereby optimizing the parameters of the reinforcement learning TST model. This algorithm facilitates easier training by circumventing the challenges posed by discrete training inherent in automatic regression decoding processes. However, due to the high variance associated with the sampling gradient, training using this method can be unstable. For instance, Luo et al. \cite{luo2019dual} proposed learning two Sequence-to-Sequence (Seq2Seq) models across two styles using reinforcement learning, without disentangling style and content. Figure \ref{fig:reinforce} illustrates the proposed dual reinforcement learning framework. The authors approached the learning of source-to-target and target-to-source styles as a dual-task. They designed the style classifier reward, $R_s$, and reconstruction reward, $R_c$, to promote style transfer accuracy and content preservation. The overall reward, calculated as the harmonic mean of these two rewards, was used as a feedback signal to steer learning within the dual-task framework. This approach enables training through reinforcement learning without the need for parallel data or content-style disentanglement.

\paragraph{Purely Unsupervised Methods}
The majority of the TST studies mentioned above operate under the assumption that style-specific corpora, whether parallel or non-parallel, are accessible. This section introduces a different category of TST models capable of performing style transfer in a purely unsupervised context, wherein only a mixed corpus of unspecified style is available.

There are relatively fewer studies proposing methods for performing TST in a purely unsupervised setting \cite{radford2017learning, jain2019unsupervised, xu2020variational, shen2020educating}. In an earlier investigation, Radford et al. \cite{radford2017learning} examined the properties of byte-level recurrent language models, wherein an LSTM model is trained on text that is preprocessed as sequences of UTF-8 encoded bytes, all within an unsupervised framework. Notably, the researchers identified a single neuron unit within the trained LSTM model that directly correlates with sentiment, and utilized manipulations of neuron units to transfer sentiments across sentences. Xu et al. \cite{xu2020variational} conducted a similar study, detecting a latent dimension responsible for sentiment with over 90\% accuracy. They subsequently employed unsupervised representation learning to disentangle style and content from a mixed corpus of unspecified styles, achieving satisfactory results in sentiment transfer tasks.

Jain et al. \cite{jain2019unsupervised} introduced an unsupervised training approach for text formalization utilizing unlabeled data. To facilitate the unsupervised training of the encoder-decoder framework, they employed external scoring mechanisms to supply style information. These scorers, relying on established language processing tools, determined the learning scheme of the encoder-decoder based on the actions it undertakes. The authors annotated the formality level of sentences using the scores provided by the external scorers, enabling the TST model to effectively capture formality information.

Shen et al. \cite{shen2020educating} expanded upon adversarial auto-encoders (AAE) by incorporating a denoising objective that reconstructs original sentences from their perturbed versions. The proposed denoising AAE model is designed to map similar sentences to analogous latent representations while delineating the boundaries between different representation clusters more distinctly. For sentiment transfer tasks, they calculated a single "sentiment vector" by averaging the latent code $\bm{z}$ separately for 100 non-parallel positive and negative sentences within the development set. Subsequently, they determined the difference between these two averages. To alter the sentiment of a sentence from the test set—by converting a positive sentiment to negative or vice versa—they employed a straightforward addition or subtraction of the sentiment vector.

The competitive performance of purely unsupervised TST methods is promising. However, it should be noted that most of the methods mentioned above were primarily evaluated on sentiment transfer tasks. In particular, the studies by Radford et al. \cite{radford2017learning} and Xu et al. \cite{xu2020variational}, which identified the latent attribute corresponding to sentiment in text, do not clearly demonstrate whether these methods can be extended to other stylistic properties, such as text formality. Further research is necessary to assess whether purely unsupervised methods can be effectively generalized to other TST tasks.

\section{Reproducibility Study}
\label{sec:evaluation}

While most existing TST methods have been evaluated in their original studies using downstream tasks such as sentiment and formality transfer, these experiments frequently relied on few or no baseline comparisons. In response, we conducted a reproducibility study\footnote{Code implementations of the reproduced models are compiled in this repository: \url{https://gitlab.com/bottle_shop/style/tst_survey}} and benchmarked 19 TST models on two widely recognized corpora: Yelp reviews and the GYAFC, corresponding to sentiment transfer and formality transfer tasks, respectively. To our knowledge, this represents the first instance where such a comprehensive array of TST models has been evaluated concurrently on the same datasets. Notably, the experimental results from this study offer valuable insights into the performance of each TST algorithm and their comparative effectiveness, specifically in terms of \textit{transfer strength}, \textit{content preservation}, and \textit{fluency}. This section is structured as follows: Firstly, we outline the experimental setup of our reproducibility study. Subsequently, we present and analyze the experimental results for both \textit{sentiment transfer} and \textit{formality transfer} tasks. Additionally, we conduct trade-off analyses to examine how the relationships between different evaluation criteria impact the performance of TST models. Finally, a human evaluation is performed on a subset of representative TST models, and we report the findings.

\subsection{Experimental Setup}
\textbf{Environment Settings.} The experiments were conducted on a system running Ubuntu 18.04.4 LTS, equipped with 24 cores, 128 GB RAM, and a clock speed of 2.9 GHz. For models based on deep neural networks, an Nvidia GTX 2080Ti GPU was utilized. We adhered to the environmental requirements and hyperparameter settings specified in the released code implementations of the TST models to reproduce the experimental results accurately. Table \ref{tbl:dataset_statistics} provides details regarding the training, validation, and test splits of the Yelp and GYAFC datasets employed in our experiments.

 \begin{table}[t]
\centering
\caption{Dataset statistics for Yelp and GYAFC.}
\label{tbl:dataset_statistics}
\begin{tabular}{|c|c|c|c|c|c|}
\hline
Dataset & Subset & Attributes & Train & Dev & Test \\
\hline
\multirow{2}*{Yelp} & \multirow{2}*{-} & Positive & 267,314 & 38,205 & 76,392 \\
\cline{3-6}
 & & Negative & 176,787 & 25,278 & 50,278 \\
\hline
\multirow{2}*{GYAFC} & \multirow{2}*{F\&R} & Informal & 51,967 & 2,788 & 1,332 \\
\cline{3-6}
&  & Formal & 51,967 & 2,247 & 1,019 \\
\hline
\end{tabular}
\end{table}

\textbf{Evaluation Metrics.} We adopt the following evaluation metrics to measure the performance of the TST models. Specifically, we apply Style Transfer Accuracy (\textit{ACC}) to measure the transfer strength. To measure content preservation, we adopt \textit{BLEU}, \textit{sBLEU}, Cosine Similarity (\textit{CS}), and Word Overlap (\textit{WO}). Noted that the \textit{BLEU} score is only computed for the GYAFC dataset as the human references of the sentences in test set are available. We compute the perplexity score (\textit{PPL}) to quantify the fluency of the transferred sentences. In practice, we use the script \textbf{multi-bleu.perl} \footnote{\url{https://github.com/OpenNMT/OpenNMT/blob/master/benchmark/3rdParty/multi-bleu.perl}} to calculate \textit{BLEU} and \textit{sBLEU} scores. For \textit{WO}, we exclude both stopwords and
sentiment words \footnote{\url{https://www.cs.uic.edu/~liub/FBS/sentiment-analysis.html}}. Finally, we compute two average metrics that consider all the aspects of evaluation:
   
\begin{itemize}
    \item Geometric Mean (\textit{G-Score}): We compute the geometric mean of \textit{ACC} (transfer strength), \textit{sBLEU} (content preservation), \textit{WO} (content preservation) and \textit{1/PPL} (fluency). We excluded the measure \textit{CS} in the mean calculation due to its insensitivity and take the inverse of the calculated perplexity score because a lower \textit{PPL} score corresponds to better fluency.
    \item Harmonic Mean (\textit{H-Score}): Different averaging methods reflect different priorities. Thus, we also compute the harmonic mean of \textit{ACC}, \textit{sBLEU}, \textit{WO}, and \textit{1/PPL}.
\end{itemize}

\textbf{Reproduced Models.} Our reproducibility study is confined to these 19 TST models because they have made their implementation codes publicly available. We aspire to encourage other researchers to publish their codes and datasets, thereby fostering development in this field. Moreover, we have classified and color-coded the models according to the strategies outlined in Section \ref{sec:taxonomy}. \orange{Orange} indicates methods that explicitly disentangle style and content, \green{green} designates methods that apply an implicit disentanglement strategy, and \blue{blue} corresponds to methods that perform TST without disentangling style and content. Specifically, we reproduced and implemented the following TST models:

\orange{\textbf{Explicit Style-Content Disentanglement}}

\begin{itemize}

    \item \orange{\textbf{DeleteOnly; Template; Del\&Retri}} \cite{li2018delete}: The style keyword replacement method explicitly disentangles the style and content of a sentence by replacing specific keywords. The authors introduce three variants of their approach: \orange{DeleteOnly}, \orange{Template}, and \orange{Del\&Retri}. The \orange{DeleteOnly} variant initially removes style-attributed keywords from the source sentence. Afterwards, the latent representation of the source sentence is combined with the target style attribute and processed through a sequence model to generate a sentence in the desired style. The \orange{Template} model replaces the deleted style-attributed keywords directly with those of the target style. The \orange{Del\&Retri} model begins with the same keyword removal as the \orange{DeleteOnly} method. It then retrieves a new sentence associated with the target style attribute. Finally, both the keyword-removed source sentence and the retrieved sentence are input into a sequence model to produce the sentence in the transferred style.
    
    \item \orange{\textbf{B-GST; G-GST}} \cite{sudhakar2019transforming}: This TST model builds upon the work of \cite{li2018delete} and introduces the Generative Style Transformer (GST) for text style transfer. There are two versions of the GST model: the Blind Generative Style Transformer (\orange{B-GST}) and the Guided Generative Style Transformer (\orange{G-GST}).

    \item \orange{\textbf{PTO}} \cite{wu2019hierarchical}: This model employs a style keyword replacement technique, utilizing reinforcement learning to implement a hierarchical \textit{Point-Then-Operate} (\orange{PTO}) sequence operation. The \orange{PTO} process involves two agents: a high-level agent that iteratively recommends positions for operations, and a low-level agent that modifies the sentence based on these high-level suggestions. This framework explicitly replaces style-attributed keywords to achieve text style transfer.
    
    \item \orange{\textbf{UST}} \cite{xu2018unpaired}: This style keyword replacement method utilizes cycled reinforcement learning to iteratively replace style-attributed keywords while preserving content for text style transfer. Originally, this model was developed for sentiment transfer tasks.
    
    \item \orange{\textbf{SMAE}} \cite{zhang2018learning}: This model employs a style keyword replacement approach for TST by explicitly disentangling style and content. Initially designed for sentiment transfer, it detects sentiment attribute words and then uses a sentiment-memory-based auto-encoder to modify sentiment without relying on parallel data.

All the explicit style-content disentanglement methods are discussed in detail in Section \ref{sec:tax-explicit}.

\end{itemize}

\green{\textbf{Implicit Style-Content Disentanglement}}

\begin{itemize}
    
    \item \green{\textbf{DRLST}} \cite{john2019disentangled}: This adversarial learning TST model incorporates auxiliary multi-task and adversarial objectives for style prediction and bag-of-word prediction, respectively, to facilitate text style transfer.

    \item \green{\textbf{BST}} \cite{prabhumoye2018style}: This back-translation-based TST model utilizes a pre-trained back-translation model to rephrase sentences, reducing their stylistic attributes. Separate style-specific decoders are then employed to perform style transfer.
    
    \item \green{\textbf{CAAE}} \cite{shen2017style}: This adversarial learning TST model implicitly disentangles the style of the text by assuming a shared latent content distribution across different text corpora. It proposes a method leveraging refined alignment of latent representations to achieve text style transfer.
    
    \item \green{\textbf{Ctrl-Gen}} \cite{hu2017toward}: This attribute-controlled TST model uses variational auto-encoders and a style classifier to guide the learning of style attributes, thereby controlling text generation across various styles.
    
    \item \green{\textbf{ARAE}} \cite{zhao2018adversarially}: A general natural language generation approach that employs adversarial learning to alter specific attributes within text. TST is one of the applications proposed for this model in its original publication.
    
    \item \green{\textbf{Multi-Dec; Style-Emb}} \cite{fu2018style}: This adversarial learning TST model employs a style classifier to disentangle style from content representation for the style transfer task. Two variants are proposed: the multi-decoder (\green{Multi-Dec}) model uses different decoders to generate text with varying styles, while the style embedding (\green{Style-Emb}) model concatenates a style embedding vector with content representation for generating text in different styles using a single decoder.
    
\end{itemize}

\blue{\textbf{Without Style-Content Disentanglement}}
\begin{itemize}

    \item \blue{\textbf{DualRL}} \cite{luo2019dual}: This reinforcement learning-based TST model employs two seq2seq models to facilitate text style transfer between two distinct styles. It approaches the learning process for transferring from source-to-target style and target-to-source style as a dual task, where each task mutually reinforces the other, enabling TST without disentangling style and content.

    \item \blue{\textbf{DAST; DAST-C}} \cite{li2019domain}: This attribute-controlled TST model executes text style transfer in a domain-aware manner. Two variations are proposed: the Domain Adaptation Style (\blue{DAST}) model and the \blue{DAST} with Generic Content Information (\blue{DAST-C}). These models learn latent style attributes and domain vectors to accomplish TST across different domains.
    
    \item \blue{\textbf{PFST}} \cite{He2020A}: This probabilistic deep generative TST model integrates a language model prior with an encoder-decoder transducer to infer latent representations of sentences within an assumed parallel-style corpus. These inferred latent representations are used to generate sentences of specific styles through a decoder.
    
\end{itemize}

\subsection{Sentiment Transfer} 

\begin{table}[htb]
\centering
\small
\caption{TST results in Yelp dataset (sentiment transfer task).}
\label{tbl:resultyelp}
\begin{tabular}{cccccccc}
\hline
Model & \textit{ACC(\%)} & \textit{sBLEU} & \textit{CS} & \textit{WO} & \textit{PPL} & \textit{G-Score} & \textit{H-Score} \\\hline
\orange{DeleteOnly} &84.2 &28.7 &0.893 &0.501  &115  &1.80  &0.034 \\
\orange{Template}  &78.2 &48.1 &0.850 &0.603 &1959  &1.04  &0.002 \\
\orange{Del\&Retri}  &88.1 &30.0 &0.897 &0.464 &101  &1.87  &0.039  \\
\orange{B-GST} &89.2 &46.5 &0.959 &0.649  &216  &1.88  &0.018  \\
\orange{G-GST} &72.7  &52.0  &0.967  &0.617  &407  &1.55  &0.010  \\
\orange{PTO} &82.3 &57.4 &\textbf{0.982} &0.737  &245  &1.94  &0.016  \\
\orange{UST} &74.0  &41.0  &0.929  &0.448  &394  &1.36  &0.010 \\
\orange{SMAE} &84.4  &14.8  &0.907  &0.294  &210  &1.315  &0.019  \\
\hline
\green{DRLST} &91.2 &7.6 &0.904  &0.484  &\textbf{86} & 1.41 & \textbf{0.045} \\
\green{BST} &83.1 &2.3 &0.827 &0.076 &261  &0.49  &0.015 \\
\green{CAAE} &82.7 &11.2 &0.901 &0.277  &145  &1.15  &0.027  \\
\green{ARAE} &83.2 &18.0 &0.874 &0.270  &138  &1.31  &0.028  \\
\green{Ctrl-Gen} &89.6 &49.5 &0.953 &0.707  &384  &1.69 &0.010  \\
\green{Multi-Dec} &69.6 &17.2  &0.887  &0.244  &299  &0.99  &0.013  \\
\green{Style-Emb} &47.5 &31.4  &0.926  &0.433  &217  &1.31  &0.018 \\
\hline                                                                    
\blue{DualRL}  &79.0 &\textbf{58.3} &0.97 &\textbf{0.801} &134 & \textbf{2.29}  &0.030  \\
\blue{DAST} &90.7  &49.7  &0.961  &0.705  &323  &1.77  &0.012 \\
\blue{DAST-C} &\textbf{93.6}  &41.2  &0.933  &0.560  &450  &1.48  &0.009 \\
\blue{PFST} &85.3  &41.7  &0.902  &0.527  &104  &2.06  &0.038 \\
\hline
\end{tabular}
\end{table}

Table \ref{tbl:resultyelp} presents the performance of various TST models on the sentiment transfer task. It is evident that no TST model excelled in all evaluation metrics. However, \blue{DualRL}, \orange{PTO}, \orange{B-GST}, and \blue{PFST} demonstrated a well-balanced trade-off between text fluency, content preservation, and style transfer accuracy. While \green{DRLST} achieved the second-best transfer accuracy, it suffered from a notably low \textit{sBLEU} score, indicating its ineffectiveness in preserving the original sentence's content. Additionally, it is observed that most implicit style-content disentanglement methods exhibited poor performance in content preservation, potentially due to the loss of some content information during the disentanglement process. Furthermore, different averaging methods, specifically \textit{G-Score} and \textit{H-Score}, assigned varying weights to the evaluation metrics. For instance, \textit{H-Score} prioritizes the perplexity scores of generated sentences more heavily. Consequently, \green{DRLST}, which had the lowest \textit{PPL} score, also attained the highest \textit{H-Score}. Conversely, the \orange{Template} model, which had the highest \textit{PPL} score, received the lowest \textit{H-Score}.

Interestingly, our observations indicate that style keyword replacement methods, such as \orange{DeleteOnly}, \orange{Template}, \orange{Del\&Retri}, \orange{B-GST}, \orange{G-GST}, \orange{UST}, \orange{PTO}, and \orange{SMAE}, demonstrate strong performance in sentiment transfer tasks. These methods have achieved high transfer accuracy while effectively preserving content information, as evidenced by high scores in \textit{sBLEU}, \textit{CS}, and \textit{WO}. A plausible explanation for their success may stem from the nature of the task; the sentiment of a sentence can be efficiently altered by substituting keywords associated with the original sentiment. For instance, replacing "fresh" with "rotten" alters the sentiment from positive to negative. However, it is noteworthy that the \orange{Template} method—an algorithm focused on replacing sentiment-related keywords—exhibits a high perplexity score, indicating suboptimal performance in sentence fluency. This observation underscores the need for more sophisticated generative approaches, which can mitigate the creation of implausible sentences resulting from simplistic keyword substitution.

\subsection{Formality Transfer}

\begin{table}[htb]
\centering
\small
\caption{TST results in the GYAFC dataset (formality transfer task).}
\label{tbl:resultGYAFC}
\begin{tabular}{cccccccccc}
\hline
Model & \textit{ACC(\%)} & \textit{sBLEU} & \textit{hBLEU} & \textit{CS} & \textit{WO} & \textit{PPL} & \textit{G-Score} & \textit{H-Score}  \\\hline
\orange{DeleteOnly} &26.0 &35.4 &16.2 &0.945 &0.431  &82  &1.48  &\textbf{0.047}  \\
\orange{Template}  &51.5 &45.1 &19.0 &0.943 &0.509 &111  &1.81  &0.035 \\
\orange{Del\&Retri}  &50.6 &22.1 &11.8 &0.934 &0.345 &94  &1.42  &0.041  \\
\orange{B-GST} &30.3 &22.5 &11.6 &\textbf{0.951} &\textbf{0.557}  &117  &1.34  &0.034\\
\orange{G-GST} &31.0 &20.7 &10.2 &0.941  &0.556  &127  &1.29  &0.031  \\
\orange{UST} & 23.6  &0.5 &0.5 &0.881  &0.012  &\textbf{28}  &0.27  &0.035 \\
\orange{SMAE} &21.6  &6.5 &1.2  &0.898  &0.079  &74  &0.62  &0.046 \\
\hline
\green{DRLST} &71.1 &4.2 &2.7 &0.909  &0.342  &86 & 1.04 & 0.045  \\
\green{BST} &69.7 &0.5 &0.5 &0.883 &0.04 &69  &0.38  &0.042  \\
\green{CAAE} &72.3 &1.8 &1.5 &0.896 &0.028  &55  &0.51  &0.044\\
\green{ARAE} &76.2 &4.8 &2.2 &0.903 &0.042  &77 &0.67  &0.040\\
\green{Ctrl-Gen} &73.1 &57.0 &15.6 &0.943 &0.446  &168  &1.82 &0.023  \\
\green{Multi-Dec} &22.2 &13.4 &5.9 &0.911  &0.168  &146  &0.76  &0.026  \\
\green{Style-Emb} &27.7  &8.3 &3.6 &0.897  &0.102  &136  &0.64  &0.027  \\
\hline
\blue{DualRL}  &56.7 &\textbf{61.6} &18.8 &0.944 &0.447 &122  &\textbf{1.89}  &0.032 \\
\blue{DAST} &73.1  &50.6 &14.3 &0.934  &0.350  &204  &1.59  &0.019 \\
\blue{DAST-C} &78.2  &48.5 &13.8 &0.927  &0.328  &308  &1.42  &0.013  \\
\blue{PFST} &50.8  &55.3 &16.5 &0.940  &0.466  &200  &0.51  &0.020  \\
\hline
Human0 &78.1  &20.5 &43.5 &0.942  &0.393  &80  &1.67  &0.048  \\
Human1 &\textbf{78.7}  &18.2 &43.2 &0.931  &0.342  &199  &1.25  &0.020  \\
Human2 &78.2  &18.6 &43.4 &0.932  &0.354  &192  &1.28  &0.021 \\
Human3 &77.4  &18.8 &43.5 &0.931  &0.354  &196  &1.27  &0.020 \\
\hline
\end{tabular}
\end{table}

Table \ref{tbl:resultGYAFC} presents the performance of various TST models on the formality transfer task. Consistent with observations from the sentiment transfer task, no TST model excels across all evaluation metrics. The average style transfer accuracy for the GYAFC dataset is 52.9\%, which is markedly lower than the average score of 84.4\% observed for the Yelp dataset, underscoring the challenges inherent in the formality transfer task. Additionally, most models performed less effectively on this task compared to the sentiment transfer task. It is unsurprising that keyword replacement methods did not fare well in the formality transfer task, with most achieving low style transfer accuracy. Some adversarial learning-based TST models, such as \green{CAAE} \cite{shen2017style} and \green{DRLST} \cite{john2019disentangled}, attained high style transfer accuracy but demonstrated very low content preservation due to the absence of mechanisms to maintain content integrity during the generative process. Interestingly, attribute-controlled TST methods such as \green{Ctrl-Gen} \cite{hu2017toward}, \blue{DAST} \cite{li2019domain}, and \green{DAST-C} \cite{li2019domain} exhibited strong performance in both style transfer accuracy and content preservation. Consistent with sentiment transfer observations, implicit style-content disentanglement methods—namely, \green{DRLST}, \green{BST}, \green{CAAE}, \green{ARAE}, and \green{Multi-Dec}—performed poorly in content preservation for the formality transfer task, with content preservation scores even lower than those in the sentiment transfer task. Methods that undertake TST tasks without style-content disentanglement achieved the highest performance in the formality transfer task.

The GYAFC dataset includes performance data for four human annotators tasked with executing formality transfer on the test set, as detailed at the bottom of Table \ref{tbl:resultGYAFC}. On average, these human references achieved a style transfer accuracy of 78.1\%. This is deemed a strong performance, especially considering that the pre-trained binary classifier achieved only 83.4\% accuracy on the same test set. Additionally, it is important to note that text formality is a subjective concept, and the four human references may hold differing views regarding the degree of formality required in the text.

Given that the GYAFC dataset is a parallel dataset, meaning that it contains corresponding sentences in both source and target styles, we are able to calculate the \textit{hBLEU} score between the sentence in the transferred style and its corresponding sentence in the target style. It comes as no surprise that human annotators have achieved the highest \textit{hBLEU} scores, indicating that the sentences they produce closely resemble the corresponding target style sentences. In contrast, TST models exhibit poor performance on \textit{hBLEU} scores. Moreover, it is observed that the average content preservation metric scores for TST models are lower in the formality transfer task than in the sentiment transfer task. For instance, the \textit{WO} scores are higher in sentiment transfer because the style transfer can be achieved by replacing only a few keywords. In the formality transfer task, however, the text requires more significant and intricate modifications to achieve style transfer. Consequently, there is less word overlap between the original and the transferred sentence, resulting in lower \textit{WO} scores. This limitation of existing metrics in evaluating content preservation during formality transfer underscores the necessity for developing improved evaluation methods for this challenging task.

\section{Summary}
Despite being a relatively new branch within the field of natural language processing, TST has recently been the subject of substantial research activity. This rapid expansion in TST research has led to the development of numerous novel and intriguing TST models. This survey endeavors to organize these innovative models through a taxonomy (see Fig. \ref{fig:taxonomy}) and provides a summary of the common techniques employed by contemporary TST models for transferring text styles. Additionally, it highlights key research trends within TST, such as the transition from models that attempt to disentangle text style from content to those that aim to perform TST without any style-content disentanglement. While we anticipate that the trend towards performing TST without such disentanglement will persist, we assert that exploring style representation remains an intriguing research avenue worthy of further investigation.

Beyond discussing common TST techniques, this survey includes a comprehensive reproducibility study, where 19 state-of-the-art TST algorithms were replicated and benchmarked using two publicly available datasets. To the best of our knowledge, this constitutes the first large-scale reproducibility study focusing on TST methods. The study's findings reveal that no single TST method excels across all evaluation metrics, underscoring the complexity of the TST task. Different methods exhibit strengths in various aspects, preventing the straightforward declaration of a superior approach. Our evaluation analysis further advocates for the development of improved evaluation metrics for TST.

We anticipate that TST research will continue to thrive and that the industry will discover more exciting applications for existing TST methods. It is our hope that this survey offers readers a comprehensive understanding of the essential aspects of this field, elucidates the primary types of TST methods, and illuminates potential directions for future research.

\chapter{LLM-Adapters: An Adapter Family for Parameter-Efficient Fine-Tuning of Large Language Models} 

\label{llm_adapters} 

\section{Objective}

Large language models (LLMs), such as ChatGPT~\cite{openai-chatgpt-2022} and GPT-4~\cite{openai-gpt4-2023}, have demonstrated unprecedented performance across various natural language processing (NLP) tasks~\cite{qin-chatgpt-2023} and multi-modal tasks~\cite{hugginggpt}. These LLMs often possess sizes exceeding hundreds of billions of parameters and are closed-source. Consequently, this has spurred the development of accessible and cost-effective alternatives such as LLaMA~\cite{llama}.
These alternatives involve fine-tuning open-source LLMs utilizing either task-specific data (e.g., ChatDoctor~\cite{yunxiang2023chatdoctor}) or instructional data (e.g., Alpaca~\cite{alpaca}). However, full-model fine-tuning (FFT) is computationally and storage-intensive, thereby presenting significant challenges in practical implementation.

Prior to the emergence of FFT of LLMs (e.g., LLaMA), a compelling solution called parameter-efficient fine-tuning (PEFT)~\cite{adapters} has been proposed in the NLP field, specifically for pre-trained models (e.g., BERT~\cite{devlin2018bert}), offering a promising approach for efficiently fine-tuning LLMs. 
The advantage of PEFT lies in its ability to fine-tune only a small set of external parameters rather than the entire backbone model while still achieving comparable or even superior performance~\cite{peft}.
Moreover, PEFT can effectively mitigate catastrophic forgetting in comparison to FFT~\cite{adamix}.
As shown in Table~\ref{tab:peft_category}, the advantage of PEFT has resulted in the developing of diverse PEFT modules, encompassing series adapters~\cite{adapters, adamix, sparseadapter, lets}, parallel adapters~\cite{parallel_adapter}, reparameterization-based methods~\cite{lora, krona}, and prompt-based learning methods~\cite{prompt_tuning, prefix}.

By incorporating these PEFT modules into backbone models (i.e., LLMs), we can capitalize on the remarkable capabilities of backbone models without requiring extensive computational resources. 
This opens up opportunities for a broader range of applications, enabling even those with limited access to high-performance computing to harness the power of LLMs in their specific tasks.
Despite the success of PEFT for pre-trained models, it remains unclear which PEFT module, in combination with which layer and hyperparameter configuration, is most suitable for a given task or dataset when meeting LLMs (e.g., LLaMA~\cite{llama}). Therefore, further investigation is needed to determine the optimal PEFT setup that maximizes performance across different tasks and datasets.

Motivated by this, in this chapter, we conduct a comprehensive empirical study of PEFT of three representative open-source LLMs, including BLOOM~\cite{bloom}, GPT-J~\cite{gpt-j}, and LLaMA~\cite{llama}.
Specifically, we undertake an empirical study to address the following three research questions: ($i$) What is the optimal placement and configuration of different PEFT methods? ($ii$) How's the performance of different adapters across downstream tasks? And ($iii$) What are the differences in performance between in-distribution (ID) and out-of-distribution (OOD) scenarios for PEFT methods? 
The findings of our study are as follows: 
\begin{enumerate}
    \item \textbf{The optimal placement for the series adapter, parallel adapter, and LoRA is after the MLP layers, parallel with the MLP layers, and located after both the Attention layers and MLP layers simultaneously, respectively};
    \item \textbf{Smaller language models with the PEFT approach can attain competitive or superior performance on specific tasks compared to larger language models.
    For instance, LLaMA-13B with LoRA can outperform GPT-3.5 (>175B) on MultiArith, AddSub, and SingleEq
    }; 
    
    \item \textbf{The ID fine-tuned LLaMA-13B with adapters outperforms ChatGPT on commonsense reasoning tasks indicating that smaller language models have the potential to outperform larger language models on specific tasks with ID fine-tuning data.} 
    
\end{enumerate}

Our contributions can be summarized as follows:
\begin{itemize}
    \item We conduct a comprehensive empirical study of various PEFT methods applied in different open-source LLMs. 
    \item To facilitate our empirical study, we construct two high-quality training datasets to enhance PEFT performance in math reasoning and commonsense reasoning tasks. 
    
    \item We develop a user-friendly framework, LLM-Adapter, seamlessly integrates diverse adapters into LLMs, empowering researchers to implement adapter-based PEFT methods for a wide range of tasks. 
    
    \item We conduct extensive experiments to answer the three research questions to serve as inspiration for future research. 
\end{itemize}

It is imperative to emphasize that we propose the LLM-Adapters framework as a foundational work to facilitate the exploration of parameter-efficient fine-tuning methods for text style transfer.

\section{Experiment Setup}
\subsection{Benchmarks}

\begin{table}[htb]
\centering
\setlength{\tabcolsep}{3pt}
\caption{Details of datasets being evaluated. Math: arithmetic reasoning. CS: commonsense reasoning.}
\begin{tabular}
{lcccr}
\hline
Dataset & Domain & \# train & \# test &Answer  \\\hline
{MultiArith} & Math   & -    &600      &Number \\
{AddSub}     & Math   & -    &395      &Number \\
{GSM8K}      & Math   & 8.8K &1,319    &Number \\
{AQuA}       & Math   & 100K &254      &Option \\
{SingleEq}   & Math   & -    &508      &Number \\
{SVAMP}      & Math   & -    &1,000    &Number \\
{BoolQ}      & CS     & 9.4K &3,270    &Yes/No \\
{PIQA}       & CS     & 16.1K&1,830    &Option \\
{SIQA}       & CS     & 33.4K&1,954    &Option \\
{HellaSwag}  & CS     & 39.9K&10,042   &Option \\
{WinoGrande} & CS     & 63.2K&1,267    &Option \\
{ARC-e}      & CS     & 1.1K &2,376    &Option \\
{ARC-c}      & CS     & 2.3K &1,172    &Option \\
{OBQA}       & CS     & 5.0K &500      &Option \\
\hline
\end{tabular}
\label{tab:dataset_description}
\end{table}
We conduct extensive empirical studies on fourteen benchmark datasets from two categories of reasoning problems: 
\textbf{Arithmetic Reasoning:} (1) the {GSM8K}~\citep{gsm8k} dataset consists of high quality linguistically diverse grade school math word problems created by human problem writers, (2) the {SVAMP}~\citep{svamp} benchmark consists of one-unknown arithmetic word problems for up-to-4 grade level students by making simple changes to a set of problems from another existing dataset, (3) the {MultiArith}~\citep{mutli_arith} dataset of math word problems requiring multiple reasoning steps and operations, (4) the {AddSub}~\cite{addsub} dataset of addition and subtraction arithmetic word problems, (5) the {AQuA}~\citep{aqua} dataset of algebraic word problems with natural language rationales, and (6) the {SingleEq}~\citep{singleeq} dataset of grade-school algebra word problems that map to single equations with varying length; 
\textbf{Commonsense Reasoning:} (1) the BoolQ \cite{boolq} dataset is a question-answering dataset for yes/no questions containing 15942 examples. These questions are naturally occurring and generated in unprompted and unconstrained settings, (2) the PIQA \cite{piqa} dataset of questions with two solutions requiring physical commonsense to answer, (3) the SIQA \cite{siqa} focuses on reasoning about people's actions and their social implications, (4) the HellaSwag dataset of commonsense NLI questions including a context and several endings which complete the context, (5) the WinoGrande \cite{winogrande} dataset is formulated as a fill-in-a-blank task with binary options, and the goal is to choose the right option for a given sentence which requires commonsense reasoning, (6) the ARC-c and (7) the ARC-e are the Challenge Set and Easy Set of ARC \cite{arc} dataset of genuine grade-school level, multiple-choice science questions, and (8) the OBQA dataset contains questions requiring multi-step reasoning, use of additional common and commonsense knowledge, and rich text comprehension. Table \ref{tab:dataset_description} shows the dataset statistics.

\subsection{Fine-tuning Data Collection}
In order to perform fine-tuning on adapters, we acquire two high-quality training datasets specifically designed for math reasoning and commonsense reasoning. Table \ref{tab:dataset_description} reveals that only GSM8K and AQuA datasets provide training sets for arithmetic reasoning. To enhance the diversity of our data, we incorporate the training sets from GSM8K, MAWPS, MAWPS-single \cite{mawps}, and select 1000 examples from AQuA for the purpose of collecting the fine-tuning data. However, it is worth noting that the chosen datasets solely offer equations and corresponding answers. In order to augment the reasoning capabilities of our model, particularly in terms of providing step-by-step rationales, we leverage ChatGPT as the teacher model. By utilizing zero-shot chain-of-thought prompts, ChatGPT generates reasoning steps. We have included the specific prompt templates used to collect the math reasoning dataset in Appendix \ref{sec:math_templates}. To ensure the quality of the data, we eliminate samples that contain incorrect answers. As a result, we obtain a set of 10K math reasoning samples, referred to as Math10K, which we consider for further analysis and fine-tuning. 

To facilitate fine-tuning in the domain of commonsense reasoning, we construct fine-tuning data by formatting the training sets from BoolQ, PIQA, SIQA, HellaSwag, WinoGrande, ARC-e, ARC-c, and OBQA with pre-defined templates. As each dataset in the commonsense reasoning domain entails distinct tasks, we adopt a structured template by initially describing the task's goal, followed by the corresponding content and answer. The template utilized for creating the fine-tuning data can be found in \ref{sec:commonsense_templates}. Upon completion of this process, we obtain a collection of 170K commonsense reasoning samples, which we refer to as Commonsense170K. These datasets will be made publicly available to encourage further research and exploration in this area. 

\subsection{Implementations}
To facilitate the seamless utilization of PEFT methods in both research and practical applications, we have developed a user-friendly framework, LLM-Adapter. LLM-Adapters seamlessly integrates diverse adapters into LLMs, empowering researchers to implement adapter-based PEFT methods for a wide range of tasks. We utilize LLaMA (7B, 13B) \cite{llama}, BLOOMz (7B) \cite{bloom}, and GPT-J (6B) \cite{gpt-j} as the base models for our experiments. As for the four categories of PEFT methods, we select Prefix-Tuning \cite{prefix}, Series Adapter \cite{adapters}, LoRA \cite{lora}, and Parallel adapter \cite{parallel_adapter} as representative candidates to examine their efficacy. For consistency across all fine-tuning experiments, we maintain a batch size of 16. The learning rate for Prefix-Tuning is set to 3e-2, while the rest of the methods adopt a learning rate of 3e-4. Each of the PEFT methods is fine-tuned for three epochs on the fine-tuning datasets. It is important to note that we fine-tune a single model for either the math or commonsense reasoning task, and subsequently evaluate its performance across all corresponding datasets.

\section{Experiment Results}
\subsection{Placement and Configuration}
\begin{figure}[htb] 
	\centering
	\includegraphics[scale = 0.5]{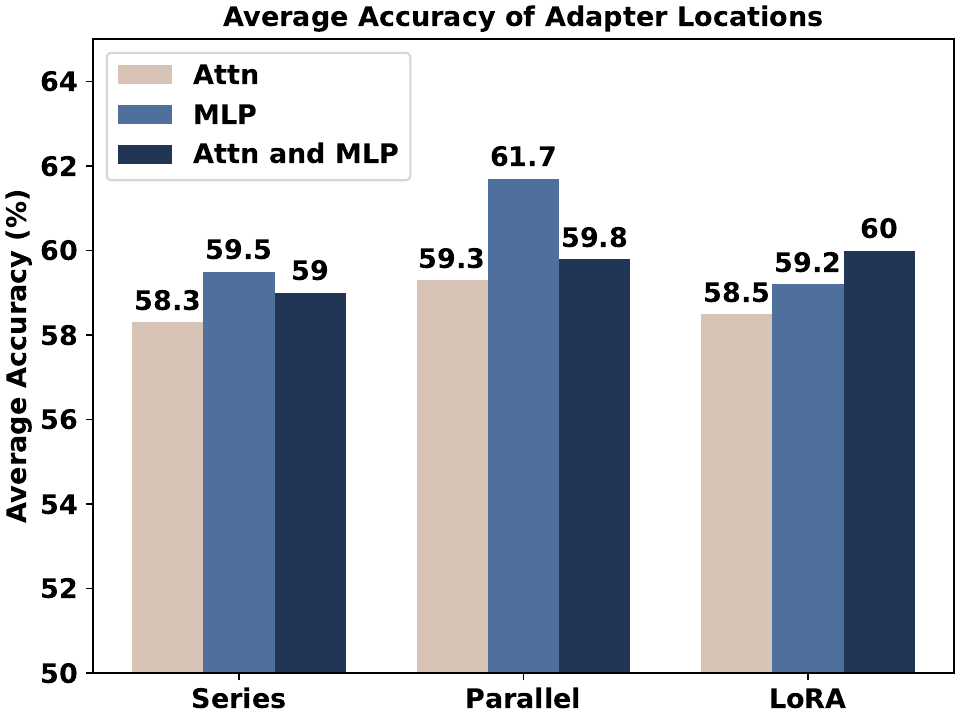} 
	\caption{The average accuracy of different adapter locations on math reasoning datasets.}
        \label{fig:placement}
\end{figure}

To address the research question, \textit{``What is the optimal placement and configuration for various types of adapters?''}, we employ LLaMA-7B as the base model to assess different adapter settings within the context of the math reasoning task. Our empirical study begins by determining the most effective placement for the Series Adapter, Parallel Adapter, and LoRA. Prefix-Tuning is excluded from this analysis since its placement is predetermined. For the Series Adapter, we explore its placement options after the multi-head attention layers, MLP layers, or both of them. As for the Parallel Adapter and LoRA, we integrate them into the multi-head attention layers, MLP layers, or both of them, in order to assess their respective performances. The detailed results on each dataset are shown in Appendix \ref{placement_analysis}.  Figure \ref{fig:placement} shows the average accuracy on math reasoning datasets. We can observe that for the Series Adapter, the best position is to place it after the MLP layers, achieving an average accuracy of $59.5\%$ on the math reasoning datasets. As for the Parallel Adapter, when we place it within the MLP layers, it achieves the best performance of $61.7\%$. Regarding LoRA, we need to insert it simultaneously into both the Multi-head Attention layers and MLP layers to achieve the best performance of $60\%$. 

\begin{figure}[htb] 
    \centering
    \begin{tabular}{cc}
	\includegraphics[scale = 0.35]{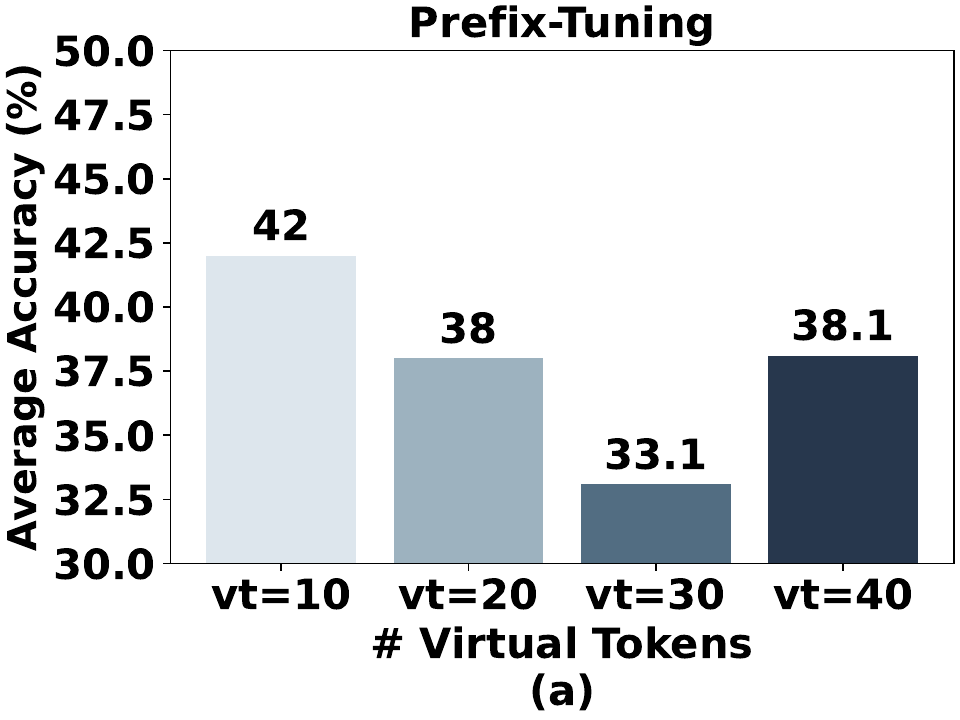} &
        \includegraphics[scale = 0.35]{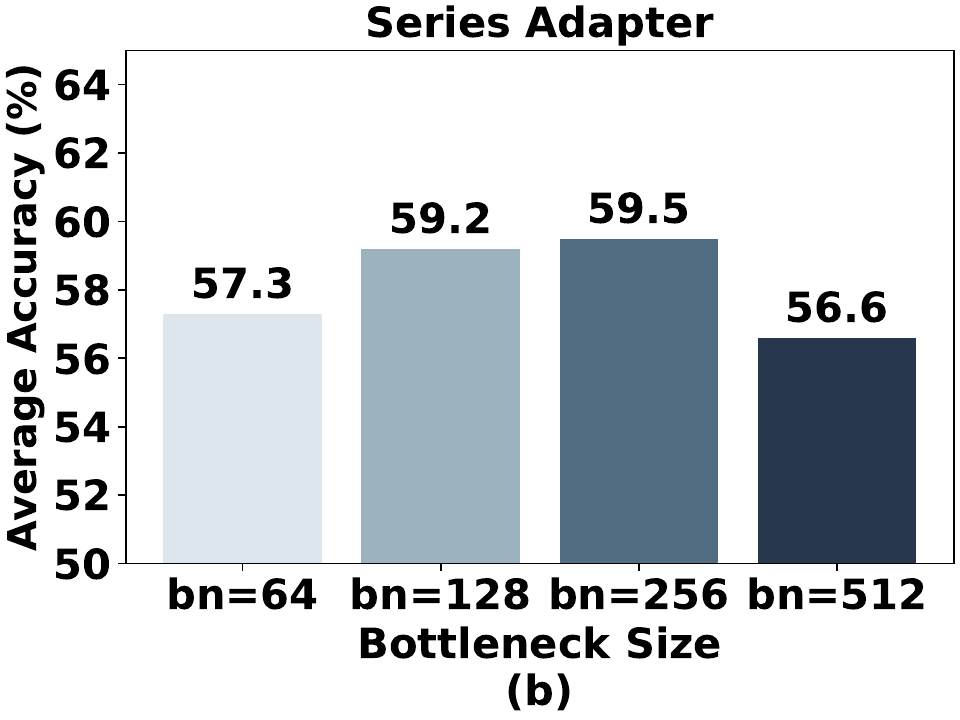} \\
        \includegraphics[scale = 0.35]{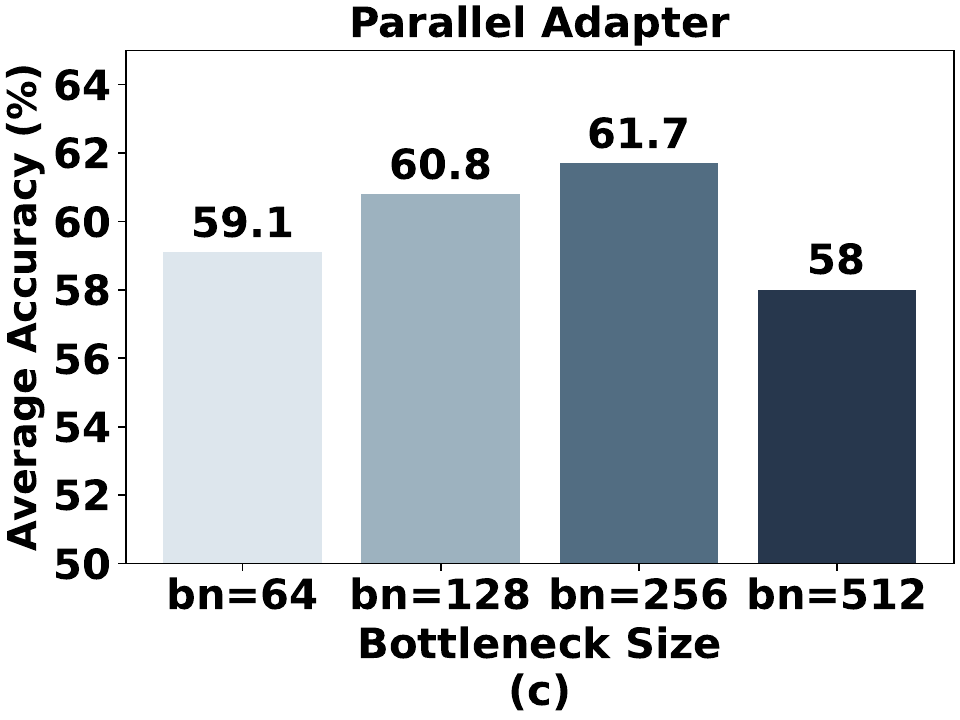} &
        \includegraphics[scale = 0.35]{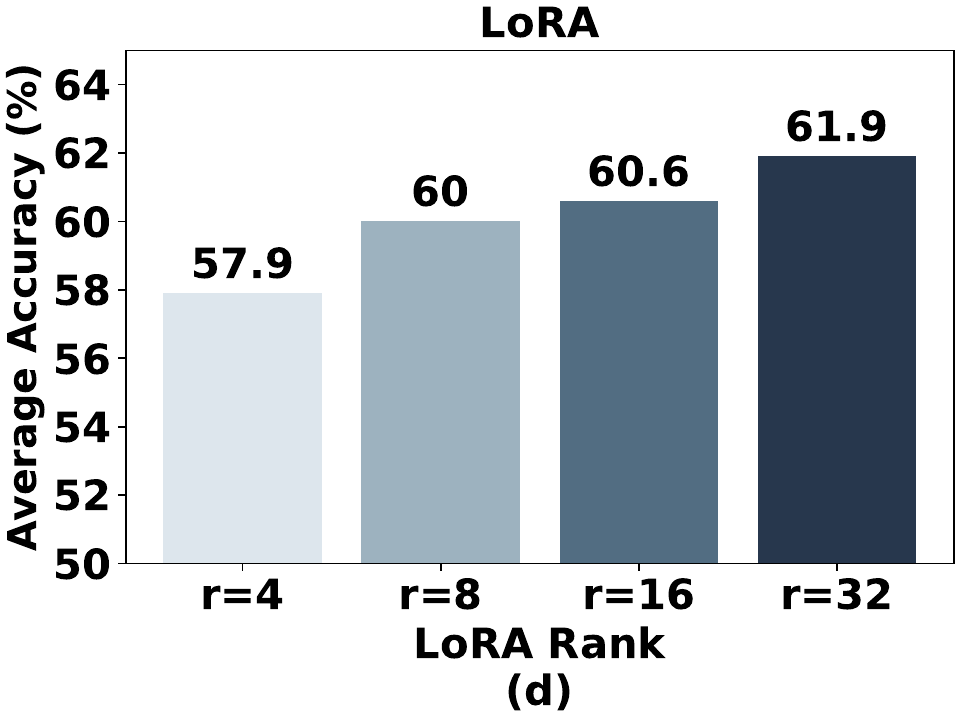} \\
    \end{tabular}
    \caption{The average accuracy of different variable settings on math reasoning datasets. Where "vt" refers to the number of virtual tokens, "bn" denotes the bottleneck size, while "r" is the LoRA rank. }
    \label{fig:configuration}
\end{figure}

In order to determine the optimal configuration of various adapters, we conduct an analysis of the most crucial variable for each type of the PEFT methods. We compare the average accuracy on math reasoning datasets. The placement of adapters follows the optimal settings derived from the placement analysis. Regarding Prefix-tuning, we assess the performance with different numbers of virtual tokens ($vt$) set at $[10, 20, 30, 40]$. For Series and Parallel Adapters, we evaluate the impact of the bottleneck size ($bn$) with values of $[64, 128, 256, 512]$. For LoRA, we examine the influence of different rank values ($r$) at $[4, 8, 16, 32]$. The detailed results for each dataset can be found in Appendix \ref{configuration_analysis}. Figure \ref{fig:configuration} presents the average accuracy of different variables on math reasoning datasets. It can be noted that when the number of virtual tokens in Prefix-Tuning is set to 10, Prefix-Tuning attains an average accuracy of $42.0\%$ on math reasoning datasets. By configuring the bottleneck dimension to 256, Series and Parallel Adapter demonstrate the highest level of performance. However, when the bottleneck size is increased to 512, the accuracy of both Series and Parallel Adapter decreases. The typical setting for LoRA rank is set to 8, but we have discovered that a larger rank can enhance the performance of LoRA. When the rank is increased from 8 to 32, the average accuracy of LoRA increases from $60.0\%$ $61.9\%$. 

In order to enhance the breadth of our research findings, we conducted additional experiments involving the placement of adapters on various LLMs such as GPT-J and BLOOMz. These experiments were conducted across different model sizes, specifically 7B and 13B parameters. Furthermore, we extended our investigation to encompass diverse tasks, including Commonsense tasks. This approach enabled us to generalize our observations across a wider spectrum of LLMs, sizes, and tasks, thus providing a more comprehensive understanding of the adapter placement strategies. The detailed experiment results can be found in Appendix \ref{placement_analysis}

Based on our comprehensive placement and configuration analysis, we have determined the optimal settings for each adapter, which will be consistently employed throughout the subsequent experiments. 
\begin{itemize}
    \item \textbf{For Prefix-Tuning, we establish the number of virtual tokens at 10.
    \item For Series and Parallel Adapter, we seamlessly incorporate them into the MLP layers, configuring the bottleneck size to 256.
    \item Regarding LoRA, we seamlessly integrate it into both the Multi-head Attention layers and the MLP layers with rank 32.}
\end{itemize}

\subsection{Arithmetic Reasoning}
\begin{table*}[htb]
\caption{Accuracy comparison of LLMs with different adapters on six math reasoning datasets. We use GPT-3.5 \texttt{text-Davinci-003} for Zero-shot CoT \cite{kojima2022large} as the baseline. 
}
\centering
\setlength{\tabcolsep}{3pt}
\small
\begin{tabular}{llccccccc}
\hline
LLM & Method & MultiArith  &GSM8K &AddSub &AQuA &SingleEq &SVAMP & Avg \\\hline

GPT-3.5$_\text{175B}$ & - &  {83.8}  & \textbf{56.4}& {85.3}& \textbf{38.9} & {88.1} & \textbf{69.9} & \textbf{70.4}\\
\hline

\multirow{4}{*}{BLOOMz$_\text{7B}$}& Prefix & 68.8&	13.8&	47.1&	12.5&	49.4&	24.1&	36.0 \\
& Series  &80.7&	14.3&	72.6&	20.5&	69.3&	38.1&	49.3 \\
& Parallel & 85.8&	18.5&	77.7&	18.9&	74.8&	36.4&	52.0\\
& LoRA &82.8&	17.4&	72.4&	21.3&	69.9&	41.0&	50.8 \\

\hline

\multirow{4}{*}{GPT-J$_\text{6B}$}& Prefix &  74.5&	16.0&	65.6&	14.7&	61.4&	31.0&	43.9\\
& Series  & 91.7&	19.5&	85.8&	15.0&	81.7&	43.6&	56.2\\
& Parallel &92.2&	18.9&	83.8&	17.9&	80.7&	41.1&	55.8 \\
& LoRA &90.7&	23.0&	84.1&	16.1&	84.1&	46.0&	57.3 \\

\hline

\multirow{4}{*}{LLaMA$_\text{7B}$}& Prefix &63.2&	24.4&	57.0&	14.2&	55.3&	38.1&	42.0  \\
& Series  &92.8&	33.3&	80.0&	15.0&	83.5&	52.3&	59.5 \\
& Parallel &94.5&	35.3&	86.6&	18.1&	86.0&	49.6&	61.7 \\
& LoRA &\textbf{95.0}&	37.5&	83.3&	18.9&	84.4&	52.1&	61.9 \\
\hline

\multirow{4}{*}{LLaMA$_\text{13B}$}& Prefix &72.2&	31.1&	56.0&	15.7&	62.8&	41.4&	46.5  \\
& Series  &93.0&	44.0&	80.5&	22.0&	87.6&	50.8&	63.0 \\
& Parallel &94.3&	43.3&	83.0&	20.5&	89.6&	55.7&	64.4\\
& LoRA &94.8&	47.5&	\textbf{87.3}&	18.5&	\textbf{89.8}&	54.6&	65.4 \\

\hline
\end{tabular}

\label{tab:math_results}

\end{table*}

In order to evaluate the effectiveness of adapters on the Arithmetic Reasoning task, we conducted a study where adapters are fine-tuned on the Math10K dataset and subsequently evaluated on six different math reasoning datasets. As our baseline, we utilize the GPT-3.5 model, specifically the \texttt{text-Davinci-003} variant, for Zero-shot CoT according to \citet{kojima2022large}. The results of the GPT-3.5 model can be found in \citet{wang2023plan}. Table \ref{tab:math_results} reports the performance of different PEFT methods and the baseline. On average, the GPT-3.5 model (175B) outperforms adapter-based PEFT LLMs in terms of accuracy. However, for simpler math reasoning datasets such as MultiArith, AddSub, and SingleEq, adapter-based methods like LLaMA-13B with LoRA outperform GPT-3.5. Notably, LLaMA-13B with LoRA achieves an average accuracy of 65.4\%, which is approximately 92.8\% of the performance exhibited by GPT-3.5. This suggests that with sufficient task-specific training data, adapter-based PEFT of smaller LLMs has the potential to achieve performance comparable to that of extremely large language models. The utilization of adapter-based PEFT yields superior performance by smaller language models compared to GPT-3.5 specifically in simpler tasks such as MultiArith, AddSub, and SingleEq. However, challenges persist in more complex tasks like GSM8K and SVAMP, which require a higher level of language comprehension and proficiency from the underlying base model, thereby resulting in a discernible performance gap.
Regarding the different adapters employed, LoRA achieves remarkable performance while utilizing significantly fewer trainable parameters. This implies that excessive learnable parameters may not be necessary for task-specific fine-tuning. Overall, these findings demonstrate the potential for adapter-based PEFT of smaller LLMs to achieve high performance on specific tasks with few trainable parameters.

\subsection{Commonsense Reasoning}

\begin{table*}[htb]
\caption{Accuracy comparison of LLMs with different adapters on eight
commonsense reasoning datasets. The ChatGPT results are obtained by Zero-shot CoT with \texttt{gpt-3.5-turbo} API. 
}
\centering
\setlength{\tabcolsep}{3pt}
\small
\begin{tabular}{llccccccccc}
\hline
LLM & Method & BoolQ  &PIQA &SIQA &HellaSwag &WinoGrande &ARC-e &ARC-c &OBQA &Avg \\\hline

GPT-3$_\text{175B}$ &-& 60.5&	81.0&	-&	78.9&	70.2&	68.8&	51.4&	57.6& -\\
PaLM$_\text{540B}$ &-& \textbf{88.0}&	82.3&	-	&83.4&	81.1&	76.6&	53.0&	53.4& - \\
ChatGPT & - & 73.1&	\textbf{85.4}&	68.5&	78.5&	66.1&	\textbf{89.8}&	\textbf{79.9}&	74.8&	77.0 \\
\hline

\multirow{4}{*}{BLOOMz$_\text{7B}$}& Prefix &45.6&	53.7&	46.3&	26.7&	49.5&	52.1&	39.7&	44.3&	44.7  \\
& Series  &65.4&	70.4&	73.6&	53.4&	69.3&	72.3&	55.9&	68.0&	66.0 \\
& Parallel &64.1&	71.5&	72.1&	52.9&	67.0&	70.5&	54.7&	69.6&	65.3 \\
& LoRA &65.9&	75.3&	74.5&	57.3&	72.5&	74.6&	57.8&	73.4&	68.9 \\

\hline

\multirow{4}{*}{GPT-J$_\text{6B}$}& Prefix & 63.1&	66.9&	68.7&	34.4&	64.5&	64.4&	46.8&	59.0&	58.5 \\
& Series  &62.1&	63.5&	72.3&	30.6&	68.0&	63.9&	48.1&	63.8&	59.0 \\
& Parallel &62.2&	69.7&	70.0&	41.7&	65.0&	60.2&	44.6&	58.2&	59.0 \\
& LoRA &62.4&	68.6&	49.5&	43.1&	57.3&	43.4&	31.0&	46.6&	50.2 \\

\hline

\multirow{4}{*}{LLaMA$_\text{7B}$}& Prefix &64.3&	76.8&	73.9&	42.1&	72.1&	72.9&	54.0&	60.6&	64.6  \\
& Series  &63.0&	79.2&	76.3&	67.9&	75.7&	74.5&	57.1&	72.4&	70.8 \\
& Parallel &67.9&	76.4&	78.8&	69.8&	78.9&	73.7&	57.3&	75.2&	72.3 \\
& LoRA &68.9&	80.7&	77.4&	78.1&	78.8&	77.8&	61.3&	74.8&	74.7 \\
\hline

\multirow{4}{*}{LLaMA$_\text{13B}$}& Prefix &65.3&	75.4&	72.1&	55.2&	68.6&	79.5&	62.9&	68.0&	68.4  \\
& Series  &71.8&	83.0&	79.2&	88.1&	82.4&	82.5&	67.3&	81.8&	79.5 \\
& Parallel &72.5&	84.8&	79.8&	\textbf{92.1}&	\textbf{84.7}&	84.2&	71.2&	\textbf{82.4}&	\textbf{81.5} \\
& LoRA &72.1&	83.5&	\textbf{80.5}&	90.5&	83.7&	82.8&	68.3&	\textbf{82.4}&	80.5 \\

\hline
\end{tabular}
\label{tab:commonsense_results}

\end{table*}

Additionally, we assess the efficacy of various PEFT methods for commonsense reasoning tasks. The adapters undergo fine-tuning using the Commonsense170K dataset. Our baseline models for commonsense reasoning include GPT-3 (175B), PaLM (540B), and ChatGPT. The results for GPT-3 and PaLM can be found in the study by \citet{llama}. To evaluate ChatGPT's performance in commonsense reasoning, we employ the \texttt{gpt-3.5-turbo} API with a zero-shot CoT. The zero-shot CoT prompts align with the template used for collecting our commonsense fine-tuning dataset, as outlined in Appendix \ref{sec:commonsense_templates}. Table \ref{tab:commonsense_results} presents the performance of the PEFT methods utilizing different LLMs  alongside the baselines. Remarkably, LLaMA-13B with Series Adapter, Parallel Adapter, and LoRA outperform all the baselines, including ChatGPT, which has been hailed as the most impressive LLM to date. LLaMA-13B with Parallel Adapter achieves an average accuracy of 81.5\%, representing a 4.5\% improvement over ChatGPT. It is worth noting that all the training sets from the commonsense reasoning datasets are included in the fine-tuning data Commonsense170K. Furthermore, we observe that the performance of the PEFT methods is influenced by the underlying capabilities of the base models. LLaMA-7B and LLaMA-13B demonstrate superior commonsense reasoning abilities compared to the BLOOMz and GPT-J models.

\subsection{ID and OOD Analysis}
When comparing the performance of PEFT methods on math reasoning and commonsense reasoning tasks, we can observe that PEFT methods exhibit more remarkable results in the realm of commonsense reasoning. Moving forward, we will analyze the factors contributing to this phenomenon from both the in-distribution (ID) and out-of-distribution (OOD) perspectives. In the context of commonsense reasoning, the fine-tuning data set, Commonsense170K, encompasses all the training sets from the commonsense reasoning datasets. Notably, PEFT methods have demonstrated the ability to outperform ChatGPT. This observation implies that, by utilizing ID fine-tuning data, smaller language models like LLaMA-13B could surpass larger language models such as ChatGPT and PaLM in specific downstream tasks. However, when considering math reasoning tasks, the fine-tuning data set, Math10K, only includes the training sets of GSM8K and AQuA. In this regard, it has been observed that PEFT methods, particularly LLaMA-13B with LoRA, exhibit superior performance compared to GPT-3.5 on MultiArith, AddSub, and SingleEq. These findings suggest that PEFT methods can enhance the math reasoning abilities of LLMs and can be successfully applied to OOD datasets. Nonetheless, when evaluating the performance of PEFT methods on the ID datasets GSM8K and AQuA, a performance gap is still evident compared to GPT-3.5. This discrepancy is likely due to the higher complexity of GSM8K and AQuA datasets in terms of math reasoning, while the reasoning capabilities of smaller LLMs remain limited. Consequently, identifying strategies to improve the performance of PEFT methods on complex math reasoning tasks represents a potential avenue for future research.

\section{Qualitative Study}

\begin{table*}[htb]
\caption{An example randomly sampled from GSM8K. The outputs of ChatGPT and LLaMA-13B with different PEFT methods.}
\centering
\small
\begin{tabular}{lp{11cm}}\hline
Question &  Mr Boarden is remodeling his bathroom. For every square foot, he needs 24 mosaic tiles. How many mosaic tiles would Mr Boarden need to cover two thirds of his 36 sq ft bathroom? \\
\hline
ChatGPT & 
1. Find two-thirds of 36: Multiply 36 by 2/3. \textbf{(36 * 2/3) = 24 square feet}

2.Determine the number of mosaic tiles needed per square foot: Given that Mr. Boarden needs 24 mosaic tiles per square foot, multiply the result from step 1 by 24. \textbf{(24 * 24) = 576 mosaic tiles}

Therefore, Mr. Boarden would need 576 mosaic tiles to cover two-thirds of his 36 square foot bathroom.
\\ \hline

LLaMA$_\text{13B}$-Prefix & 1 square foot = 4 square inches, 1 square inch = 16 square centimeters. So, 1 sq ft = 16 sq cm. Therefore, 36 sq ft = 36 x 16 sq cm = 576 sq cm.
To cover two-thirds of the bathroom, Mr Boarden needs to cover 576 sq cm x 2/3 = 384 sq cm.
Since 1 sq cm = 1 mosaic tile, Mr Boarden needs 384 mosaic tiles.
Therefore, Mr Boarden needs 384 mosaic tiles to cover two-thirds of his 36 sq ft bathroom. \\ \hline

LLaMA$_\text{13B}$-Series & 1. Find the area of the bathroom: 36 sq ft

2. Find two-thirds of the area: \textbf{2/3 x 36 sq ft = 24 sq ft}

3. Multiply the area by the number of tiles needed per square foot: \textbf{24 sq ft x 24 tiles/sq ft = 576 tiles}

4. Therefore, Mr Boarden would need 576 mosaic tiles to cover two-thirds of his 36 sq ft bathroom. \\ \hline

LLaMA$_\text{13B}$-Parallel &  1. Calculate the total area of the bathroom: 36 sq ft

2. Determine how much area Mr Boarden wants to cover with mosaic tiles: \textbf{2/3 x 36 sq ft = 24 sq ft}

3. Calculate the number of mosaic tiles needed: \textbf{24 sq ft x 24} mosaic tiles per sq ft = 600 mosaic tiles

Therefore, Mr Boarden would need 600 mosaic tiles to cover two thirds of his 36 sq ft bathroom.
\\\hline
LLaMA$_\text{13B}$-LoRA &  1. Calculate the total area of the bathroom: 36 sq ft

2. Determine how much area needs to be covered: \textbf{2/3 x 36 sq ft = 24 sq ft}

3. Calculate the number of mosaic tiles needed: \textbf{24 sq ft x 24 mosaic tiles per sq ft = 576 mosaic tiles}

Therefore, Mr Boarden would need 576 mosaic tiles to cover two thirds of his 36 sq ft bathroom.  \\\hline

\hline
\end{tabular}
\label{tab:case_study}
\end{table*}

The previous sections have presented the quantitative analysis. In this section, we will provide qualitative examples to demonstrate the quality of outputs from different models. Table \ref{tab:case_study} displays a randomly selected question from GSM8K along with the outputs of ChatGPT and LLaMA-13B models using various PEFT methods. More detailed examples can be found in Appendix \ref{sec:examples}. ChatGPT demonstrates a comprehensive understanding of the question and generates two steps, "(36 * 2/3) = 24 square feet" and "(24 * 24) = 576 mosaic tiles," effectively solving the problem. However, the language understanding ability of LLaMA-13B-Prefix models is limited, leading LLaMA-13B-Prefix to take the wrong direction in the first step. On the other hand, LLaMA-13B with Series Adapter produces a high-quality answer by providing the crucial two steps and performing the correct calculations to obtain the accurate result. Interestingly, LLaMA-13B-Parallel and LLaMA-13B-LoRA generate almost identical rationales. However, LLaMA-13B-Parallel produces an incorrect answer due to a calculation error, stating "24 sq ft x 24 mosaic tiles per sq ft = 600 mosaic tiles". In general, when equipped with task-specific fine-tuning data, smaller language models like LLaMA-13B can generate impressive, high-quality answers that are comparable to those produced by ChatGPT.

\section{Summary}
In this chapter, we develop a user-friendly framework, LLM-Adapter, seamlessly integrates diverse adapters into LLMs, empowering researchers to implement adapter-based PEFT methods for a wide range of tasks. To evaluate different PEFT methods on downstream tasks, we construct two high-quality fine-tuning datasets to enhance PEFT performance on math reasoning and commonsense reasoning tasks. By utilizing the LLM-Adapter toolkit and the constructed fine-tuning datasets, we conduct a comprehensive empirical study and find the answer of research questions on the optimal placement and configuration of different PEFT methods, the impact of adapter architectures, and the influence of ID and OOD scenarios. We propose the LLM-Adapters framework as a foundational work to facilitate the exploration of parameter-efficient fine-tuning methods for text style transfer.
\chapter{Adapter-TST: A Parameter Efficient Method for Multiple-Attribute Text Style Transfer} 

\label{adapter_tst} 

\section{Objective}

\textbf{Motivation.} Text style transfer (TST) is a popular natural language generation task that aims to change the stylistic properties (e.g., sentiment, formality, tense, voice) of the text while preserving the style-independent content \cite{hu2022text}. Existing studies explore performing text style transfer on attributes like age, or gender \cite{lampleSSDRB19}, sentiment \cite{li2018delete,Luo19DualRL,fu2018style}, formality \cite{rao2018dear}, politeness \cite{madaan-etal-2020-politeness,hu2022current}, and author writing style \cite{syed2020adapting}. Nevertheless, most of the existing TST studies are confined to single-attribute TST tasks.

\begin{figure}[t] 
	\centering
	\includegraphics[scale = 0.16]{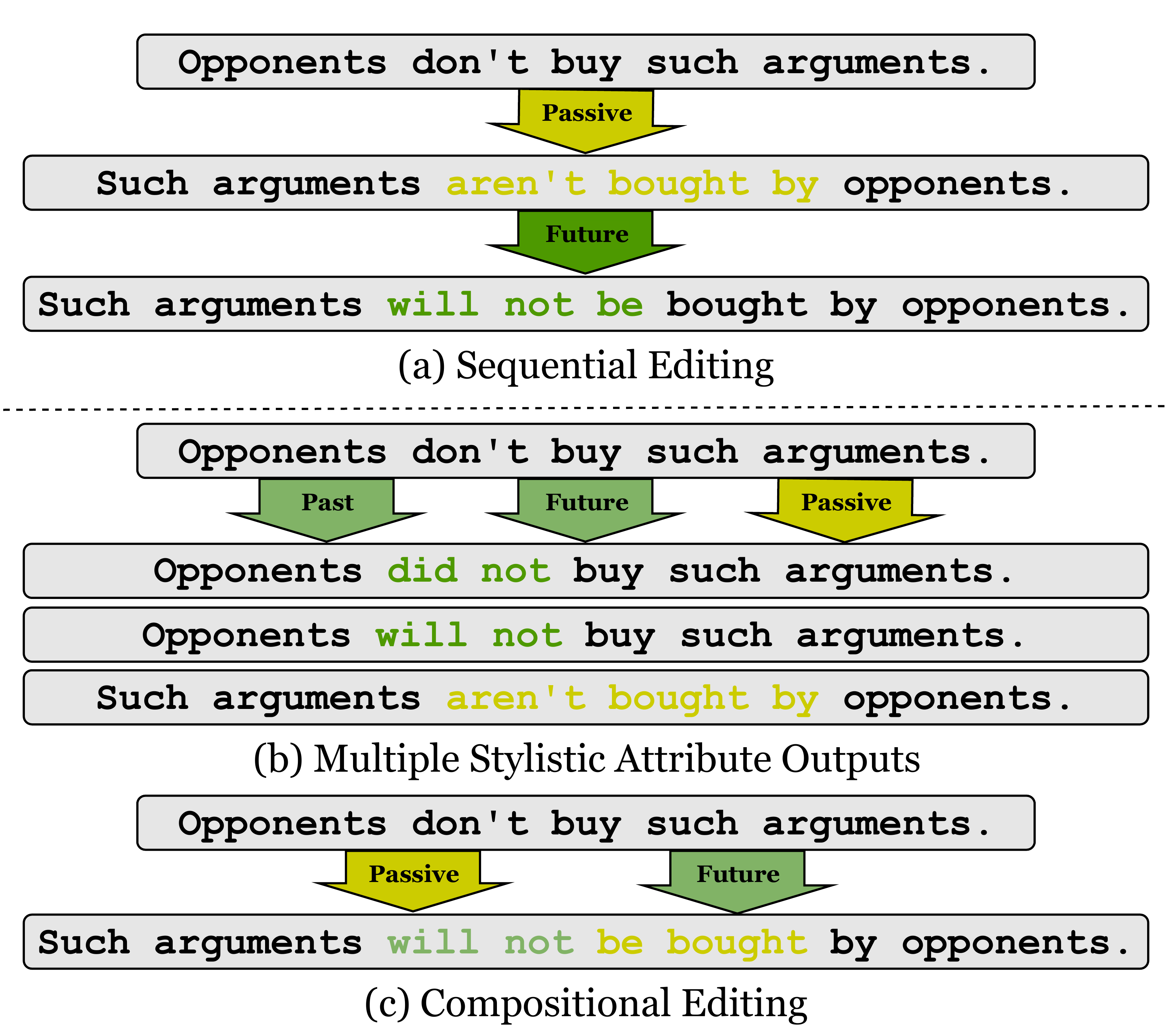} 
	\caption{Examples of different settings of multiple-attribute TST. (a) Existing single-attribute TST models perform sequential editing by transferring the text style sequentially to achieve compositional editing. Multiple-attribute TST models can (b) generate multiple outputs simultaneously in the corresponding target style, or (c) perform compositional editing by transferring different target styles. The proposed \textsf{Adapter-TST} enables a single PLM to achieve both settings (b) and (c) by configuring the adapters' connection method.}
	\label{fig:example}
\end{figure}

Few works have explored multiple-attribute TST tasks, where TST models are designed to control and transfer text in multiple target stylistic attributes. \citet{lampleSSDRB19} attempts style transfer with multiple attributes by conditioning on the average embedding of each target attribute and using a combination of denoising autoencoder (DAE) and back-translation techniques. \citet{goyal-etal-2021-multi} propose an approach to initialize an encoder-decoder setup with a transformer-based language model that is pre-trained on a generic corpus and enhances its capability of re-writing to multiple target style dimensions by utilizing multiple style-aware language models as discriminators. 

A possible approach to perform single and multiple attribute TST tasks is to leverage large pre-trained language models (PLMs). The PLMs have been pre-trained on large corpora, which allows them to capture natural language's syntactic and semantic information. This characteristic of PLMs makes them well-suited for TST tasks, where the model needs to understand the content and style of the input text. \citet{syed2020adapting} fine-tune a denoising autoencoder (DAE) for the stylized re-writing task by initializing the encoder and decoder with a pre-trained language model trained on Masked Language Modeling (MLM) objectives \cite{devlin-etal-2019-bert}. \citet{wang2019harnessing} fine-tune GPT-2 model \cite{radford2019language} using the text formality transfer rules harnessed from analyzing the GYAFC parallel dataset \cite{rao2018dear}. The fine-tuned GPT-2 model was subsequently used to transfer the formality of text (e.g., informal to formal text). However, fine-tuning PLMs for multiple-attribute TST remains challenging as a significant amount of computational resources and style-labeled data are required to perform TST for each stylistic attribute.

\textbf{Research Objectives.} To address these research gaps, we propose \textsf{Adapter-TST}, a parameter-efficient framework that utilizes BART \cite{lewis-etal-2020-bart} or T5 \cite{raffel2020exploring} as the backbone model and trains neural adapters to capture multiple stylistic attributes for multiple-attribute TST. During the training of \textsf{Adapter-TST}, we freeze the original parameters of the pre-trained BART or T5 model and only update the parameters of adapters to relax the dependence on computational resources and supervised data. The proposed \textsf{Adapter-TST} model is flexible to handle different settings of multiple-attribute TST by configuring the connection method among adapters. Figure \ref{fig:example} illustrates the different settings of multiple-attribute TST tasks. Paralleling the adapters in \textsf{Adapter-TST} can generate multiple outputs in the corresponding target style simultaneously (setting b) and stacking the adapters for compositional editing in terms of different target styles at the same time (setting c). We conduct experiments on the traditional sentiment transfer task and multiple-attribute TST tasks, including multiple stylistic attribute outputs and compositional editing. Results of automatic and human evaluations show that \textsf{Adapter-TST} can outperform the state-of-the-art baselines to transfer and generate high-quality text with lower computational resources. 

\textbf{Contributions.} We summarize our contributions as follows: (i) We introduce an \textsf{Adapter-TST}, which is a parameter-efficient framework that can perform multiple-attribute TST tasks with significantly lower computational resources. (ii) Included in the \textsf{Adapter-TST} are two TST configurations, \textit{parallel} and \textit{stacking}, which support  multiple-output TST and compositional editing, respectively. (iii) We conducted extensive experiments on real-world datasets. The automatic and human evaluation results show that \textsf{Adapter-TST} can outperform the state-of-the-art baselines to transfer and generate high-quality text.

\section{Methodology}

This section proposes \textsf{Adapter-TST}, which adds neural adapters into each transformer layer to capture different attribute information for multiple-attribute TST. We first introduce the adapter structure used in \textsf{Adapter-TST} and its parameter efficiency. Subsequently, we explain how the adapters are configured for different multiple-attribute TST settings, namely, \textit{multiple stylistic attribute outputs} and \textit{compositional editing}. Finally, we describe the training objectives of \textsf{Adapter-TST}.

\subsection{Adapter Structure}

\begin{figure}[t] 
	\centering
	\includegraphics[scale = 0.15]{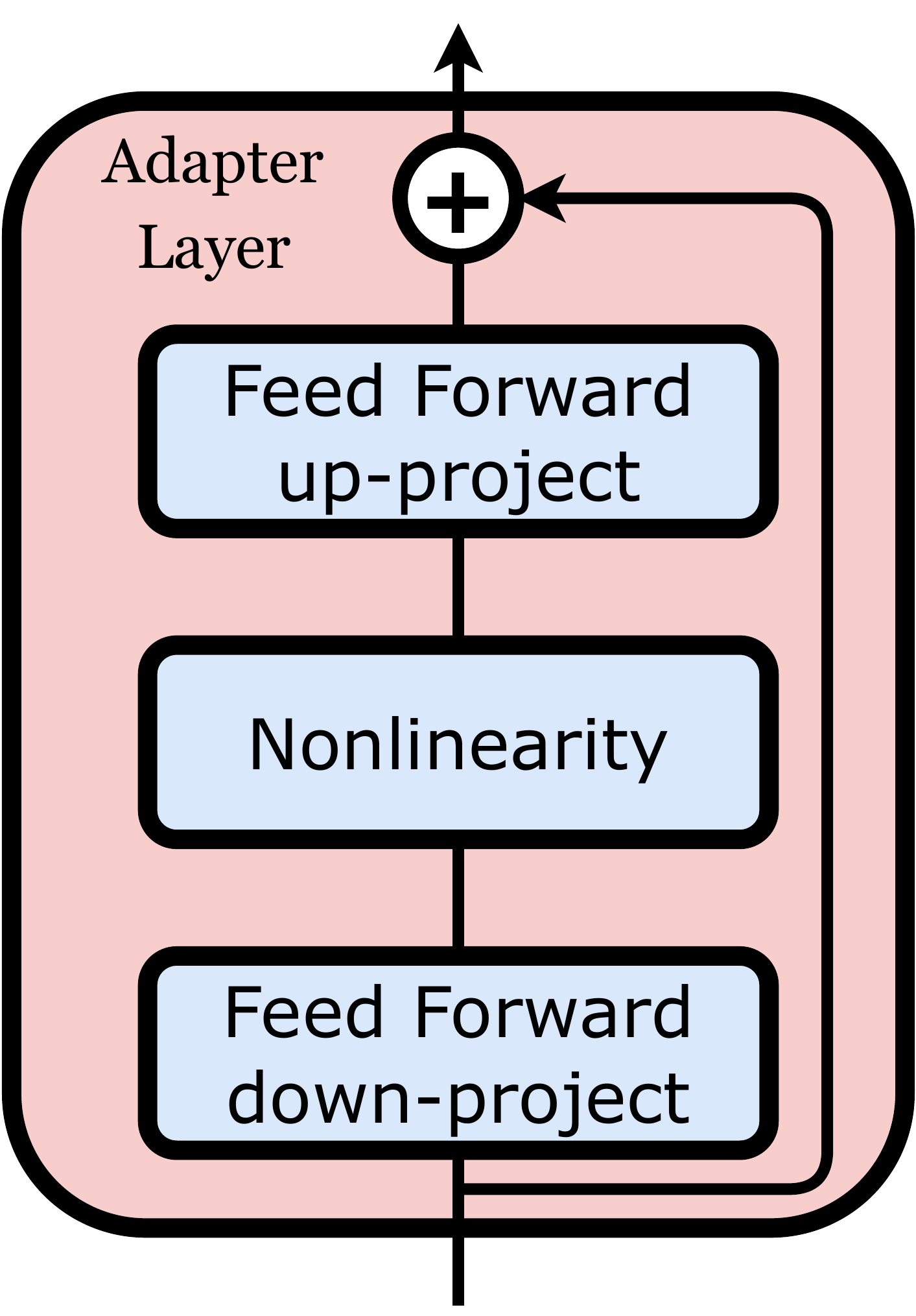} 
	\caption{Structure of the adapter layer. The adapter layer consists of a bottleneck with up and down projection layers, and a skip connection between two projection layer.}
	\label{fig:adapter_layer}
\end{figure}

\begin{figure*}[t] 
	\centering
	\includegraphics[scale = 0.09]{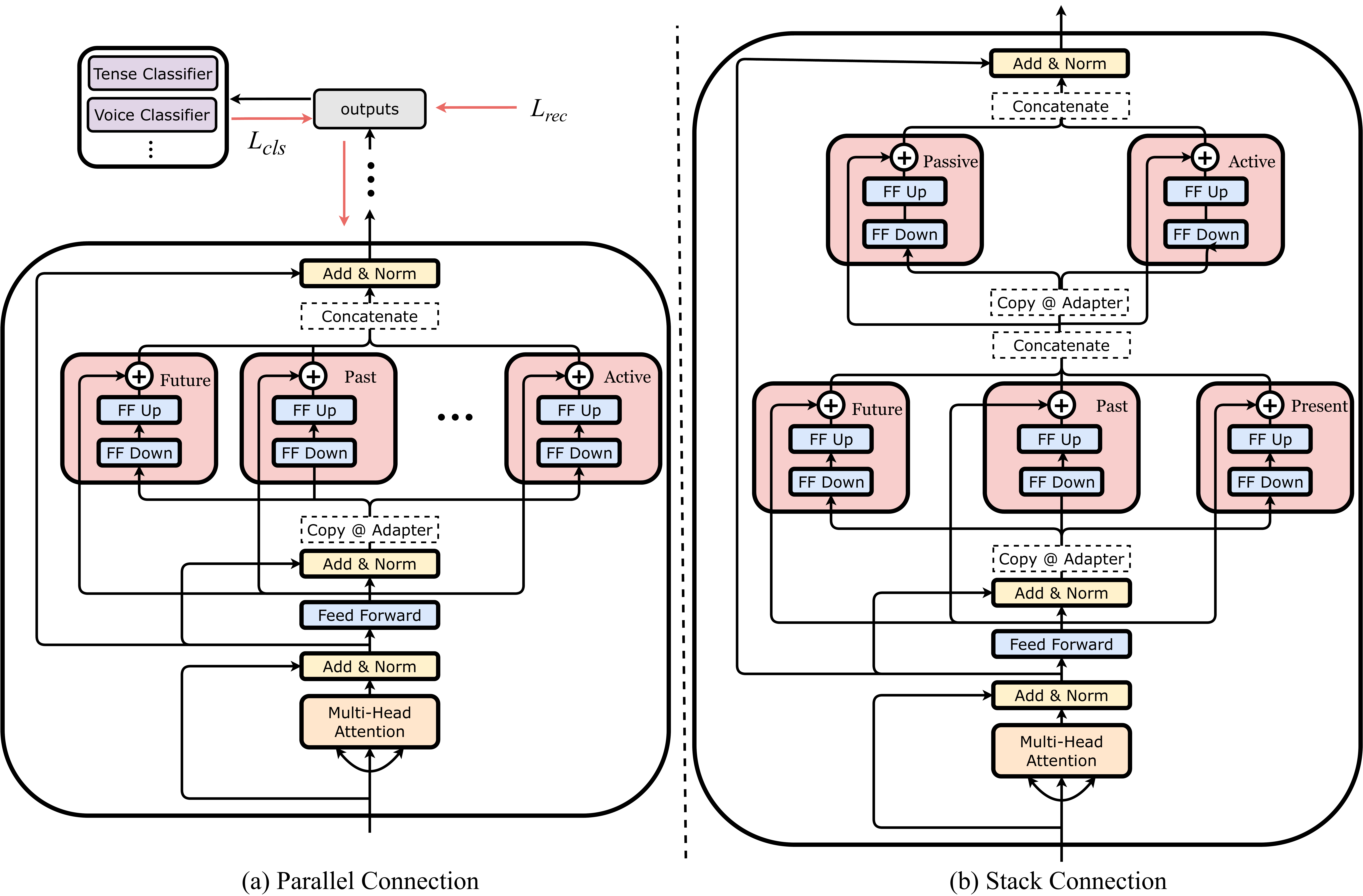} 
	\caption{\textsf{Adapter-TST} Configurations - Left: Paralleling the adapters enables a single PLM to model different attributes simultaneously and generate multiple outputs in the corresponding target style. Right: Stacking the adapters for compositional editing in terms of different target styles at the same time. \textit{Stack} connection is used for inference to verify the relevant attribute information captured by adapters.}
	\label{fig:model}
\end{figure*}

We present an adapter structure in Figure \ref{fig:adapter_layer}. The adapter consists of a bottleneck that contains few parameters relative to the attention and feedforward layers in the original model. A skip connection is applied across two projection layers. In our proposed \textsf{Adapter-TST}, these adapters will be trained to capture different stylistic attributes. In contrast to \citet{houlsby2019parameter}, adding the adapter module twice to each transformer layer, we propose simplifying the approach by just adding the adapter layer into each transformer once, making our \textsf{Adapter-TST}'s architecture more parameter efficient. 

We use BART-large (24-layer, 1024-hidden, 16-heads, 406M parameters) or T5-large (24-layer, 1024-hidden, 16-heads, 770M parameters) as the backbone model in \textsf{Adapter-TST}. As for each adapter layer, we denote the hidden dimensions of the down-projection and up-projection layers as $H_d = 64$ and $H_u = 1024$. The bottleneck adapter layers are plugged into each layer of BART-large or T5-large, and different adapter layers do not share parameters. Thus the total number of parameters for each attribute-specific adapter is about 3.17M, which is only 0.78\% of the original BART-large model and 0.41\% of the T5-large model, making the training process parameter efficient. Note that the original parameters of BART-large or T5-large are frozen during multiple-attribute TST training, and only the parameters of adapters are trainable and initialized randomly. 

\subsection{Adapter-TST Configurations}
\textsf{Adapter-TST} has two configurations, \textit{parallel} and \textit{stack}, which support two multiple-attribute TST task settings: \textit{multiple stylistic attribute outputs} and \textit{compositional editing}, respectively. To better understand the two Configurations of \textsf{Adapter-TST}, we take the multiple-attribute TST task with \textit{tense} and \textit{voice}  attributes as an example. \textit{Tense} has three attribute values (\textit{Future}, \textit{Past}, \textit{Present}), while \textit{Voice} has two attribute values (\textit{Passive}, \textit{Active}). Thus, we add five attribute-specific adapters \textit{Adapter(Future, Past, Present, Passive, Active)} to the base model for all the possible attribute values, respectively. Each adapter is employed to learn to generate sentences with corresponding attribute values while preserving the semantic content of the inputs. 

\textbf{Parallel Connection.} We define the multiple stylistic attribute outputs as follows: given a sentence $x = \{x_1, ..., x_n\}$ with $n$ tokens and $y_{tense}, y_{voice}$ labels, the \textsf{Adapter-TST} model is required to generate multiple outputs with all possible other attribute values at the same time. For instance, as shown in Figure \ref{fig:example}(b), given an input sentence with present tense and active voice, the multiple-attribute TST models need to generate three sentences in the past tense, future tense, and passive voice simultaneously. The multiple stylistic attribute output setting requires TST models to capture all the stylistic attributes and have the capability of performing style transfer among the attribute values. \textsf{Adapter-TST} performs the multiple stylistic attribute output by utilizing the \textit{Parallel} connection configuration shown in Figure \ref{fig:model}(a). Specifically, we plug the paralleled adapters \textit{Parallel(Future, Past, Present, Passive, Active)} into each transformer layer of the base model. During training, each training sample passes all the attribute-specific adapters, but adapters will take different actions according to the attribute values of input sentences. The adapter learns to reconstruct the input sentence for training samples with the same attribute value as an adapter. Conversely, when training samples with different attribute values, the adapter learns to transfer the attribute of the input sentence while preserving the original content. The outputs of all the adapters are concatenated together to the next layer. The replication is only performed once in the first transformer layer. In the latter transformer layers, we distribute the hidden states to corresponding adapters to make sure that the input of an adapter in the current layer is the output of the adapter with the same attribute value in the preceding layer.

\textbf{Stack Connection.} Compositional editing requires TST models to change multiple attributes simultaneously while preserving the original content. For instance, as shown in Figure \ref{fig:example}(c), given an input sentence with present tense and active voice, the multiple-attribute TST models need to generate one sentence both in future tense and passive voice. 
\textsf{Adapter-TST} performs compositional editing by using the \textit{Stack} connection method shown in Figure \ref{fig:model}(b), where adapters belonging to the same attribute are parallelized because a sentence should only contain one attribute value for a specific attribute. Specifically, we have \textit{Parallel(Future, Past, Present)} and \textit{Parallel(Passive, Active)} for tense and voice attributes. The two sets of paralleled adapters are stacked as \textit{Stack(Parallel(Future, Past, Present), Parallel(Passive, Active))} to learn to transfer multiple attributes. Similar to the \textit{Parallel} connection method, the hidden states are replicated according to the number of adapters in the \textit{Parallel} connection module. It's worth noting that, to demonstrate the attribute-specific adapters captured the attribute information, we only use the \textit{Stack} connection method in inference time. During inference, we reload the parameters of adapters trained in multiple stylistic attribute outputs tasks and change the connection among the adapters to \textit{Stack}. 

\subsection{Training Objectives.} 
The TST task aims to transfer the style of inputs while preserving the original semantic content. Thus, we train \textsf{Adapter-TST} with classification loss $L_{cls}$ for style transfer and reconstruction $L_{rec}$ for content preservation. During training, the original parameters of BART-large or T5-large are frozen, and only the parameters of adapters are trainable. 

\textbf{Classification Loss $L_{cls}$}: The classification loss ensures that the transferred sentence conforms to the target attribute value. To this end, we first pre-train a TextCNN-based \cite{kim2014convolutional} binary attribute classifier $D$ for each attribute, then apply the pre-trained attribute classifiers to guide the updates of adapters' parameters such that the output sentence is predicted to be in the target style: 

\begin{equation}
    L_{cls} = - \mathbb{E}_{(x,y) \sim D}[logP(y_t|x')]
\end{equation}

where $x'$ is sampled from the distribution of model outputs at each decoding time step, and $y_t$ is the target attribute value. Policy gradient algorithm \cite{sutton1999policy} is used to for discrete training with the attribute classifiers. 

\textbf{Reconstruction Loss $L_{rec}$}: The reconstruction loss attempts to preserve the original content information in the transferred sentences. Specifically, the loss function constricts the adapters to capture informative features to reconstruct the original sentence using the learned representations. Formally, we define $L_{rec}$ as follows:

\begin{equation}
    L_{rec} = -logP(x|z_i,y_i)
\end{equation}

where $y_i$ is the $i$-th attribute value of the input sentence, $z_i$ denotes the hidden representation extracted by the corresponding adapter. The input sentences are only reconstructed by the corresponding adapter and transferred by other adapters.

\textbf{Putting them together,} the final joint training loss $L$ is as follows:

\begin{equation}
    L = (1 - \lambda) L_{rec} + \lambda L_{cls}
\end{equation}

Where $\lambda$ is a balancing hyper-parameter to ensure that the transferred sentence has the target style while preserving the original content.

\section{Experiments}

\subsection{Experiment Setting}
\textbf{Datasets.} We evaluate the proposed \textsf{Adapter-TST} model on sentiment transfer  and multiple-attribute TST tasks using the Yelp\footnote{https://github.com/luofuli/DualRL} and StylePTB \cite{lyu-etal-2021-styleptb} datasets, respectively. We adopt the train, development, and test split for the Yelp dataset as \cite{Luo19DualRL}. \citet{lyu-etal-2021-styleptb} introduce StylePTB \footnote{https://github.com/lvyiwei1/StylePTB}, a large-scale benchmark with compositions of multiple-attribute TST tasks which allow the modeling of fine-grained stylistic changes. In our experiments, we choose four subsets for multiple-attribute TST: Tense-Voice, Tense-PP-Front$\leftrightarrow$Back, Tense--PP-Removal, and Tense-ADJADV-Removal. Specifically, the four subsets include five attributes, tense with three attribute values (\textit{Future}, \textit{Past}, \textit{Present}), voice with two attribute values (\textit{Passive}, \textit{Active}), proposition position with two attribute values (\textit{Front}, \textit{Back}),  proposition removal with two attribute values (\textit{Adding}, \textit{Removing}), and adjectives\&adverbs removal with two attribute values (\textit{Adding}, \textit{Removing}). Table~\ref{tbl:dataset_statistics}  shows the training, validation, and test splits of the Yelp and StylePTB datasets used in our experiments.

\begin{table}
\caption{Dataset statistics for Yelp and StylePTB.}
\small
\centering
\begin{tabular}{c|c|c|c}
\hline
\textbf{Dataset} &  \textbf{Train} & \textbf{Dev} & \textbf{Test} \\
\hline\hline
Yelp  & 443K & 1,000 & 1,000 \\
Tense-voice & 28K & 1,538 & 1,564 \\
Tense-PP-Front$\leftrightarrow$Back & 5K & 270 & 284 \\
Tense-PP-Removal & 32K & 1,796 & 1,834 \\
Tense-ADJADV-Removal & 33K & 1,838 & 1,819 \\
\hline
\end{tabular}
\label{tbl:dataset_statistics}
\end{table}

\textbf{Baselines.} For sentiment transfer, we benchmark \textsf{Adapter-TST} against nine state-of-the-art TST models: \textit{BackTrans} \cite{prabhumoye2018style}, \textit{CrossAlign} \cite{shen2017style}, \textit{DualRL} \cite{Luo19DualRL}, \textit{Unpaired} \cite{li2019domain}, \textit{UnsuperMT} \cite{zhang2018style}, \textit{Style Transformer} \cite{dai2019style}, \textit{DeleteOnly}, \textit{Template}, and \textit{Del\&Retri} \cite{li2018delete}. For multiple stylistic attribute outputs task, \textit{Style Transformer} \cite{dai2019style}, a transformer-based model for single-attribute TST, is selected as a baseline. We train multiple \textit{Style Transformer} models for each attribute and perform style transfer separately. For compositional editing, we use the trained \textit{Style Transformer} models to perform sequential editing, which transfers one attribute after another to compare results with our model. We term this baseline as \textit{Sequential Style Transformer} setup. 

\textbf{Training.} The experiments were performed on an Ubuntu 20.04.3 LTS system with 24 cores, 128 GB RAM, and Nvidia RTX 3090. The model implementation is based on AdapterHub \cite{pfeiffer2020AdapterHub} and Huggingface Transformers \cite{wolf-etal-2020-transformers}. For the balancing hyper-parameter $\lambda$, we choose the best-performed one from (0.9, 1) as the BART-large and T5-large models can copy the input without training with TST objectives. 

\subsection{Automatic Evaluation}
We evaluate the proposed model and baselines on three criteria commonly used in TST studies: \textit{transfer strength}, \textit{content preservation}, and \textit{fluency}. An attribute classifier is first pre-trained to predict the attribute label of the input sentence. The classifier is subsequently used to approximate the style transfer accuracy (ACC) of the sentences’ transferred attributes by considering the target attribute value as the ground truth. To quantitatively measure the amount of original content preserved after style transfer operations, we employ BERTscore \cite{zhang2019bertscore} between style-transferred and original sentences. For fluency, We use GPT-2 \cite{radford2019language} to measure the perplexity (PPL) of transferred sentences. The sentences with smaller PPL scores are considered more fluent. Finally, we compute the geometric mean of ACC, BERTscore, and 1/PPL. We take the inverse of the calculated perplexity score because a smaller PPL score corresponds to better fluency. When there is more than one accuracy in the multiple-attribute TST tasks, we use the average accuracy to compute G-score.

\begin{table}[t]
\caption{Performance of models on Yelp dataset (Sentiment Transfer Task). The best performances are \textbf{bold}. }
\small
\centering
\begin{tabular}{ccccc}
\hline
\textbf{Model} & \textbf{ACC}  & \textbf{BS} & \textbf{PPL} & \textbf{G} \\
\hline
BackTrans & 94.5 & 0.88 & 11.3 & 1.95 \\
CrossAlign & 74.3 & 0.89 & 35.3 & 1.23 \\
DeleteOnly & 87.6 & 0.91 & 36.4 & 1.30 \\
Del\&Retri & 90.2 & 0.91 & 34.0 & 1.34 \\
DualRL & 88.9 & \textbf{0.95} & 27.1 & 1.46 \\
Template & 83.7 & 0.92 & 47.2 & 1.18 \\
Unpaired & 50.6 & 0.91 & 53.1 & 0.95 \\
UnsuperMT & 96.2 & 0.93 & 33.5 & 1.39 \\
Style Transformer & 85.8 & \textbf{0.95} & 10.1 & 2.00 \\
\hline
\textsf{Adapter-TST-BART} & 90.1 & 0.91 & 8.2 & 2.15 \\
\textsf{Adapter-TST-T5} & \textbf{97.3} & 0.89 & \textbf{4.8} & \textbf{2.62} \\
\hline
\end{tabular}
\label{tbl:yelp_results}
\end{table}

\begin{table*}[t]
\caption{Automatic evaluation results of models on multiple stylistic attribute outputs task. The best performances are \textbf{bold}.}
\small
\setlength{\tabcolsep}{3pt}
\centering
\begin{tabular}{cc|ccccc|ccccc}
\hline
& & \multicolumn{5}{c}{Tense-Voice} & \multicolumn{5}{|c}{Tense-ADJADV-Removal} \\
\textbf{Model} & \textbf{Attri} & \textbf{Tense}  & \textbf{Voice}  & \textbf{BS} & \textbf{PPL} & \textbf{G} & \textbf{Tense}  & \textbf{Removal}  & \textbf{BS} & \textbf{PPL} & \textbf{G} \\
\hline
Style Transformer & single & 91.1 & - & 0.91 & 15.3 & 1.76 & 92.6 & - & 0.92 & 27.0 & 1.47 \\
Style Transformer & single & - & \textbf{87.2} & 0.85 & 11 & 1.89 & - & 83.7 & 0.93 & 21.7 & 1.53 \\
\textsf{Adapter-TST-BART} & multi & \textbf{96.9} & 81.9 & \textbf{0.96} & 4.7 & 2.63 & \textbf{96.2} & 76.5 & \textbf{0.95} & 11.8 & 1.91 \\
\textsf{Adapter-TST-T5} & multi & 95.9 & 83.4 & 0.94 & \textbf{2.6} & \textbf{3.19} & 95.7 & \textbf{85.2} & 0.91 & \textbf{3.8} & \textbf{2.79} \\
\hline
& & \multicolumn{5}{c}{Tense-PP-Front$\leftrightarrow$Back} & \multicolumn{5}{|c}{Tense-PP-Removal} \\
\textbf{Model} & \textbf{Attri} & \textbf{Tense}  & \textbf{F$\leftrightarrow$Back}  & \textbf{BS} & \textbf{PPL} & \textbf{G} & \textbf{Tense}  & \textbf{Removal}  & \textbf{BS} & \textbf{PPL} & \textbf{G} \\
\hline
Style Transformer & single & \textbf{95.}7 & - & 0.83 & 6.8 & 2.27 & 94.9 & - & 0.91 & 27 & 1.47 \\
Style Transformer & single & - & \textbf{57.2} & 0.83 & 10.4 & 1.66 & - & 87.2 & 0.91 & 26.1 & 1.45 \\
\textsf{Adapter-TST-BART}  & multi & 88.2 & 48.9 & \textbf{0.96} & 4 & 2.54 & \textbf{96} & 74.5 & \textbf{0.96} & 12.5 & \textbf{1.87} \\
\textsf{Adapter-TST-T5}  & multi & 75.9 & 48.2 & 0.83 & \textbf{1.7} & \textbf{3.12} & 95.6 & 95.0 & 0.92 & \textbf{3.8} & \textbf{2.85} \\
\hline
\end{tabular}
\label{tbl:multi_output_results}
\end{table*}

\begin{table*}[t]
\caption{Automatic evaluation results of models on compositional editing task. The best performances are \textbf{bold}.}
\small
\setlength{\tabcolsep}{3pt}
\centering
\begin{tabular}{c|ccccc|ccccc}
\hline
& \multicolumn{5}{c}{Tense-Voice} & \multicolumn{5}{|c}{Tense-ADJADV-Removal} \\
\textbf{Model}  & \textbf{Tense}  & \textbf{Voice}  & \textbf{BS} & \textbf{PPL} & \textbf{G} & \textbf{Tense}  & \textbf{Removal}  & \textbf{BS} & \textbf{PPL} & \textbf{G} \\
\hline
Sequential Style Transformer & 80.2 & \textbf{88.1} & 0.85 & 22.2 & 1.48 & 88.6 & 90.0 & \textbf{0.89} & 42.2 & 1.23 \\
\textsf{Adapter-TST-BART}  & \textbf{88.2} & 85.4 & 0.90 & 8.0 & 2.14 & \textbf{88.9} & \textbf{92.7} & 0.86 & 22 & 1.53 \\
\textsf{Adapter-TST-T5}  & 75.6 &	73.1&	\textbf{0.91}&	\textbf{6.5}&	\textbf{2.18} & 85.4&	77.1&	0.87&	\textbf{12.1}&	\textbf{1.80} \\
\hline
& \multicolumn{5}{c}{Tense-PP-Front$\leftrightarrow$Back} & \multicolumn{5}{|c}{Tense-PP-Removal} \\
\textbf{Model} & \textbf{Tense}  & \textbf{F$\leftrightarrow$Back}  & \textbf{BS} & \textbf{PPL} & \textbf{G} & \textbf{Tense}  & \textbf{Removal}  & \textbf{BS} & \textbf{PPL} & \textbf{G} \\
\hline
Sequential Style Transformer & 76.1 & \textbf{65.7} & 0.82 & 8.1 & 1.93 & \textbf{91.2} & 85.7 & 0.88 & 51.4 & 1.15\\
\textsf{Adapter-TST-BART} &  \textbf{88.2} & 50.0 & \textbf{0.92} & 4.9 & 2.35 & 90.1 & \textbf{88.2} & 0.86 & 20.9 & 1.54 \\
\textsf{Adapter-TST-T5} &  73.5&	51.4&	0.88&	\textbf{1.8}&	\textbf{3.13} & 82.1&	81.7&	\textbf{0.91}&	\textbf{19.7}&	\textbf{1.56} \\
\hline
\end{tabular}
\label{tbl:compsotional_results}
\end{table*}

\subsection{Automatic Evaluation Results}
Table \ref{tbl:yelp_results} shows the performance of the \textsf{Adapter-TST} model and the baselines on the sentiment transfer task. \textsf{Adapter-TST} has achieved the best G-score, outperforming the baselines. We observe that \textsf{Adapter-TST} achieves comparable performance on transfer strength and content preservation with 97.3\% transfer accuracy and 0.89 BERTscore by only updating the parameters of adapters. With the impressive generative ability of the pre-trained BART-large and T5-large models, the \textsf{Adapter-TST} model can generate high-quality text in terms of fluency and completeness. The experiment results demonstrate \textsf{Adapter-TST}'s ability to perform TST well and efficiently with fewer training parameters.

Table \ref{tbl:multi_output_results} presents the results of the proposed \textsf{Adapter-TST} model and Style Transformer baselines for the multiple stylistic attribute output task. Our \textsf{Adapter-TST} model achieves the highest G-score across all four datasets by simultaneously modeling multiple attributes using different adapters. \textsf{Adapter-TST} performs well in transferring tense attributes, surpassing the baselines on three datasets. However, modeling multiple attributes together proves to be a more challenging task. While \textsf{Adapter-TST} exhibits a slight performance gap compared to the Style Transformer model in terms of transfer accuracy, it excels in generating fluent and coherent sentences while preserving the original content. This advantage allows \textsf{Adapter-TST} to outperform the baselines in content preservation and fluency. It is also worth noting that training multiple Style Transformers for the multiple-attribute TST tasks is computationally inefficient and expensive, unlike \textsf{Adapter-TST}.

To demonstrate that the attribute-specific adapters capture the corresponding attribute information, we evaluate the proposed \textsf{Adapter-TST} model on the compositional editing task. Note that the parameters of adapters trained in the multiple stylistic attribute outputs task are reloaded, and the connection method is changed to \textit{Stack} for compositional editing. Table \ref{tbl:compsotional_results} shows the performance of the \textsf{Adapter-TST} and Sequential Style Transformer on the compositional editing task. The \textsf{Adapter-TST} model achieves the highest G-score across four datasets, similar to the results obtained in the multiple stylistic attribute output task. We observe that the average G-score of the multiple stylistic attribute outputs task is 2.24, significantly higher than compositional editing's average G-score of 1.89. The difference in the average G-score highlights the challenge of the compositional editing task. Interestingly, \textsf{Adapter-TST} achieves comparable performance on style transfer accuracy over attributes, indicating that the attribute-specific adapters effectively capture the stylistic attributes. 

\begin{table*}[t]
\caption{Human evaluation results of models on both multiple stylistic attribute outputs and compositional editing tasks. The best performances are \textbf{bold}.}
\small
\centering
\begin{tabular}{cc|ccccc}
\hline
&& \multicolumn{5}{c}{Tense-Voice}\\
\textbf{Model} & \textbf{Attribute} & \textbf{Tense}  & \textbf{Voice}  & \textbf{BS} & \textbf{PPL} & \textbf{G}  \\
\hline
Style Transformer & Tense & 90.0 & - & 3.72 & 3.23 & 10.26 \\
Style Transformer & Voice & - & \textbf{74.0} & 2.16 & 2.42 & 7.29 \\
\textsf{Adapter-TST} (ours) & Tense+Voice & \textbf{99} & 67.0 & \textbf{3.74} & \textbf{3.58} & \textbf{10.35} \\
\hline
Sequential Style Transformer & Tense+Voice & 81.0 & \textbf{85.0} & 2.56 & 2.88 & 8.49\\
\textsf{Adapter-TST} (ours) & Tense+Voice & \textbf{93.0} & 82.0 & \textbf{3.19} & \textbf{3.00} & \textbf{9.43}\\
\hline
\end{tabular}
\label{tbl:human_results}
\end{table*}

\begin{table*}[t]
\caption{Qualitative results for transfer to different target style combination across different models. Different colors highlight the transferred segments contributing to the target style.}
\small
\centering
\begin{tabular}{lp{3.6cm}p{3.6cm}p{3.6cm}}
\hline
\textbf{Target Style} & \textbf{Source Sentence} & \textbf{Style Transformer} & \textbf{\textsf{Adapter-TST}} \\
\hline
Future & The plan lacked a withdrawal timetable.  & The plan \red{will} lack be had by the buy-out group. & The plan \blue{will have} a withdrawal timetable. \\ \hline
Past & Some issues will be helped by higher earnings. & Some issues \red{were} helped by higher earnings by some issues. & Some issues \blue{were} helped by higher earnings. \\ \hline
Present & And he will question the white house dedication. & And he \red{question} the white house and he & And he \blue{says} the white house dedication. \\\hline
Future+passive & Litigation sciences doesn't make moral distinctions. & Litigation transportation \red{will not} make fuel had and this teaches us and no she \red{will be had by} either. & Moral distinctions \blue{will be done by} Litigation sciences. \\\hline
Past+Active & Third high yields are offered by them. & Third nutmeg yields \red{listed} night board them third period earnings stage  & Third high yields \blue{offered to} them. \\
\hline
\hline
\end{tabular}
\label{tbl:case_study}
\end{table*}

\subsection{Human Evaluation}

We conducted a human-based evaluation study to assess the performance of the \textsf{Adapter-TST} model in handling multiple-attribute TST tasks. The study involved randomly sampling 200 sentences from the Tense-Voice dataset. Both \textsf{Adapter-TST} and the baselines were used to generate multiple stylistic attribute outputs and perform compositional editing on the sampled sentences. Two linguistic researchers evaluated the generated sentences based on three criteria used in automated evaluation. To measure \textit{transfer strength}, evaluators indicated whether the sentences were in the target attribute value (e.g., future tense, passive voice) using a true/false indicator. For \textit{content preservation}, evaluators rated the amount of preserved content on a 5-point Likert scale, ranging from no content preserved (1) to all content preserved (5). Fluency was assessed on a 5-point Likert scale, where 1 represented unreadable sentences with numerous grammatical errors, and 5 indicated perfect and fluent sentences. To reduce biases, the model names were concealed, and the order of the models was randomized when displaying the generated sentences. This ensured that evaluators were unaware of which model generated the sentences they were evaluating.

\subsection{Human Evaluation Results}
Table \ref{tbl:human_results} shows the evaluation results. The style transfer accuracy of the models was computed using the binary feedback from the evaluators. The average scores for the criteria of content preservation and fluency were calculated using the 5-point Likert scores. \textsf{Adapter-TST} is observed to outperform the baselines in content preservation, fluency, and G-score. \textsf{Adapter-TST} is also rated to generate more syntactically sound and fluent sentences compared to the baselines. We can also observe that there is still a style transfer accuracy drop of \textsf{Adapter-TST} on attribute Voice when modeling multiple attributes at the same time. These results align with the automatic evaluations and demonstrate \textsf{Adapter-TST}'s effectiveness in performing multiple-attribute TST well and efficiently.

\section{Case Study}
We conducted case studies to showcase the style transferred outputs of both the \textsf{Adapter-TST} and Style Transformer models. Randomly sampled examples and their corresponding outputs are presented in Table \ref{tbl:case_study}, specifically for the Tense-Voice dataset. Our findings reveal that \textsf{Adapter-TST} successfully transfers the style while preserving the content and sentence structure in multiple-attribute TST tasks. In contrast, the Style Transformer model generates sentences with grammatical errors, making it challenging to determine if the style transfer was successful. Moreover, the Style Transformer model performs poorly in the task of compositional editing due to its inherent complexity. Despite the difficulty of compositional editing, \textsf{Adapter-TST} is capable of generating fluent sentences that preserve the original content.

\section{Summary}

In this chapter, we introduced a parameter-efficient framework, \textsf{Adapter-TST} with different neural adapters to capture different attribute information for multiple-attribute TST tasks. During training, the original parameters of BART-large were frozen, and only the adapters' parameters were optimized to relax the dependence on computational resources and supervised data. We conducted extensive experiments on traditional sentiment transfer and multiple-attribute TST tasks. The automatic and human-based evaluation results showed that the attribute-specific adapters in \textsf{Adapter-TST} is able to capture relevant stylistic attributes to transfer the style while preserving the original content successfully. Our case studies also demonstrated that \textsf{Adapter-TST} was able to generate high-quality text in the target style. For future work, we will continue to improve TST models' ability to model multiple attributes in terms of quality and efficiency. We will also explore plugging \textsf{Adapter-TST} on other PLMs and evaluate its effectiveness.

\chapter{Contrastive Disentanglement for Authorship Attribution} 

\label{contrastiveaa} 

\section{Objective}

\textbf{Motivation.} Authorship Attribution (AA) is an extensively researched field, with the primary objective of pinpointing the author of a text by analyzing unique linguistic patterns and stylistic nuances present in their writing \cite{zheng2023review}. The significance of AA extends across diverse domains, such as digital forensics \cite{iqbal2008novel}, where it aids in identifying the origin of contentious or illicit content, and plagiarism detection \cite{stamatatos2011plagiarism}, where it helps ascertain the authenticity of written works. The complexity and variability of individual writing styles, combined with the evolving nature of language, make AA a challenging yet indispensable tool for ensuring integrity and authenticity in digital communication and documentation.

Existing methods in AA can be broadly categorized into two main groups: traditional stylometric approaches and machine learning-based techniques. Traditional stylometric methods, as explored in works by Seroussi et al. \cite{seroussi2011authorship} and Bevendorff et al. \cite{bevendorff2019generalizing}, primarily utilize features such as word lengths, sentence lengths, and the frequency of function words to attribute authorship. These methods have been foundational in the field, offering insights into the quantifiable aspects of writing styles. With the advent of more sophisticated computational models, machine learning-based techniques, particularly those leveraging deep learning, have emerged as powerful tools for AA. These approaches, exemplified in the studies by Zhang et al. \cite{zhang-etal-2018-syntax} and Saedi et al. \cite{saedi2021siamese}, aim to capture more intricate and subtle patterns in writing styles, often achieving superior performance over traditional stylometric methods. Notably, works by Rivera-Soto et al. \cite{rivera-soto-etal-2021-learning} and Wang et al. \cite{wang2023can} demonstrate how deep learning models can learn complex stylistic signatures from textual data, marking a significant advancement in the field's ability to accurately attribute authorship.

Despite significant advancements in AA, a persistent challenge emerges in scenarios involving topic shifts, particularly when models are tested on topics not encountered during training. This challenge can be attributed to the models' inability to distinguish between topic-related content and an author's unique writing style effectively~\cite{hu2023tdrlm,sapkota2014cross}. In such cases, the standard stylistic features traditionally relied upon in AA can inadvertently capture topical variations rather than the nuanced stylistic markers unique to an author. This conflation leads to inaccuracies in authorship determination, as the models may incorrectly attribute texts based on their topical content rather than on the distinct stylistic elements they are meant to analyze. The issue underscores the need for innovative approaches that can robustly disentangle topical influence from stylistic characteristics, ensuring accurate authorship attribution across diverse and unseen topics.

Furthermore, while existing research in AA has made considerable strides, it has primarily focused on attributing authorship at the individual author level. This focus has inadvertently led to the underestimation of the rich potential inherent in regional-level AA. Investigating AA from a regional perspective can uncover distinct linguistic styles that are shared among authors within the same geographical area, deeply rooted in cultural nuances and regional dialects. For instance, the use of English in Singapore provides a compelling example, where local cultural influences and slang imbue the language with distinctive characteristics that are crucial for effective AA at a regional scale. Such regional linguistic variations offer a unique dimension that is essential for enhancing the robustness and accuracy of AA methods. Therefore, it is imperative to extend our investigative efforts to fully understand and leverage the nuances of regional linguistic variations for AA, which could significantly improve the field's methodologies and applications.

\textbf{Research Objectives.} This chapter introduces \textsf{ContrastDistAA}, an innovative approach to AA that addresses the critical challenge of topic variability. By employing contrastive learning combined with mutual information maximization, \textsf{ContrastDistAA} effectively disentangles topic-related content from stylistic features within latent representations. This dual-focus approach enables robust authorship determination in the face of topic shifts and facilitates AA analysis at both the individual and regional levels. Recognizing the limitations of existing datasets for regional-level AA, we have also developed a new dataset specifically designed to support and enhance the evaluation of AA tasks across geographical and cultural boundaries. Through comprehensive experiments, \textsf{ContrastDistAA} is rigorously evaluated against leading state-of-the-art baselines in AA, demonstrating its superior performance and versatility in handling both regional and individual-level attribution tasks.

\textbf{Contributions.} In this study, we make several significant contributions to the field of AA:
\begin{itemize}
    \item We present a pioneering regional-level AA task, accompanied by a meticulously curated dataset designed to challenge and evaluate AA methods in this novel context. This dataset fills a critical void in AA research by facilitating the exploration of authorship across diverse cultural and countries.
    \item Our proposed \textsf{ContrastDistAA} framework represents a methodological leap in AA, employing contrastive learning and mutual information maximization to effectively separate content and style within textual data. This allows for unprecedented accuracy in AA tasks, particularly in scenarios plagued by topic shifts.
    \item Through extensive experimentation, we benchmark \textsf{ContrastDistAA} against contemporary state-of-the-art AA methods across a variety of settings. The results unequivocally demonstrate \textsf{ContrastDistAA}'s superior performance in both individual and regional-level AA tasks, showcasing its effectiveness and adaptability.
\end{itemize}

Our contributions extend beyond mere performance enhancements in AA; they pave the way for new insights into the complex dynamics between linguistic styles and cultural influences. By addressing the overlooked aspect of regional variability in AA, our work offers fresh perspectives and deepens the understanding within the AA domain, thereby enriching the broader discourse in computational linguistics.

\section{Methodology}

\begin{figure*}[t] 
	\centering
	\includegraphics[width=0.85\textwidth]{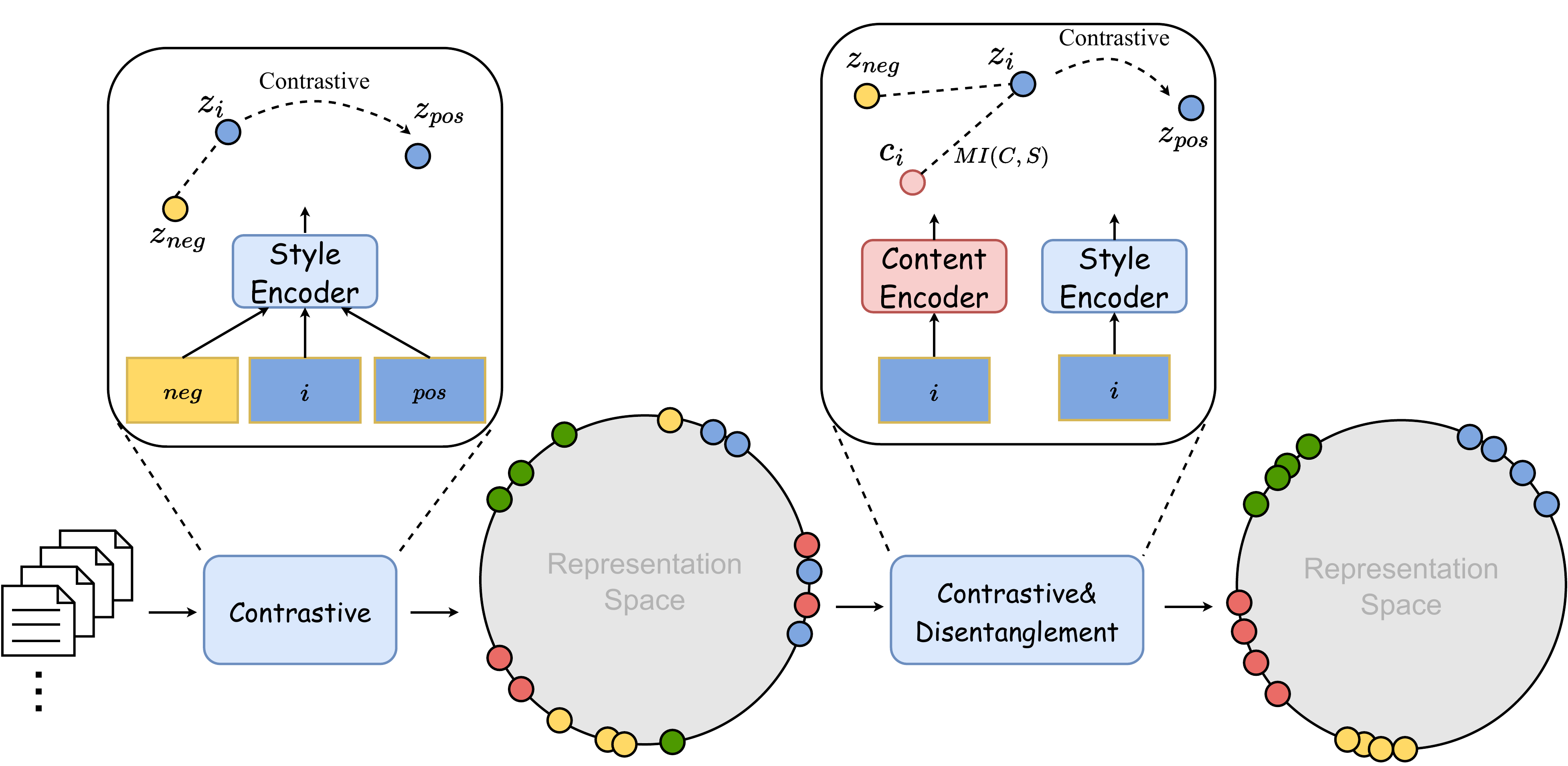} 
	\vspace{-5pt}
	\caption{The architecture overview of \textsf{ContrastDistAA} model. The proposed models contains two-stages training process: (i) training using contrastive loss, and (ii) training using both contrastive loss with disentanglement loss.}
        \label{fig:model_overview}
	\vspace{-8pt}
\end{figure*}

This section outlines our proposed model, \textsf{ContrastDistAA}, designed to learn a disentangled representation of writing style for AA. As depicted in Figure~\ref{fig:model_overview}, \textsf{ContrastDistAA} is structured in two distinct phases. The initial phase employs supervised contrastive loss to extract key stylistic features from labeled data. However, given the potential for content-related information to be intertwined with style, thus impacting the robustness of AA models, the subsequent phase of \textsf{ContrastDistAA} introduces a mutual information-based approach. This technique aims to separate style and content representations in the latent space, thereby enhancing the effectiveness of contrastive learning by clearly differentiating between style and content-specific attributes, including topical elements.

In subsequent sections, we will first review the contrastive learning component and the associated contrastive losses. This is followed by an introduction to mutual information, which is applied to disentangled representation learning for AA.

\subsection{Contrastive Learning}
Self-supervised representation learning has seen considerable progress in recent years, largely attributable to the application of contrastive learning~\cite{wu2018unsupervised,hénaff2020dataefficient,oord2018representation,chen2020simple}. The fundamental mechanism of contrastive learning involves drawing an anchor and a ``\textit{positive}'' sample closer in an embedding space, while simultaneously distancing the anchor from multiple ``\textit{negative}'' samples, thus yielding meaningful representations. Specifically for AA tasks, we define ``\textit{positive pair}'' consists of a text sample authored by the same individual as the anchor within a minibatch. In contrast, ``\textit{negative pairs}'' are formed by aligning the anchor with randomly chosen samples from different authors within the same minibatch.

The initial phase of \textsf{ContrastDistAA} involves applying contrastive learning to train a style encoder, which extracts style features from texts authored by individuals or authors from specific regions. We utilize BERT~\cite{devlin2019bert}, acclaimed for its proficiency in capturing writing styles, as the style encoder. This encoder transforms discrete text into representations within latent space. Following this, supervised contrastive loss is applied to align representations of texts by the same author or from the same region more closely, while simultaneously distinguishing those from different authors or regions. This methodology enhances the style encoder's ability to discern and learn discriminative style representations.

\subsubsection{Supervised Contrastive Loss for AA}
In the \textsf{ContrastDistAA} model, we implement a supervised contrastive loss for AA. Consider a batch consisting of $N$ textual samples from distinct authors. Let $i \in I \equiv \{1,2,\cdots,N\}$ represent an individual sample in the minibatch, and let $A(i) \equiv I \backslash \{i\}$ denote the set of other texts excluding $i$. The negative samples for anchor $i$, denoted as $NEG(i) \equiv \{ neg \in A(i) : y_{neg} \neq y_i\}$, are those not sharing the same author as $i$, while $POS(i) \equiv \{ pos \in A(i) : y_{pos} = y_i\}$ represents the positive samples, sharing the same author as $i$. The supervised contrastive loss is particularly effective in scenarios where multiple samples belong to the same class, as it utilizes the available labels \cite{khosla2021supervised}. The formulation of the supervised contrastive loss for AA tasks is as follows:

\begin{equation}
\begin{split}
    L^{sup} = \sum_{i \in I} \frac{-1}{|POS(i)|} \sum_{pos \in POS(i)}  
    log \frac{exp(z_i \cdot z_{pos} / \tau)}{\sum_{neg \in NEG(i) exp(z_i \cdot z_{neg} / \tau)}}
\end{split}
\end{equation}

where $z_i = StyleEncoder(x_i)$, the $\cdot$ symbol denotes the inner product, $\tau \in \mathcal{R^+}$ is a scalar temperature parameter, $POS(i) \equiv \{ pos \in A(i) : y_{pos} = y_i\}$ is the set of indices of all positive samples distinct from i, and $|POS(i)|$ is its cardinality.

\subsection{Mutual Information for Style-Content Disentanglement}
The style encoder, trained using supervised contrastive loss, becomes proficient at extracting representations that encapsulate both style and content attributes. Therefore, to refine the style encoder's focus on capturing writing style more distinctly, we integrate mutual information with contrastive learning. This synergy aims to separate style and content information within the latent space.

Mutual information, a fundamental concept in information theory, measures the dependence between two random variables. For our model, mutual information between style ($z$) and content ($c$) representations is crucial. Its mathematical definition involves the expectation of the logarithm of the ratio of the joint distribution of $z$ and $c$ to their respective marginal distributions, which can be expressed as follows:

\begin{equation}
    I(z;c) = \mathbb{E}_{p(z,c)} [log \frac{p(z,c)}{p(z)p(c)} ]
    \label{eq:mi}
\end{equation}

In practice, accurately calculating mutual information is challenging due to the intractability of the integral involved~\cite{chen2016infogan,belghazi2018mutual,poole2019variational}. To address this, we employ the Contrastive Log-ratio Upper Bound (CLUB) estimation method~\cite{cheng2020club}. This approach is particularly suitable when conditional distributions such as $p(z|c)$ or $p(c|z)$ are not explicitly available. $p(z|c)$ is approximated by using a variational distribution $q_{\theta}(z|c)$, parameterized by $\theta$, leading to the definition of the variational CLUB term (vCLUB) as follows:

In disentangled representation learning, a common objective is to minimize the mutual information between varying types of embeddings, aligning with our training target~\cite{poole2019variational}. However, determining the exact value of mutual information presents challenges in practical settings, as the integral in Eq. \ref{eq:mi} is often intractable. To overcome this, several mutual information estimation methods have been proposed~\cite{chen2016infogan,belghazi2018mutual,poole2019variational}. We employ the estimation method known as the Contrastive Log-ratio Upper Bound (CLUB)~\cite{cheng2020club}, which is suitable for the scenario where the conditional distributions $p(z|c)$ or $p(c|z)$ is not provided. A variational distribution $q_{\theta}(z|c)$ with parameter $\theta$ is used to approximate $p(z|c)$. Consequently, a variational CLUB term (vCLUB) is defined as follows:

\begin{equation}
\begin{split}
     I_{vCLUB}(z;c) := \mathbb{E}_{p(z,c)}[log q_{\theta}(z|c)] - \mathbb{E}_{p(z)}\mathbb{E}_{p(c)} [log q_{\theta}(z| & c)]
\end{split}
\end{equation}

The unbiased estimator for vCLUB is derived from a set of samples, effectively quantifying the mutual information in a computationally feasible manner, which is expressed as follows:

\begin{equation}
\begin{split}
     \hat{I}_{vCLUB} = \frac{1}{N^2} \sum^N_{i=1} \sum^N_{j=1} [log q_{\theta}(z_i| c_i) - log q_{\theta}(z_j| c_i) ] \\
     = \frac{1}{N} \sum^N_{i=1}[log q_{\theta}(z_i| c_i)- \frac{1}{N} \sum^N_{j=1} log q_{\theta}(z_j| c_i)].
\end{split}
\end{equation}

In summary, to facilitate style-content disentanglement in \textsf{ContrastDistAA}, we first deploy a content encoder, also a BERT model, to extract content representations, denoted as $c$. Meanwhile, the pre-trained style encoder from the first stage extracts style representations, denoted as $z$. Each post $i$ thus has two distinct representations: the content representation $c_i$ and the style representation $s_i$. Here, we apply the vCLUB estimator to minimize the mutual information between these content and style representations, refining the distinctiveness of each. Concurrently, the supervised contrastive loss continues to enhance the style encoder’s ability to capture writing style nuances. During the evaluation phase, only the style encoder is used to extract style representations from posts authored by individuals or from specific regions. The regional or individual author style representations are then calculated by averaging the post-style representations, facilitating a comprehensive and nuanced assessment of writing styles.

\section{Experiments}
\begin{table}[t]
    \centering
    \caption{Statistics of datasets}
    \begin{tabular}{l l l l l}
        \toprule
        \textbf{Dataset} & {\textbf{\#Users}} & {\textbf{\#Train}} & {\textbf{\#Valid}} & {\textbf{\#Test}} \\
        \midrule
        Regional Tweets & 87,836 & 382,598 & 42,513 & 42,513 \\
        CCAT50 & 50 & 1,766 & 442 & 465 \\
        Twitter1000 & 1,000 & 6,000 & 2,000 & 2,000 \\
        IMDB62 & 62 & 37,200 & 12,400 & 12,400  \\
        \bottomrule    
    \end{tabular}
    \label{tab:GroupStatistics}
\end{table}

\begin{table}[t]
    \centering
    \caption{Statistics of Regional Tweet dataset. \#avg means average number of tweets for top 100 authors}
    \begin{tabular}{l c c c}
        \toprule
        \textbf{Region} & {\textbf{\#Tweets}} & {\textbf{\#Unique Users}} & {\textbf{\#avg}} \\
        \midrule
        KL & 35,568 & 9,172 & 78 \\
        Manila & 85,389 & 20,511 & 166 \\
        Jakarta & 55,406 & 17,141 & 96 \\
        Hanoi & 86,205 & 11,131 & 345 \\
        Bangkok & 87,947 & 14,841 & 210 \\
        SG & 74,596 & 15,040 & 159 \\
        \bottomrule    
    \end{tabular}
    \label{tab:RegionalStatistics}
\end{table}

\begin{table*}[t]
    \small
    \centering
    \caption{Macro and Micro F1 scores for baselines and \textsf{ContrastDistAA} on four benchmark datasets.}
    \resizebox{\linewidth}{!}{
        \begin{tabular}{lcc cc cc cc}
            \toprule
            & \multicolumn{2}{c}{\textbf{Regional Tweet}} & \multicolumn{2}{c}{\textbf{CCAT50}} & \multicolumn{2}{c}{\textbf{Twitter1000}} & \multicolumn{2}{c}{\textbf{IMDB62}} \\
            \cmidrule{2-3} \cmidrule{4-5} \cmidrule{6-7} \cmidrule{8-9}
            \textbf{Method} & \textbf{Macro F1} & \textbf{Micro F1} & \textbf{Macro F1} & \textbf{Micro F1} & \textbf{Macro F1} & \textbf{Micro F1} & \textbf{Macro F1} & \textbf{Micro F1} \\
            \midrule
            LR-Stylo & 0.176 & 0.251 & 0.013 & 0.037 & 0.019 & 0.035 & 0.013 & 0.037 \\
            LR-TF-IDF & 0.402 & 0.446 & 0.554 & 0.554 & 0.566 & 0.566 & 0.554 & 0.554 \\
            LR-Char & 0.252 & 0.308 & 0.180 & 0.209 & 0.077 & 0.128 & 0.503 & 0.503 \\
            LSTM & 0.186 & 0.290 & 0.244 & 0.274 & 0.124 & 0.126 & 0.307 & 0.326 \\
            BertAA & 0.433 & 0.472 & 0.518 & 0.512 & 0.226 & 0.249 & 0.627 & 0.654 \\
            DistilBERT & 0.407 & 0.449 & 0.453 & 0.447 & 0.213 & 0.242 & 0.402 & 0.441 \\
            Roberta & 0.476 & 0.522 & 0.466 & 0.497 & 0.622 & 0.626 & 0.735 & 0.749 \\
            \midrule
            \textsf{ContrastDistAA} & \textbf{0.510} & \textbf{0.550} & \textbf{0.578} & \textbf{0.584} & \textbf{0.960} & \textbf{0.961} & \textbf{0.813} & \textbf{0.816} \\
            \textsf{ContrastDistAA}  (w/o dist) & 0.505 & 0.508 & 0.552 & 0.566 & 0.960 & 0.916 & 0.803 & 0.816 \\
            \bottomrule
        \end{tabular}
        }
\label{tab:ResultAllDatasets}
\end{table*}

\subsection{Experimental Settings}
\textbf{Datasets}. To evaluate \textsf{ContrastDistAA} effectively on both individual and regional AA tasks, we utilize four datasets in our experiments. The statistical distributions of the datasets are shown in Table~\ref{tab:GroupStatistics}.

\textit{Regional Tweets}: Targeting the identification of regional writing styles within Southeast Asia, this dataset comprises 425,111 English tweets collected via the Twitter API over 2021-2022 from 87,836 users spanning six regions: Singapore, Kuala Lumpur, Manila, Jakarta, Hanoi, and Bangkok. The dataset was curated with a focus on tweets exceeding three words to ensure quality and relevance. As detailed in Table~\ref{tab:RegionalStatistics}, it showcases a wide range of tweet volumes across regions, from 35,000 to 88,000, reflecting the diverse user base of 9,000 to 21,000 individuals per region. This variation underscores the dataset's richness and potential for comprehensive regional AA analysis. The dataset not only fills a crucial gap in regional AA research but also demonstrates significant potential for individual-level AA studies. Preprocessing revealed that the average tweet volume per user, among the top 100 authors in each region, varies from 78 to 355 tweets, indicating active engagement and diverse expression styles. An anonymized version of this dataset is poised for release to the academic community, fostering further research and exploration in AA. 

In our study, we utilized this dataset specifically for regional AA analysis, complementing it with other datasets for individual AA tasks to facilitate a thorough comparison with existing literature. For experimental purposes, the dataset was partitioned into training, validation, and testing sets following an 8:1:1 ratio, ensuring a balanced approach to model training and evaluation.

\textit{CCAT50}: A subset of the Reuters Corpus and a prominent resource in AA research, the CCAT50 dataset~\cite{liu2012novel} focuses on the top 50 contributors in the CCAT (corporate/industrial) subtopic. It consists of 5,000 texts (50 per author) divided into distinct training, validation, and testing sets following a 6:2:2 ratio, based on the processed version by \cite{tyo2022state}.

\textit{Twitter1000}: Derived from a larger Twitter dataset used in AA research \cite{shrestha2017convolutional, schwartz2013authorship}, Twitter1000 includes tweets from the top 1,000 authors by volume, with 100 tweets randomly selected from each. The dataset is organized into training, validation, and testing subsets, also following a 6:2:2 ratio.

\textit{IMDB62}: Recognized for long-text AA studies~\cite{imdb}, the IMDB62 dataset includes contributions from 62 authors, each providing 1,000 texts. Similar to the others, this dataset is partitioned into training, validation, and testing sets in a 6:2:2 ratio.

\noindent\textbf{Evaluation Metrics.}
Following existing AA studies,  we adopt Macro-F1 and Micro-F1 as the evaluation metrics in our experiments. 

\subsection{Baselines} 
We benchmark our model against commonly used and state-of-the-art AA models. These baselines are trained or fine-tuned to perform both the regional-level and individual-level AA tasks. 

\textbf{LR-Stylo}: This logistic regression model, leveraging stylometric features as inputs, is grounded in prior research \cite{sari2018neural, aborisade2018classification}. Based on \cite{fabien-etal-2020-bertaa}, it uses ten different stylometric features like text length and word count for classification.

\textbf{LR-TF-IDF}: Employing Term Frequency - Inverse Document Frequency (TF-IDF) at the word level, this logistic regression classifier follows the approach of \cite{fabien-etal-2020-bertaa}. Pre-processing includes stemming and stop-word removal before constructing the TF-IDF features.

\textbf{LR-Char}: This model uses character N-gram-based features, shown to be effective in AA \cite{bischoff2020importance, shrestha2017convolutional, altakrori-etal-2021-topic-confusion}. Following \cite{tyo2022state}, the logistic regression classifier is trained with a mix of character N-gram, part-of-speech N-gram, and summary statistics.

\textbf{LSTM}: An LSTM model, inspired by recent studies \cite{oliva2022improving}, incorporates a dense layer followed by a max pooling layer. It focuses on the hidden states of the LSTM for AA tasks.

\textbf{BertAA}: Utilizing a pre-trained BERT language model, BertAA \cite{fabien-etal-2020-bertaa} is fine-tuned specifically for AA, integrating a dense layer and softmax activation function for AA classification.

\textbf{DistilBert}: Known for its efficiency as a compact language model, DistilBERT \cite{distilbert} is fine-tuned for AA tasks.

\textbf{Roberta}: Employing the Roberta model \cite{liu2019roberta}, we follow the original hyperparameters and fine-tune it on AA datasets over a specific number of epochs.

\begin{figure*}[t] 
    \centering
    \begin{tabular}{@{} c@{\hspace{0.01pt}} c@{\hspace{0.05pt}} c @{}}
        \includegraphics[width=0.333\linewidth]{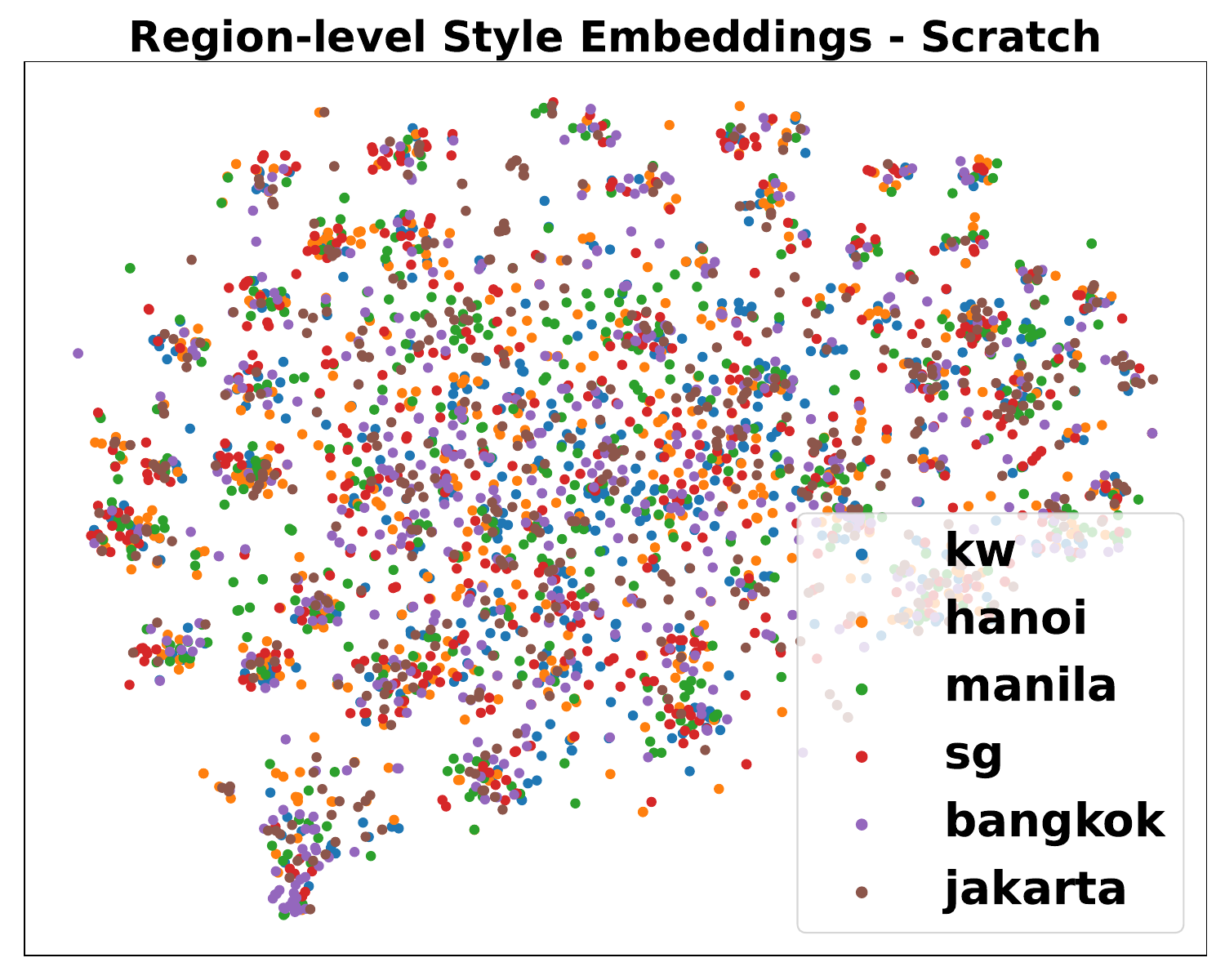} &
        \includegraphics[width=0.333\linewidth]{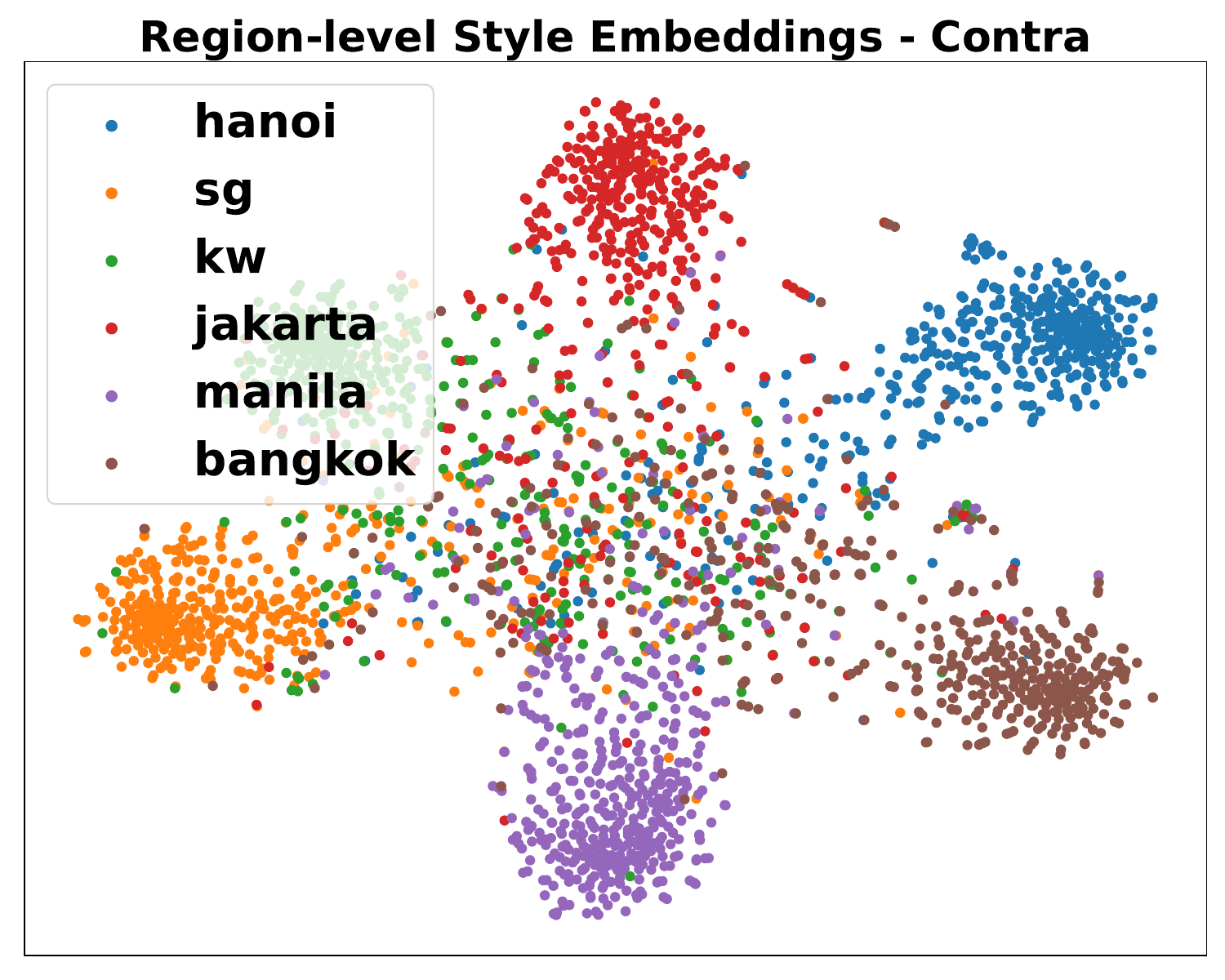} &
        \includegraphics[width=0.333\linewidth]{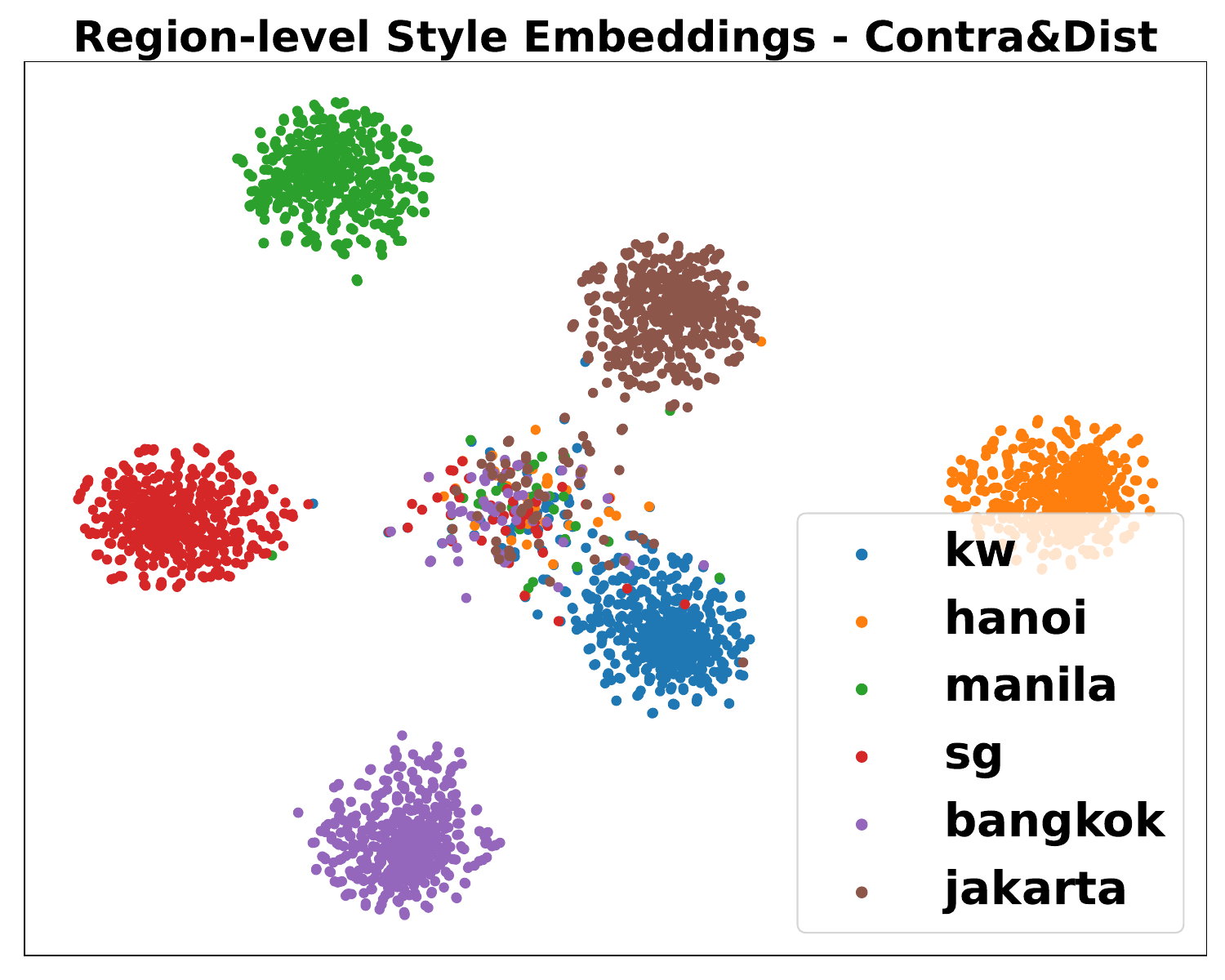} \\
        (a) & (b) & (c) \\
        \includegraphics[width=0.333\linewidth]{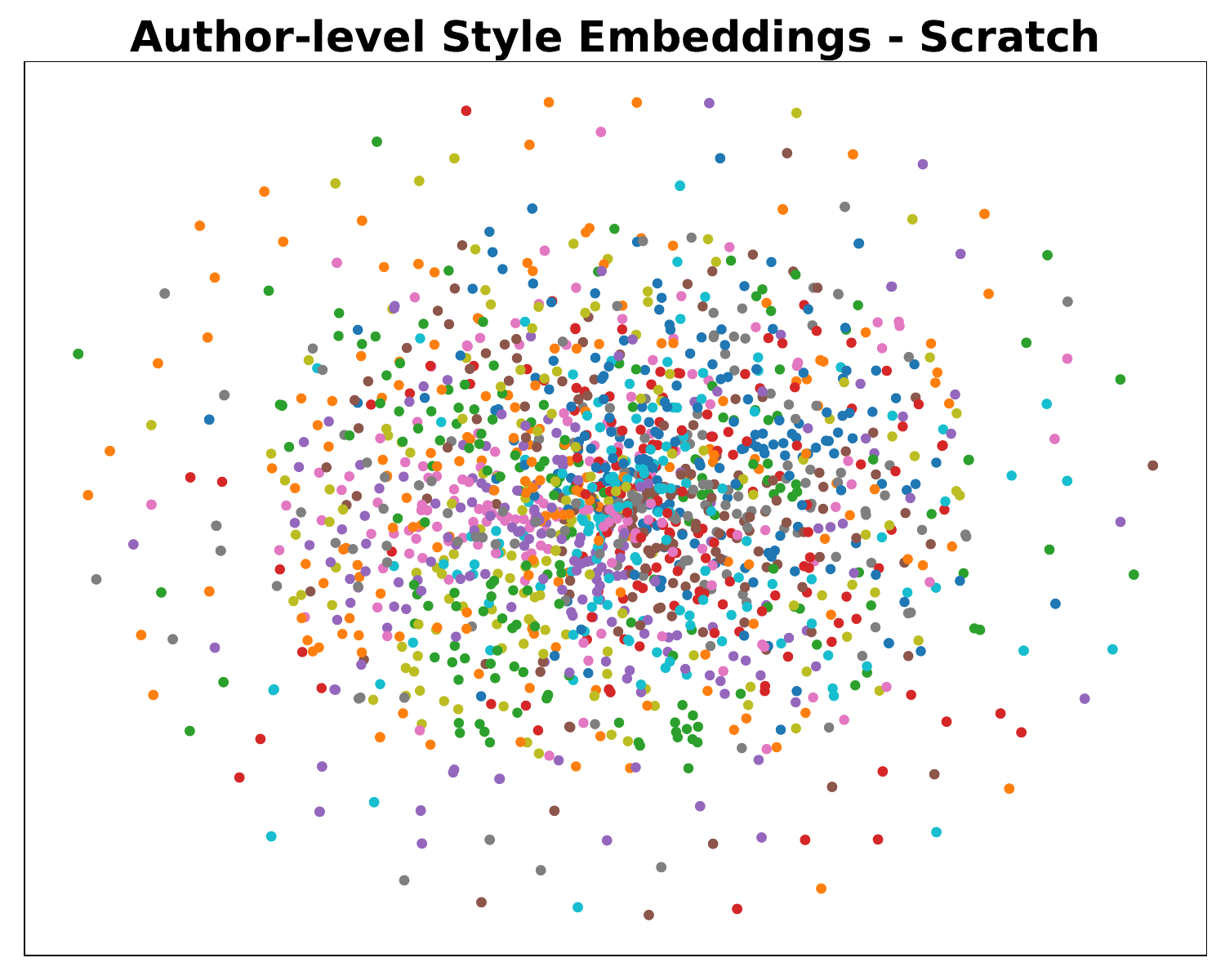} &
        \includegraphics[width=0.333\linewidth]{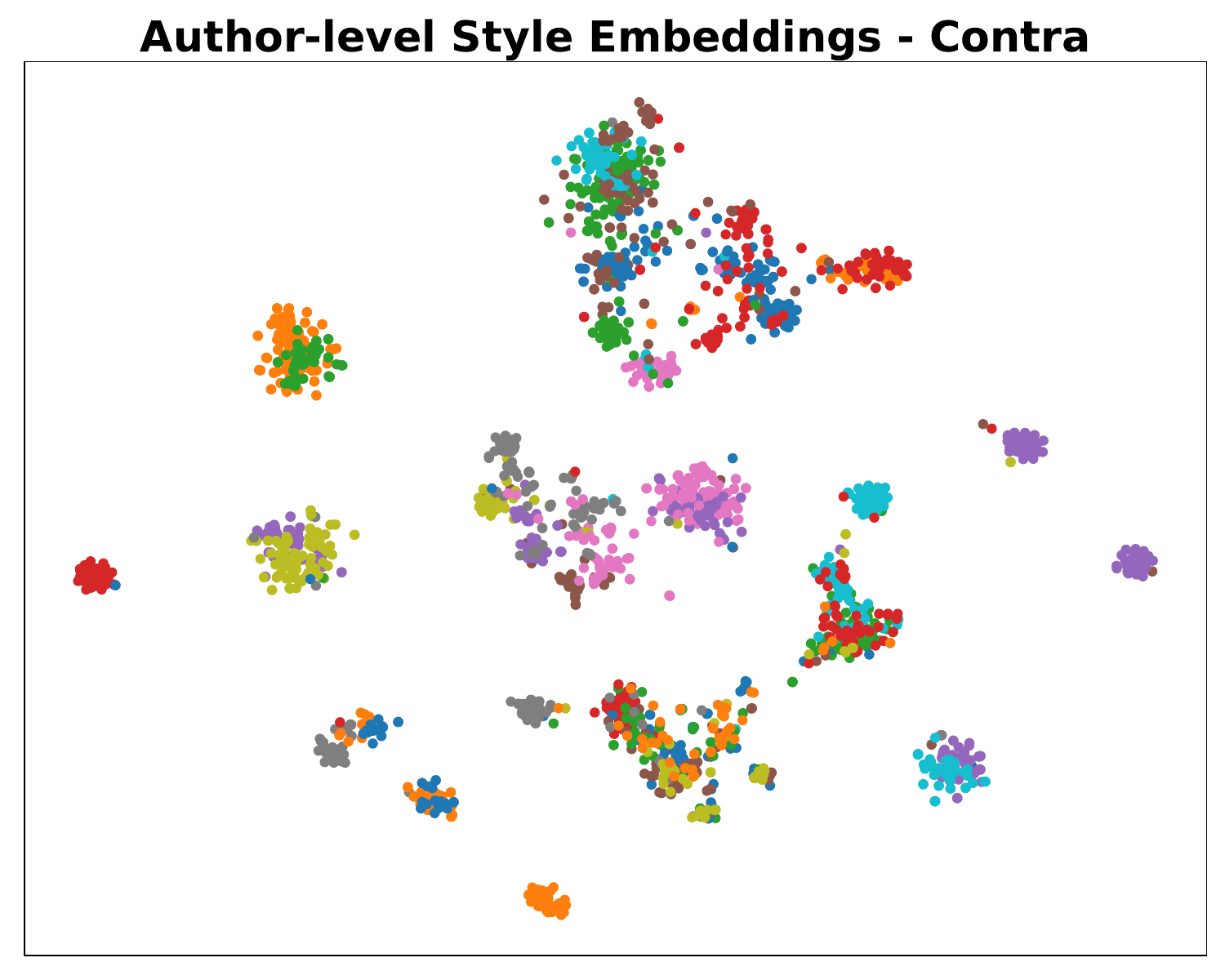} &
        \includegraphics[width=0.333\linewidth]{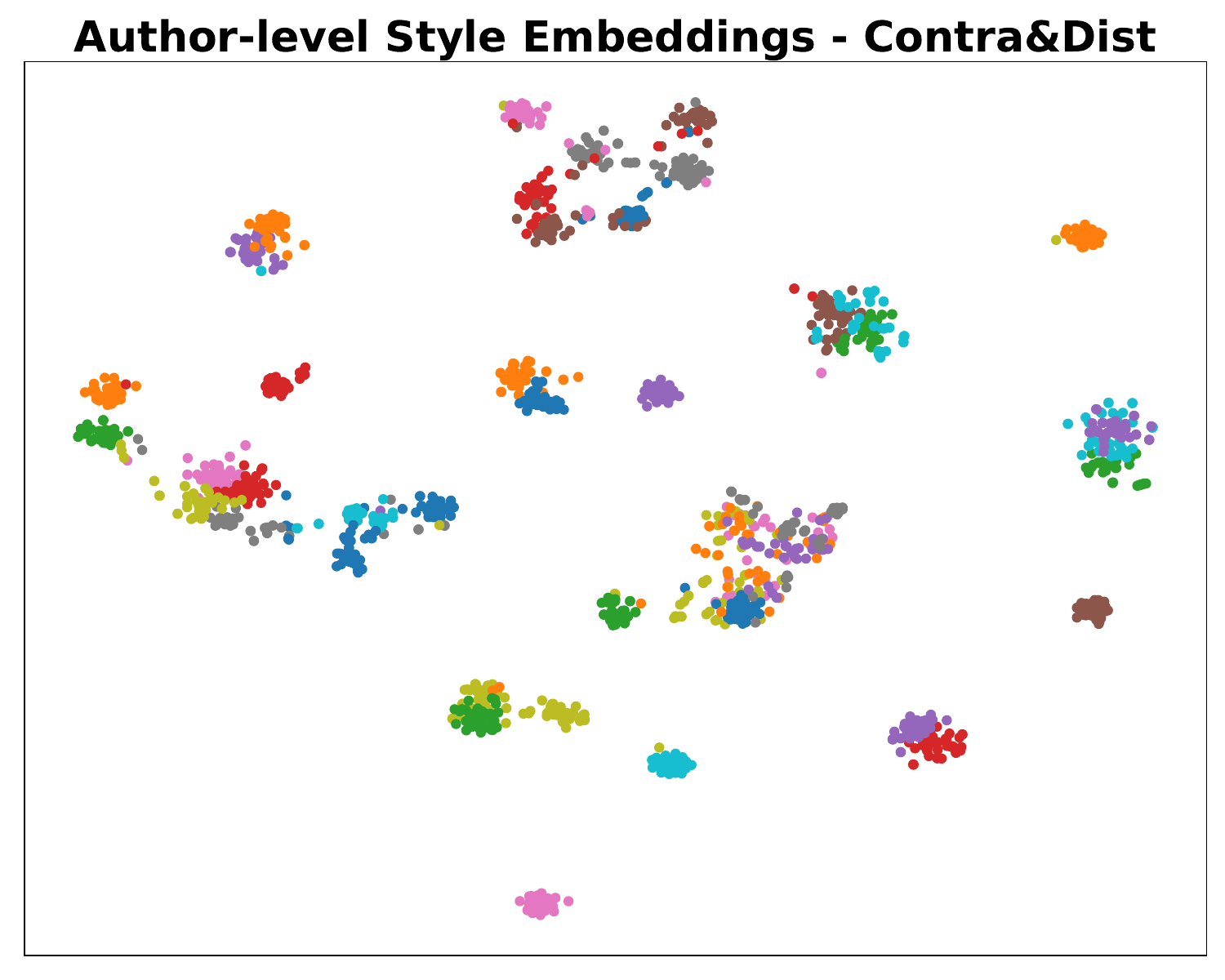} \\
        (d) & (e) & (f) \\
    \end{tabular}
    \vspace{-3pt}
    \caption{t-SNE visualization of posts from Regional Tweets and CCAT50. Specifically, we select 100 posts from each region in the Regional Tweets dataset and 50 posts from each author in CCAT50. The top three visualizations display the posts from Regional Tweets, while the bottom three pertain to CCAT50.}
    \label{fig:t-SNE visualization}
    \vspace{-10pt}
\end{figure*}

\subsection{Implementation.} 
Our experiments were carried out on a system operating on Ubuntu 20.04.3 LTS, equipped with robust hardware specifications including 24 CPU cores, 128 GB of RAM, and a base clock speed of 2.9 GHz. To facilitate efficient training of the pre-trained models, Nvidia GTX 3090 graphics cards were utilized. BERT, with its pre-trained weights, served dual roles as both the style and content encoders in our experiment, which was divided into two distinct stages.

We train \textsf{ContrastDistAA} in two stages. In the first stage, the style encoder was the sole focus, trained using supervised contrastive loss over 30 epochs. The subsequent stage marked the joint training of both the content and style encoders. This phase, extending for an additional 20 epochs, employed supervised contrastive loss alongside a mutual information estimator. The implementation of the mutual information estimator was based on the source code\footnote{https://github.com/Linear95/CLUB} provided by \cite{cheng2020club}. Consistency in training parameters was maintained throughout, with a learning rate set at 1e-3 and a batch size of 32 for both stages. This setup ensured a balanced and rigorous training process for the \textsf{ContrastDistAA} model.

\subsection{Experiment Results}

In our study, the efficacy of the \textsf{ContrastDistAA} model was thoroughly assessed on both regional and individual-level datasets, with its performance benchmarked against a range of established baseline models. The comparative results, evaluated using F1 scores, are detailed in Table~\ref{tab:ResultAllDatasets}.

The \textsf{ContrastDistAA} model consistently exhibited superior performance across these datasets. For instance, within the Regional Tweets dataset, it attained a Micro F1 score of 0.55, representing a notable 7\% improvement compared to the next closest model, BertAA. In the context of the CCAT50 dataset, \textsf{ContrastDistAA} surpassed all baselines in every evaluated metric, achieving a significant 16\% improvement in Micro F1 scores. The model also demonstrated exceptional performance on the Twitter1000 dataset, registering a substantial 29\% increase in F1 scores. Furthermore, on the IMDB62 dataset, \textsf{ContrastDistAA} achieved a 6.7\% improvement in performance, indicative of its robustness even in the presence of textual complexity. These results collectively affirm the \textsf{ContrastDistAA} model's capability in effectively discerning writing styles at both regional and individual levels, thereby establishing it as a state-of-the-art benchmark in the AA tasks.

Interestingly, we also noted that the models' F1 scores are generally lower for the Regional Tweets dataset, suggesting the difficulty of the region-level AA task. The individual authors typically have more distinct and consistent writing styles compared to a group of authors from a region. This uniqueness in individual writing styles makes it easier for models to attribute authorship accurately, leading to higher F1 scores. In contrast, regional-level AA deals with broader, less distinct writing styles shared by a group, which can be more challenging to differentiate.

\begin{table}[htb]
    \centering
    \small
    \caption{Examples showcasing the unique writing expressions (highlighted in \hl{yellow}) from each region. The similarity score is the cosine similarity between the post representation and the region style embedding.}
    \begin{tabular}{ l p{10cm} c } 
        \toprule
        \textbf{Regions} & \textbf{Sentences} & \textbf{Similarity}\\
        \midrule
        \multirow{3}{4.5em}{Bangkok} & 1. @USER thank u \hl{naa} & 0.825 \\ \cmidrule{2-3}
        & 2. @USER You're very welcome I feel honored and very happy . \hl{ka} \:pleading\_face\: 
        \:two\_hearts\: & 0.977 \\ \cmidrule{2-3}
        & 3. @USER You make all of us lazy people feel ashamed on a Sunday morning \hl{na krub} . & 0.990 \\ 
        \midrule
        \multirow{3}{4.5em}{Hanoi} 
        & 1. isit Indonesian \#Booth in \hl{Ly Thao To} Park , DATE & 0.995 \\ \cmidrule{2-3}
        & 2. Those light is fierce ! \#welldone @USER \hl{Trang Tien} Plaza HTTPURL & 0.996 \\ \cmidrule{2-3}
        & 3. try some coconut coffee \:hot\_beverage\: \@USER \hl{Cong Caphe} HTTPURL & 0.996 \\ 
        \midrule
        \multirow{3}{4.5em}{Jakarta} 
        & 1. @USER \hl{Serem amat} :loudly\_crying\_face: & 0.991 \\ \cmidrule{2-3}
        & 2. @USER batman who laughs \hl{lumayan} lah atleast & 0.951 \\  \cmidrule{2-3}
        & 3. \hl{Mantul} the babbies \hl{nyusul} the daddies & 0.992 \\ 
        \midrule
        \multirow{3}{4.5em}{Manila} 
        & 1. \hl{Salamat} sa live selling at unboxing ! Lol char . Love you bestie ! Congratulations ! HTTPURL & 0.996 \\ \cmidrule{2-3}
        & 2. Wow , \hl{salamat} po sa Dios \! To God be the Glory \:sparkles\: Are Your Prayers Heard \#PureDoctrinesOfChrist HTTPURL & 0.996 \\  \cmidrule{2-3}
        & 3. DATE \hl{nabudol} ako sa film life . Excited for youuuuuu . @USER Stay Broke , Shoot Film . HTTPUR . HTTPURL & 0.959 \\ 
        \midrule
        \multirow{3}{4.5em}{Singapore} 
        & 1. STOp . the tarot card readings gotta STOOOOOOOop pls \hl{lah} & 0.857 \\ \cmidrule{2-3}
        & 2. So much things on my mind rn ! \hl{Inshallah} all goes well & 0.651 \\  \cmidrule{2-3}
        & 3. @USER i no have scandal \hl{lehhh} u my one and only & 0.984 \\ 
        \midrule
        \multirow{3}{4.5em}{Kuala Lumpur} 
        & 1. Pusing \hl{lah} kot \hl{mana pun} , no one else is calling it democratic . Except PN of course &  0.908 \\ \cmidrule{2-3}
        & 2. Say goodbye to grainy spycam footage . \hl{Tak main lah} video quality Nokia & 0.501 \\  \cmidrule{2-3}
        & 3. adut \: \hl{saya} order 138 utk pastikan bontot staff \hl{saya} 8p m \- 5pm tak ke Pavilion & 0.964 \\ 
        \bottomrule
    \end{tabular}
    \label{RegionStyle}
\end{table}

\subsection{Ablation Study}

We also conduct an ablation study, which aimed to assess the impact of the dual-stage training process on \textsf{ContrastDistAA}. This study involved comparing the model's performance after the initial training phase, which utilized solely contrastive loss, against its performance following the second training stage that integrated both contrastive and disentanglement losses. The results, detailed in last two rows of Table~\ref{tab:ResultAllDatasets}, emphasize the significant contribution of representation disentanglement learning to the model's efficacy.

Crucially, the findings reveal that \textsf{ContrastDistAA} demonstrates an improvement in F1 scores when the disentanglement loss is incorporated in the second training stage, compared to the model trained only with contrastive loss. This improvement underscores the value of the second training stage in enhancing the model's capability. By effectively separating content-related elements from style-related information in the training process, the model becomes more adept at isolating and recognizing distinctive stylistic features inherent to different regional writings. This separation is key to the improved performance, illustrating the effectiveness of the comprehensive two-stage training approach in \textsf{ContrastDistAA}.

\subsection{Qualitative Analysis} 
To demonstrate the efficacy of \textsf{ContrastDistAA}, we employed the t-SNE algorithm \cite{van2008visualizing} to visually represent post style embeddings in two-dimensional space. This visualization aimed to show how different training methodologies influence the distribution of post representations. We selected 100 posts from each region in the Regional Tweets dataset and 50 posts per author from the CCAT50 dataset, extracting their latent representations using three approaches: (i) BERT in its basic form, (ii) a style encoder trained with contrastive loss, and (iii) a style encoder trained using both contrastive loss and mutual information.

Figure \ref{fig:t-SNE visualization} presents these representations. The first three visualizations pertain to posts from the Regional Tweets dataset, while the latter three focus on the CCAT50 dataset. Notably, with the application of contrastive loss, distinct clusters emerge, indicating the style encoder's ability to capture style information effectively. However, challenges are evident, such as the central clustering in Figure \ref{fig:t-SNE visualization} (b), reflecting the limitations of contrastive learning with complex samples. The incorporation of mutual information for disentangling content and style in latent space results in more distinct clustering patterns, as seen in Figure \ref{fig:t-SNE visualization} (c). This suggests that the integration of both contrastive and disentanglement learning notably enhances the style encoder's capability to discern style information, thereby improving its application in AA tasks.

\subsection{Case Studies for Regional AA} 
To highlight the unique writing styles prevalent in different regions, we conducted a linguistic analysis of posts from these areas. This involved selecting three posts from each region and calculating the cosine similarity between their representations and the corresponding regional style representations, providing insights into how closely these posts align with predominant regional writing styles.

Our analysis revealed distinct linguistic features characteristic of each region, often embodied in specific words or expressions that encapsulate regional nuances and evoke emotional responses. For instance, authors from Bangkok frequently conclude sentences with unique words such as ``\textit{kub}'', ``\textit{naka}'', ``\textit{krub}'', or ``\textit{na}'' adding an expressive and emotive quality to their writing. In Jakarta, authors use expressions like "lumayan" to indicate a moderate experience, "seem amat" for excitement, and "mantul" to denote something extraordinary, showcasing the rich and diverse writing style of this region. Hanoi's writing style, influenced by the modern Latin script and its use of diacritical marks, often features Vietnamese words without these marks. This use reflects a blend of traditional and contemporary linguistic practices, allowing for effective communication while honoring the linguistic heritage and subtleties of the region. These findings underscore the distinct linguistic identities of each region, as mirrored in their writing styles.

\section{Summary}

In this chapter, we introduced \textsf{ContrastDistAA}, a model designed to effectively separate content and style information, thereby enhancing AA performance. A significant contribution of our research is the introduction of the regional-level AA task, along with a dedicated dataset to evaluate AA methods in this new context. Through comprehensive experiments, \textsf{ContrastDistAA} was benchmarked against state-of-the-art AA techniques, demonstrating its superior performance in both individual-level and regional-level AA tasks. 

The results from our case studies indicate that \textsf{ContrastDistAA} is adept at identifying unique linguistic features indicative of regional writing styles. Specifically, the contrastive learning and representation disentanglement approach have helped to effectively segregate content from stylistic features for AA tasks. This capability is crucial for understanding how linguistic styles and cultural influences interplay in AA. Our research addresses a previously unexplored aspect of AA and offers fresh perspectives on the relationship between linguistic styles and cultural elements. 

For future work, we will focus on further exploring regional and cultural writing styles. We aim to include a broader range of cultural characteristics and regional diversity, thereby enhancing the understanding of AA in diverse linguistic and cultural contexts. This ongoing research will continue to expand the horizons of AA, contributing to a deeper understanding of the intricate relationship between authorship, language, and culture.

\chapter{InstructAV: Instruction Fine-tuning Large Language Models for Authorship Verification} 

\label{instractav} 

\section{Objective}

Authorship Verification (AV) is a task aimed at determining if two texts were written by the same author, with significant implications across forensics, literature, and digital security domains \citep{av_define, stamatatos2016authorship}. Traditionally, AV relied on stylometric analysis, utilizing linguistic and stylistic features, such as word and sentence lengths, and function word frequencies, to distinguish between authors \citep{seroussi2011authorship,bevendorff2019generalizing}. However, the advent of machine learning, particularly deep learning models like BERT \citep{bert} and RoBERTa \citep{roberta}, has revolutionized this field. These modern approaches, leveraging complex patterns in text, have shown superior performance over conventional stylometric techniques in identifying authorship \citep{saedi2021siamese,konstantinou2022different,graphav}. This paradigm shift underscores a significant evolution in AV methodologies, emphasizing the increasing effectiveness of machine learning in text analysis \citep{zheng2023review}.

While current AV models have made notable advancements, they predominantly focus on binary classification and notably lack in providing explanatory insights. 
Explainability is not only of academic interest. It is fundamental to understanding a model's decision-making logic, and it also enhances trust and reliability in the model's output.

Additionally, the lack of clear explanations makes it hard to find and fix any biases that may be hidden inside these models, creating a significant problem for ensuring they are fair and unbiased.  Therefore, it's critical for AI models not only to be accurate but also to provide transparency and interpretability. 

This chapter presents the \textsf{InstructAV} framework, an innovative approach tailored for AV tasks. Unlike existing models, \textsf{InstructAV} is designed to accurately verify authorship across texts while concurrently furnishing detailed linguistic explanations for its determinations. A key feature of \textsf{InstructAV} is its unique capacity to integrate explainability directly into the classification process, thereby creating a direct pathway between making accurate predictions and offering deep explanations. Through rigorous testing across three diverse AV datasets, the \textsf{InstructAV} framework has demonstrated not only outstanding accuracy in authorship verification but has also set a new benchmark by producing coherent and substantiated explanations for its findings. This dual capability of \textsf{InstructAV}—merging high predictive performance with actionable insights—marks a significant leap forward in the AV domain, contributing both to the enhancement of model transparency and the advancement of explainable artificial intelligence.

Our contributions can be summarized as follows: (i) We propose the \textsf{InstructAV} framework for AV tasks to accurately determine whether two texts share the same author and to furnish robust linguistic explanations for the AV outcomes. (ii) We have curated three instruction-tuning datasets, each accompanied by dependable linguistic explanations for AV tasks. These datasets are intended to serve as valuable resources for advancing research in this field\footnote{\textcolor{black}{The
code and datasets can be found at~\url{https://github.com/Social-AI-Studio/InstructAV}.}}. (iii) Both automated and human evaluation results demonstrate the effectiveness of the \textsf{InstructAV} in providing precise AV predictions and reliable linguistic explanations.

\section{Methodology}

\begin{figure*}[t] 
    \centering
        \includegraphics[width=\textwidth]{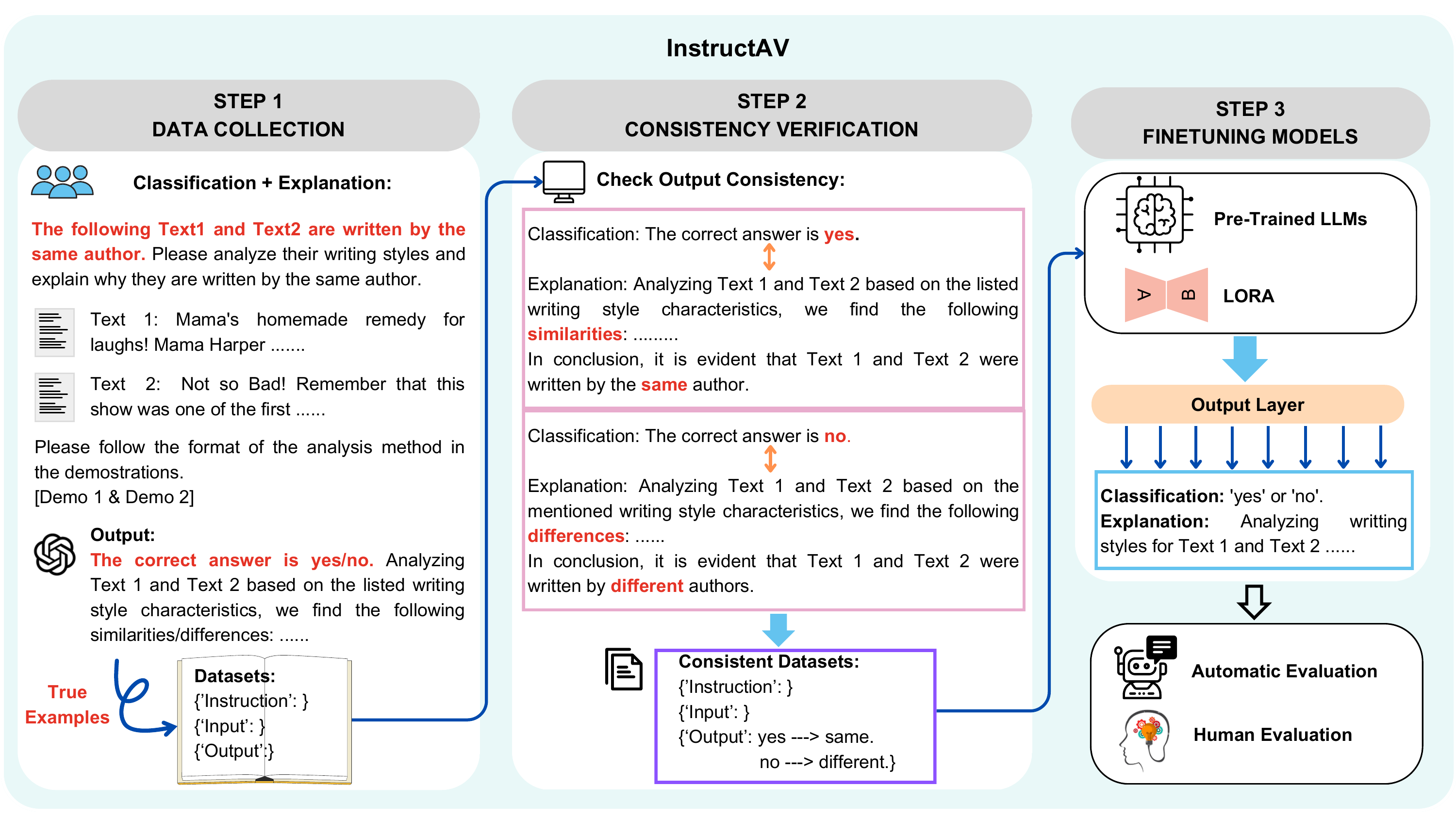}
    \caption{An illustration of \textsf{InstructAV}'s architectures.}
    \label{fig:framework}
\end{figure*}

The overview of the \textsf{InstructAV} framework, as depicted in Figure \ref{fig:framework}, comprises three primary steps: data collection, consistency verification, and the fine-tuning of LLMs using the LoRA method \citep{hu2021lora}. Initially, the framework focuses on the aggregation of explanatory data for AV samples. 
This approach uses the binary classification labels available in existing AV datasets.  Following this, a strict quality check is implemented, aimed at verifying the alignment and consistency of the explanations with the corresponding classification labels. The final stage involves the synthesis of instruction-tuning data, which is a fusion of the gathered classification labels and their associated explanations. 
This composite data then serves as the foundation for fine-tuning LLMs in conjunction with the LoRA adaptation technique. This approach ensures that the LLMs are not only accurately fine-tuned for the AV task but also enhanced in their capacity to provide coherent and reliable explanations for their predictions. The subsequent sections present the details of each component in \textsf{InstructAV}.

\subsection{Explanation Data Collection}
\label{datasets}

To augment the explanatory capabilities of the \textsf{InstructAV} model, particularly in generating dependable explanations for AV predictions, we initiated a comprehensive collection of explanations using ChatGPT, with the crucial step of informing ChatGPT about the classification labels beforehand. This process involved three datasets widely used in AV studies: the IMDB dataset~\citep{imdb}, the Twitter dataset~\citep{schwartz2013authorship}, and the Yelp Reviews dataset\footnote{https://www.yelp.com/dataset}. This selection strategically covers various dimensions of textual data, thereby ensuring a diverse and comprehensive analysis.

The IMDB dataset, characterized by its longer text length (averaging 303 words), exemplifies long-form content. In contrast, the Twitter dataset, with an average text length of 16 words, epitomizes short-form content. The Yelp Reviews dataset, averaging 154 words per text, represents medium-length content. These datasets, initially curated for authorship attribution tasks, were adapted for our study. 
We extracted 11,000 samples from each dataset to facilitate a robust AV evaluation. Each sample comprises two texts, either written by the same author or by different authors.

For generating explanations, we employed ChatGPT with true classification labels and few-shot prompts, focusing on 11 linguistic features for each AV sample. These features, identified in the research by \citep{boenninghoff2019explainable}, are crucial for analyzing writing styles in textual content. The linguistic features generated by known-label ChatGPT were recorded as labels for the AV explanations in our dataset if they accurately predict and meet the consistency check.

To link classification labels closely with linguistic explanations during the data collection phase, we crafted prompts that incorporate the classification labels. For example, a prompt might state ``\textit{The following Text1 and Text2 are written by different authors. Please analyze their writing styles and explain why they are written by different authors.}'' to guide the explanation process. A detailed example of the explanation generation prompt is shown in Appendix~\ref{app:prompt} Table~\ref{tab:prompt_example}. These prompts enhance the relevance of the collected explanations, thereby improving the explanatory capacity of the \textsf{InstructAV} model.

\paragraph{Consistency Verification.} 
Alignment between classification labels and their linguistic explanations is essential for explanation data integrity and trustworthiness. 
Models like known-label ChatGPT are skilled at generating explanations but face challenges in align explanations with classification labels. For instance, despite being informed that "\textit{The following Text1 and Text2 are written by the same author,}" ChatGPT might incorrectly respond with,  "\textit{The correct answer is no.}"  Instances like this create a mismatch between explanations and classification labels, reducing trust in the model's explanations. To enhance user trust in automated decisions, it is important to guarantee the consistency and reliability of both classifications and explanations.
As shown in Figure \ref{fig:framework}, we employ a consistency verification method to verify the alignment between the model’s analytical explanations and its classification decisions. We have instituted a comprehensive verification process to ensure the explanations are consistent with the classification.  This process leverages demonstration templates that inherently guide the model to incorporate specific expressions—such as `\textit{same/similarities}' or `\textit{different/differences}'—into its outputs. 
\textcolor{black}{During the consistency verification stage, keyword searching was performed on terms within the generated text. Matching the phrases `\textit{written by the same author}' (resp. `\textit{written by different authors}') with classification labels allows us to assess the quality of ChatGPT's linguistic explanations.}

The key to this process is the construction of instruction-tuning data, which serves a dual purpose: facilitating AV classification prediction and supporting the generation of explanations. An example of the instruction-tuning data is shown in Appendix~\ref{app:datasets} Table~\ref{tab:dataset_example}. This verification step is essential for ensuring that the \textsf{InstructAV} model is not only accurate in its predictions but also capable of generating explanations that are coherent, relevant, and aligned with the classification outcomes, thereby enhancing the overall efficacy and reliability of the \textsf{InstructAV} framework.

\subsection{Fine-tuning with LoRA}

Adapting LLMs for AV tasks can be a demanding process, particularly due to the significant computational resources and extensive labeled data typically required for fine-tuning. To address this challenge, we have incorporated a PEFT method known as LoRA~\citep{hu2021lora} into our approach for adapting LLMs to AV tasks.

LoRA presents a novel approach to updating weight matrix parameters. It decomposes the matrix into two low-rank matrices, reducing trainable parameters while processing high-dimensional matrices effectively. During the forward pass, LoRA matrices compute rank-decomposed weights, used in attention or feed-forward activations. Importantly, this method keeps the LLM's core parameters unchanged, enabling adaptation to new data distributions or tasks without overhauling the model completely.

\begin{table}[htb]
\caption{Dataset statistics for IMDB, Twitter and Yelp.}
    \small
    \centering
    \begin{tabular}{c c c c c} 
 \toprule
\textbf{Dataset}& \#Authors& \textbf{\#Train} & \textbf{\#Test} & \textbf{Avg length}\\
 \hline
 IMDB & 64& 10,000 & 1,000 & 303 \\
 Twitter & 100& 10,000 &  1,000 & 16 \\
 Yelp & 1,000& 10,000 & 1,000 & 154 \\
\bottomrule
\end{tabular}
    
    \label{tab:data_stat}
\end{table}

The LoRA architecture is encapsulated in the following equation:
\begin{equation}
    h = W_{0}x + \Delta W_{0}x = W_{o}x + BA x
\end{equation}
where $W_{0}\in \mathbb{R}^{d \times d}$ is a pre-trained weight matrix in LLMs, which is associated with two adjustable low-rank matrices, $B \in \mathbb{R}^{d \times r}$ and $A \in \mathbb{R}^{r \times d} $, $d$ corresponds to the hidden dimension of the attention layers, and $r$ is the adaptation rank, which is selected such that $r \ll d$. Within the training phase, $W_{0}$ remains frozen, exempt from gradient updates, whereas $A$ and $B$ are dynamic, containing the parameters subject to training.

\section{Experiments}

\subsection{Experiment Settings}

\textbf{Datasets.} The \textsf{InstructAV} framework evaluation utilized three distinct Authorship Verification (AV) datasets: IMDB62~\citep{imdb}, Twitter~\citep{schwartz2013authorship}, and Yelp Reviews~\footnote{https://www.yelp.com/dataset}, chosen for diversity and relevance.

To evaluate the AV classification and explanation tasks performance, we constructed two distinct types of dataset settings, each incorporating varying levels of information:

\begin{enumerate}
    \item \textit{Classification}: This dataset involves integrating a question alongside two texts as the input and employs the LoRA method to fine-tune the model for classification tasks. The expected output is a straightforward binary classification indicating whether the two texts are written by the same author, formatted as ``\textit{The correct answer is yes/no}''. 
    \item \textit{Classification and Explanation}: In this setting, we augment the classification data with linguistic analysis to empower the model to generate robust explanations for the AV predictions. The LLMs are fine-tuned to not only predict the classification labels, but also provide an analysis of eleven linguistic features of the two texts. This added layer of analysis offers a reasoned explanation behind the classification decision, thereby enhancing the model's interpretability and reliability.
\end{enumerate}

For the \textit{Classification} setting, We randomly sampled 11,000 balanced samples from the IMDB62, Twitter, and Yelp datasets, respectively. The data was then divided into a training set comprising 10,000 samples and a test set comprising 1,000 samples, with both sets maintaining balanced class distributions. 

For the \textit{Classification and Explanation} setting, we initially sampled 20,000 examples from each dataset and used the GPT-3.5-turbo API for linguistic analysis. To ensure the high quality of the generated explanations, samples with incorrect linguistic analysis were dropped during the initial consistency verification phase. For a comprehensive evaluation, balanced subsets consisting of 10,000 training samples and 1,000 testing samples were randomly selected from the verified examples within each subset. These samples formed instruction-tuning data, including text and corresponding linguistic explanations (see Section \ref{datasets}). Details regarding dataset statistics and characteristics, including the splits between training and test sets, are presented in Table \ref{tab:data_stat}.

These two dataset settings enable a thorough investigation into how the addition of explanatory components influences the performance of the classification task, providing insights into the efficacy and adaptability of the \textsf{InstructAV} framework.

\noindent \textbf{Baselines.} For the \textit{Classification} task, our baseline models comprised BERT~\citep {bert} and its variants, DistilBERT~\citep{sanh2019distilbert}, and AlBERT~\citep{lan2019albert}. These models were selected for their widespread usage and proven effectiveness in AV classification tasks~\citep{brad2021baseline,tyo2022state,fabien2020bertaa}. However, these models are not inherently designed for text generation tasks.
\textcolor{black}{Another baseline related to LLMs is the application of the LIP method proposed by ~\citet{huang2024largelanguagemodelsidentify} on GPT-4-turbo, LLaMA-2-70B, and Mistral-7B.}

For tasks involving \textit{Classification and Explanation}, autoregressive models like GPT, which generate text sequences based on preceding text, are suitable. We adopted the PromptAV methodology on GPT-3.5-Turbo as a baseline for the ChatGPT model, as proposed by~\citep{promptav}. Although PromptAV may be slightly less effective than BERT in classification accuracy, it offers advantages such as bypassing extensive training and generating explanations. Additionally, we integrated PromptAV, using GPT-4 as the foundational model, to establish a more competitive baseline.

Our comparative analysis evaluates both classification accuracy and the quality of explanations from \textsf{InstructAV} and PromptAV. This comprehensive assessment aims to understand the strengths and potential limitations of each approach in AV tasks.

\noindent \textbf{Implementation.} In our research, we employed three different LLMs to validate the efficacy of the \textsf{InstructAV} framework. These models include OPT-6.7B~\citep{zhang2022opt}, LLaMA-1-7B~\citep{touvron2023llama}, and LLaMA-2-7B~\citep{touvron2023llama2}. Our experiments were conducted using two NVIDIA A6000 GPUs, encompassing both the fine-tuning and inference phases.

For fine-tuning, we applied the LoRA approach with a rank set to 8. This introduced 4.1 million parameters, representing only 0.06\% of the total 7 billion parameters of the base models. This parameter increase highlights LoRA's effectiveness in offering an efficient fine-tuning method. Fine-tuning was performed over 3 epochs using the LLM-Adapters Toolkit~\citep{llm-adapter}, tailored for integrating adapters into LLMs.

During inference, all operations were performed deterministically with a fixed temperature of 0.1. This maintains consistency and reliability in the model's performance, crucial for accurately assessing the \textsf{InstructAV} framework's capabilities. The selection of diverse LLMs and the strategic application of LoRA for fine-tuning effectively showcase the capability of the \textsf{InstructAV} framework across various model architectures.
\textcolor{black}{To ensure the reproducibility of the experiments conducted using the \textsf{InstructAV} framework, each experiment was repeated three times with different random seeds. The mean and standard deviation were calculated from these repetitions.}

\subsection{Evaluation Metrics}
The evaluation of the \textsf{InstructAV} framework involves both AV classification and explanation generation tasks. This dual-focused approach enables an in-depth assessment of the framework’s capabilities in key areas of AV.

For AV classification tasks, accuracy serves as the primary metric. This metric effectively measures the models' proficiency in determining whether two texts have been written by the same author. 
To evaluate the framework's performance in generating explanations for AV tasks, a distinct approach is required. Here, the focus shifts towards assessing the quality of the linguistic analysis generated by the models. Unlike accuracy, which is quantitatively measurable, the quality of linguistic explanations presents a more complex evaluation challenge, as it involves subjective elements related to the coherence, relevance, and clarity of the explanations. Given the complex nature of evaluating explanation quality, we adopt a dual evaluation approach, employing both automatic and human evaluation metrics. 

\textbf{Automatic Evaluation for Explanations.} In our automated assessment of explanation quality for both \textsf{InstructAV} models and baseline models, we measure the similarity between generated and labeled explanations, using ChatGPT-generated texts as standards. Through consistency verification, we ensure explanatory labels within test sets consistently align with classification labels, affirming their accuracy. Consequently, we consider ChatGPT's linguistic analysis as \textcolor{black}{explanation labels}. We employ metrics like ROUGE-1, ROUGE-2, ROUGE-L (for content coverage and structural fluency), and BERT\_Score (for semantic quality). Higher scores indicate better explanation quality, ensuring a comprehensive evaluation based on content accuracy, logical coherence, and contextual relevance.

\textbf{Human Evaluation for Explanations.}
To supplement the automated metrics for evaluating explanation quality in the \textsf{InstructAV} framework, we also implemented a human evaluation. This approach involved randomly selecting 100 examples from each of the IMDB, Twitter, and Yelp Reviews datasets for qualitative assessment by human evaluators. These evaluations included not only explanations generated by \textsf{InstructAV} and baseline models, but also those explanation labels. For the human evaluations focusing on explanations, we selected InstrctAV (LLaMA-2-7B) along with the two most robust baselines, namely PromptAV-2shot equipped with GPT-3.5 and GPT-4 as the base models. Three evaluators were enlisted to independently evaluate each explanation across four criteria:

\textbf{Coverage}: Evaluating how many of the anticipated linguistic features are present in the explanations.
Evaluators checked the number of features covered in each explanation, with baseline models like PromptAV covering 0 to 7 features, and \textsf{InstructAV}, along with \textcolor{black}{explanation labels}, extending from 0 to 11 features.

\textbf{Relevance}: Evaluators scored the relevance of the explanations to the original texts on a 5-point Likert scale, ranging from 1 (completely irrelevant) to 5 (perfectly relevant).
    
\textbf{Reasonableness}: This involved assessing the logical soundness of each feature's explanation, using a 5-point Likert scale, from 1 (totally unreasonable) to 5 (completely reasonable).

\textbf{Persuasiveness}: Evaluators rated the persuasiveness of the explanations on a scale from 1 (completely not persuasive) to 5 (highly persuasive).

\section{Experiment Results}

\subsection{Classification Results.}

\begin{table*}[htb]
\caption{\textcolor{black}{Classification Accuracy of \textsf{InstructAV} and the baselines on different dataset settings. Highest Acc are \textbf{bolded}. All experiments were repeated three times. The table lists the average values of the three repetitions, while the standard deviations of accuracy from the experiments are provided in brackets.}}
    \small
    \centering
    \setlength{\tabcolsep}{3pt}
    \begin{tabular}{c l c c c} 
    \hline
    \textbf{Dataset Setting} &\textbf{Model} &\textbf{IMDB}&\textbf{Twitter}&\textbf{Yelp}\\
     \hline
    \multirow{9}*{Classification} & BERT &  0.677 (0.0124) & 0.702 (0.0021) & 0.622 (0.0020) \\
     & DistilBERT &  0.526 (0.0125) & 0.575 (0.0065) & 0.543 (0.0036) \\
     & ALBERT & 0.642 (0.0040) & 0.701 (0.0023) & 0.601 (0.0020) \\
     & LIP (GPT-4-turbo) & 0.732 (0.0070) & 0.612 (0.0040) & 0.632 (0.0036) \\
     & LIP (LLaMA-2-70B) & 0.533 (0.0080) & 0.554 (0.0042) & 0.528 (0.0040) \\
     & LIP (Mistral-7B) & 0.507 (0.0038) & 0.539 (0.0025) & 0.527 (0.0032) \\
     & InstructAV (LLaMA-1-7B) & 0.648 (0.0236) & 0.610 (0.0062) & 0.542 (0.0031)\\
    & \textsf{InstructAV} (OPT-6.7B) & 0.590 (0.0050) & 0.524 (0.0110) & 0.527 (0.0060)  \\ 
    & \textsf{InstructAV} (LLaMA-2-7B) &  \textbf{0.914} (0.0046) & \textbf{0.740} (0.0070) & \textbf{0.689} (0.0025)\\
     \hline
    \multirow{6}*{Classification \& } & PromptAV-2shot (GPT-3.5) &  0.623 (0.0397) & 0.628 (0.0147) & 0.534 (0.0064) \\
    & PromptAV-4shot (GPT-3.5)& 0.635 (0.0265) & 0.667 (0.0163) & 0.544 (0.0080) \\
    & PromptAV-8shot (GPT-3.5)& 0.601 (0.0070) & 0.648 (0.0075) & 0.564 (0.0081)\\
    & PromptAV (GPT-4-Turbo) & 0.755 (0.0075) & 0.729 (0.0070) & 0.597 (0.0065) \\
    Explanation & \textsf{InstructAV} (LLaMA-1-7B) & 0.825 (0.0289) & 0.625 (0.0065) & 0.596 (0.0104) \\
    & \textsf{InstructAV} (OPT-6.7B) & 0.744 (0.0095) & 0.714 (0.0070) & 0.575 (0.0140)\\
    & \textsf{InstructAV} (LLaMA-2-7B) & \textbf{0.937} (0.0017) & \textbf{0.745} (0.0063) & \textbf{0.693} (0.0442)
 \\
\hline
\end{tabular}
    \label{tab:performance}
\end{table*}

We evaluate the \textsf{InstructAV} framework for AV tasks, utilizing both classification and explanatory dataset settings. The primary goal is to explore how linguistic analysis can improve the model's performance in AV classification tasks. The results, detailed in Table \ref{tab:performance}, include the accuracy of \textsf{InstructAV} with various LLMs and baseline models. A key finding is that \textsf{InstructAV}, particularly when paired with LLaMA-1-7B and LLaMA-2-7B, outperforms baseline models in all datasets when using only classification data. Notably, with LLaMA-2-7B, \textsf{InstructAV} achieves a 25.2\% improvement over the highest-performing baseline, BERT, on the IMDB dataset.

It's imperative to underline that in our experimental setup, models were evaluated on their ability to concurrently generate classification predictions and conduct linguistic analysis without incorporating explanation data in the input. 
This methodology ensured fair comparison with approaches that focus only on classification.
Our results clearly demonstrate that all variants of the \textsf{InstructAV} framework, utilizing different LLMs, significantly benefit from training on explanatory labels. 
Notably, \textsf{InstructAV} with LLaMA-2-7B showcased a remarkable 27.1\% improvement in classification accuracy over the PromptAV approach using 2-shot prompts based on GPT-3.5. Furthermore, when compared to PromptAV-2shot employing GPT-4, \textsf{InstructAV} exhibited superior performance across all three evaluated datasets. 
These findings highlight the substantial benefits of incorporating explanatory training, indicating that providing \textsf{InstructAV} with the dual function of AV classification and generating linguistic explanations significantly boosts its classification precision.

\subsection{Automatic Evaluation Results on Explanations.}

\begin{table}[htb]
\caption{Automatic Metric Performance for \textsf{InstructAV} and PromptAV. Higher
Acc are \textbf{bolded}. }
    \centering
    \small
    \bgroup
    \def\arraystretch{1.3}
    \begin{tabular}{l c c c c } 
 \hline
 \multicolumn{5}{c}{\textbf{IMDB}} \\
 \hline
 \textbf{Model} & ROUGE-1 & ROUGE-2 & ROUGE-L & Bert\_Score \\
 \hline
PromptAV-2shot (GPT-3.5) & 0.379 & 0.147 & 0.227 & 0.844 \\
PromptAV (GPT-4) & 0.496 & 0.193 & 0.237 & 0.861 \\
\textsf{InstructAV} (LLaMA-1) & 0.677 & 0.412 & 0.496 & 0.898 \\
\textsf{InstructAV} (OPT) & 0.656 & 0.403 & 0.482 & 0.893 \\
\textsf{InstructAV} (LLaMA-2) & \textbf{0.689} & \textbf{0.434} & \textbf{0.515} & \textbf{0.907} \\
\hline
\multicolumn{5}{c}{\textbf{Twitter}} \\ 
\hline
PromptAV-2shot (GPT-3.5) & 0.445 & 0.203 & 0.287 & 0.856 \\
PromptAV (GPT-4) & 0.510 & 0.307 & 0.466 & 0.860\\
\textsf{InstructAV} (LLaMA-1) & 0.670 & 0.406 & 0.522 & 0.899\\
\textsf{InstructAV} (OPT) & 0.644 & 0.398 & 0.476 & 0.897 \\
\textsf{InstructAV} (LLaMA-2) & \textbf{0.689} & \textbf{0.420} & \textbf{0.542} & \textbf{0.904}\\
\hline
\multicolumn{5}{c}{\textbf{Yelp}} \\ 
\hline
PromptAV-2shot (GPT-3.5) & 0.377 & 0.142 & 0.243 & 0.856 \\
PromptAV (GPT-4) & 0.431 & 0.162 & 0.402 & 0.867 \\
\textsf{InstructAV} (LLaMA-1) & 0.666 & 0.407 & 0.524 & 0.906\\
\textsf{InstructAV} (OPT) & 0.629 & 0.377 & 0.486 & 0.900\\
\textsf{InstructAV} (LLaMA-2) & \textbf{0.716} & \textbf{0.429} & \textbf{0.592} & \textbf{0.912} \\
 \hline
\end{tabular}
\egroup
    \label{tab:auto_eval}
\end{table}

Our \textsf{InstructAV} framework demonstrates that training in \textcolor{black}{explanation labels} can significantly improve AV classification performance. Additionally, we aim to assess whether \textsf{InstructAV} can generate high-quality explanations. 
For this purpose, we selected PromptAV-2shot (GPT-3.5), the best-performing variation of the framework, alongside GPT-4, to benchmark against all variations of \textsf{InstructAV}. we subjected them to both automatic and human evaluations focused on the quality of their generated linguistic explanations. The results of the automatic evaluation are presented in Table \ref{tab:auto_eval}. These results reveal that \textsf{InstructAV} consistently surpasses PromptAV models (GPT-3.5 and GPT-4) across all datasets and all evaluation metrics. Notably, ROUGE-1 and ROUGE-2 scores highlight \textsf{InstructAV}'s superior performance in achieving content overlap at both the word and phrase levels. Moreover, the ROUGE-L metric indicates that \textsf{InstructAV} is more proficient in maintaining sentence-level structure and fluency. Furthermore, the BERT\_Score supports the observation that the explanations generated by \textsf{InstructAV} are semantically closer to the \textcolor{black}{explanation labels}. This comprehensive evaluation underscores \textsf{InstructAV}'s capability not only in improving AV classification accuracy but also in generating linguistically coherent and contextually relevant explanations.

\subsection{Human Evaluation Results on Explanations.}

\begin{table}[htb]
\caption{\textcolor{black}{Results of the human evaluation on explanation labels and explanations generated by the \textsf{InstructAV} and PromptAV frameworks.}}
    \small
    \centering
    \begin{tabular}{l c c c c} 
 \hline
 \multicolumn{5}{c}{\textbf{IMDB}} \\
 \hline
 \textbf{Model} & Coverage & Relevance & Reasonableness & Persuasiveness  \\
 \hline
\textcolor{black}{Explanation Label} & 11 & 4.42 & 4.12 & 4.22 \\
PromptAV-2shot (GPT-3.5) & 7 & 3.8 & 3.84 & 3.26 \\
PromptAV (GPT-4) & 7 & 3.92 & 3.87 & 3.56 \\
\textsf{InstructAV} (LLaMA-2-7B) & 11 & 4.24 & 3.98 & 4.03 \\
\hline
\multicolumn{5}{c}{\textbf{Twitter}} \\
\hline
\textcolor{black}{Explanation Label} & 11 & 4.66 & 4.42 & 4.65\\
PromptAV-2shot (GPT-3.5) & 7 & 3.01 & 3.24 & 2.94\\
PromptAV (GPT-4) & 7 & 3.67 & 3.22 & 3.27\\
\textsf{InstructAV} (LLaMA-2-7B) & 11 & 4.48 & 4.25 & 4.28\\
\hline
\multicolumn{5}{c}{\textbf{Yelp}} \\
\hline
\textcolor{black}{Explanation Label} & 11 & 4.56 & 4.44 & 4.65\\
PromptAV-2shot (GPT-3.5) & 7 & 2.87 & 3.02 & 3.06\\
PromptAV (GPT-4) & 7 & 3.75 & 3.41 & 3.50 \\
\textsf{InstructAV} (LLaMA-2-7B) & 11 & 4.3 & 4.32 & 4.36\\
 \hline
\end{tabular}
    
    \label{tab:human_eval}
\end{table}

To comprehensively evaluate the generated explanations and to assess the quality of the linguistic analysis produced by ChatGPT, which serves as our \textcolor{black}{explanation labels}, we have conducted a human evaluation using four key metrics: Coverage, Relevance, Reasonableness, and Persuasiveness. The results of this human evaluation are presented in Table \ref{tab:human_eval}.

Our findings from this human evaluation process show that our \textcolor{black}{explanation labels} achieves the highest scores. This result validates our methodological choice of using explanations generated by known-label ChatGPT as both a source for training data and a benchmark for \textcolor{black}{explanation labels} in our testing scenarios. 
Importantly, the results also reveal that \textsf{InstructAV}, particularly with the LLaMA-2-7B model, not only surpasses the performance of PromptAV-2shot models (GPT-3.5 and GPT-4) but also attains a level of explanation quality that is comparable to known-label ChatGPT. This outcome is significant as it demonstrates that \textsf{InstructAV} can produce explanations that are not only accurate but also contextually relevant, logically sound, and convincing to human evaluators. Such a capability is essential for applications where understanding the rationale behind model predictions is as important as the predictions themselves.

\begin{table}[t]
\caption{Performance of \textsf{InstructAV} on Top 25 and Bottom 25 explanation. Higher
accuracies are bolded.}
    \small
    \centering
    \begin{tabular}{c c c c c} 
 \hline
  &\textbf{IMDB}&\textbf{Twitter}&\textbf{Yelp}\\
 \hline
Top 25 & \textbf{0.92} & \textbf{0.8} & \textbf{0.84} \\
Bottom 25 & 0.88 & 0.68 & 0.72 \\
 \hline
\end{tabular}
    \label{tab:corr_eval}
\end{table}

\subsection{Correlation between Explanation and Classification.}
To explore the relationship between explanation quality and classification accuracy, we have selected two distinct subsets of \textsf{InstructAV} samples: the top 25\% with the highest average human evaluation scores and the bottom 25\% with the lowest average human evaluation scores. We then calculated the classification accuracy for each of these subsets, and the results are presented in Table \ref{tab:corr_eval}. Our analysis clearly indicates that samples associated with higher quality explanations consistently achieve superior classification accuracy. 
This finding underscores the effectiveness of training \textsf{InstructAV} to not only generate AV classification predictions but also provide linguistic explanations. Such training not only enhances the model's performance in AV classification tasks but also enhances its ability to produce valuable linguistic analysis.

\section{Ablation Study}

\begin{table*}[htb]
\caption{Classification Performance of InstructAV Framework and LLMs in Few-Shot Contexts Without Fine-Tuning. Highest Acc are \textbf{bolded}.}
    \small
    \centering
    \color{black}
    \begin{tabular}{ l l  c  c  c} 
    \hline
      & \textbf{Model} & \textbf{IMDB}&\textbf{Twitter}&\textbf{Yelp}\\
     \hline
     \multirow{3}*{0-shot} & LLaMA-1-7B & 0.000 & 0.000 & 0.000\\
     & OPT-6.7B & 0.000 & 0.000 & 0.000  \\ 
      & LLaMA-2-7B &  0.006 & 0.010 & 0.021\\
    \hline
     \multirow{3}*{2-shot} & LLaMA-1-7B & 0.189 & 0.097 & 0.226\\
      & OPT-6.7B & 0.258 & 0.179 & 0.297  \\ 
      & LLaMA-2-7B &  0.309 & 0.336 & 0.397\\
    \hline
     \multirow{3}*{4-shot} & LLaMA-1-7B & 0.014 & 0.020 & 0.127\\
      & OPT-6.7B & 0.191 & 0.220 & 0.217  \\ 
      & LLaMA-2-7B &  0.279 & 0.350 & 0.375\\
    \hline
     \multirow{3}*{8-shot} & LLaMA-1-7B & 0.003 & 0.007 & 0.002 \\
      & OPT-6.7B & 0.005 & 0.008 & 0.003  \\ 
      & LLaMA-2-7B) &  0.020 & 0.033 & 0.025
\\
     \hline
    \multirow{3}*{Finetuned} & \textsf{InstructAV} (LLaMA-1-7B) & 0.648  & 0.610 & 0.542)\\
    & \textsf{InstructAV} (OPT-6.7B) & 0.590 & 0.524 & 0.527  \\ 
    & \textsf{InstructAV} (LLaMA-2-7B) &  \textbf{0.914} & \textbf{0.740} & \textbf{0.689} \\
\hline
\end{tabular}
    \label{tab:ablation}
\end{table*}

\textcolor{black}{Ablation experiments  were conducted to investigate potential data contamination issues, specifically whether the data might have been included in the training corpora of LLMs. We performed 0-shot, 2-shot, 4-shot, and 8-shot in-context tests on each dataset using the original, untuned LLaMA-2-7B, LLaMA-1-7B, and OPT-6.7B models. Each experiment was replicated three times, with the mean results presented in Table~\ref{tab:ablation}.
The results indicate that the original models struggle to perform AV tests effectively on our dataset, especially in the 0-shot setting where the outputs are highly random. Both LLaMA-1-7B and OPT-6.7B failed to make correct judgments, and LLaMA-2-7B showed weak judgment capabilities. 
Consequently, we postulate that it is unlikely that our dataset was included in the training data of the original models.}

\section{Summary}

This chapter presents \textsf{InstructAV}, an innovative approach to AV tasks that leverages LLMs with a PEFT method. Our study establishes \textsf{InstructAV} as a significant advancement in the AV domain, showcasing its ability to enhance classification accuracy and provide clear and coherent explanations for its decisions. The contributions of this paper, including the development of the \textsf{InstructAV} framework, the creation of three instruction-tuning datasets with reliable linguistic explanations, and the demonstration of the framework's effectiveness through both automated and human evaluations, mark a crucial progress in AV research. \textsf{InstructAV}, with its dual priority on high accuracy and the ability to provide high quality explanations, positions it as a state-of-the-art AV solution.

\chapter{Conclusion and Future Work} 

\label{conclusion} 

\section{Conclusion}

This dissertation has explored innovative methodologies for learning text styles, focusing on three areas: text style transfer, authorship attribution, and authorship verification. Through four foundational works, we have advanced the field in several key areas.

\textbf{Parameter-Efficient Fine-tuning Toolkit.} In the realm of parameter-efficient fine-tuning, we introduced the LLM-Adapters framework, which facilitates the integration of various adapters into large language models. This approach not only elevates the fine-tuning efficiency of these models but also democratizes their application across a wide array of tasks by reducing the computational resources required. Furthermore, we have conducted a comprehensive empirical study examining various PEFT methods within different open-source LLMs. We propose the LLM-Adapters framework as a foundational work to facilitate the exploration of parameter-efficient fine-tuning methods for text style transfer.

\textbf{Parameter-Efficient Framework for TST.} Building on this, we introduce \textsf{Adapter-TST}, a parameter-efficient framework designed to perform multi-attribute text style transfer tasks while significantly reducing computational resource requirements. The \textsf{Adapter-TST} framework offers two configurations: \textit{parallel}, which supports multiple-output text style transfer, and \textit{stacking}, which allows for compositional editing. Comprehensive experiments conducted on real-world datasets reveal that \textsf{Adapter-TST} outperforms state-of-the-art baselines in both transferring styles and generating high-quality textual output, as evidenced by results from both automated and human evaluations.

\textbf{Disentangled Representation Learning for AA.} In authorship attribution, we tackled the challenging issue of topic variability with the \textsf{ContrastDistAA} framework. This method leverages contrastive learning and mutual information maximization, leading to powerful disentanglement of content and stylistic features. This approach achieves unparalleled accuracy in AA tasks, particularly in contexts involving topic shifts. Through extensive experimentation, we have benchmarked \textsf{ContrastDistAA} against current state-of-the-art AA methods across various scenarios. The results unequivocally demonstrate \textsf{ContrastDistAA}'s superior performance in both individual and regional-level AA tasks, underscoring its effectiveness and adaptability.


\textbf{Explanability for AV.} For authorship verification, the \textsf{InstructAV} framework was developed. This framework sets a new standard by not only delivering accurate authorship verification but also providing coherent and detailed explanations for its predictions, thereby enhancing transparency and accountability in model outputs. We have curated three instruction-tuning datasets, each supplemented with reliable linguistic explanations relevant to AV tasks. Evaluation results from both automated methods and human reviewers confirm the efficacy of \textsf{InstructAV} in providing precise AV predictions and robust linguistic explanations.

Each of these contributions underscores the potential for large language models to address complex problems in text style analysis, while our advancements in dataset creation and model framework offer significant resources for future research in computational linguistics.

\section{Future Work}


While the advancements in this thesis have laid substantial groundwork, several avenues for future research can further extend these achievements.

In the domain of parameter-efficient fine-tuning, the effectiveness of PEFT is often contingent upon several hyperparameters, including the bottleneck dimension of adapters, the rank of LoRA, and the configuration of various additive PEFT layers. Manually adjusting these hyperparameters demands considerable effort. Therefore, future research should concentrate on developing approaches that reduce reliance on manual tuning or that can automatically determine optimal configuration settings. Additionally, investigating scaling laws represents another promising research avenue. Since PEFT methods initially formulated for smaller Transformer models do not necessarily scale effectively with larger models, it is imperative to identify and adapt strategies that maintain their efficacy as foundation models expand in size. Such investigations will facilitate the customization of PEFT methodologies to accommodate the evolving landscape of large model architectures.

In the domain of text style transfer, the progression of research has stagnated or decelerated, primarily due to the absence of robust evaluation metrics. Addressing this deficiency constitutes a critical research direction for TST and is essential for achieving significant breakthroughs in the field. For instance, employing large language models to evaluate transfer accuracy and content preservation could result in evaluations that better reflect human preferences. Moreover, another crucial avenue for TST research involves broadening the spectrum of styles under investigation. Present studies primarily concentrate on text attributes like sentiment and formality. However, with advancements in large language models, researchers can delve into more intricate styles, such as emulating the writing style of particular individuals or harmonizing the writing styles of different co-authors within collaborative works.

In the fields of authorship attribution and authorship verification, a promising future research direction is code reuse detection combined with dynamic author profiling. Authors of both malware and legitimate software frequently reuse existing software codes and libraries to save time and effort, given that these components are already developed and available. Dynamic author profiling methods can be adapted for authorship attribution by monitoring changes in an author's coding style over time. Thus, identifying code reuse, employing clone detection techniques, and implementing dynamic profiling of authors represent potential avenues for future research. Another significant direction is multi-author attribution, which centers on determining the specific authorship of program code or documents attributed to multiple authors. A notable challenge in multi-author attribution lies in delineating code boundaries among different authors within a sample code and correctly identifying segments that belong to each author. Exploring the development of models for multi-author attribution that can accurately and effectively identify code segments attributed to individuals among multiple authors presents another fruitful area for future research.

These future directions not only promise to deepen our current understanding but also hold potential for significant impact across various technological and human-centric fields.


\appendix 


\chapter{Appendix of LLM-Adapters}
\label{appendixa} 

\section{Math Reasoning Prompt Templates}
\label{sec:math_templates}
We utilize ChatGPT to collect the math reasoning data for fine-tuning. Table \ref{tab:p_math} show the prompt template used to query ChatGPT. The expression "Please give the steps" is employed to guide ChatGPT to generate reasoning steps, thus, we can use the rationale information to fine-tune adapters. "Give the arabic numerals as the answer." is utilized to guide ChatGPT to generate arabic numbers as the final answer making it easier to extract the answer from the outputs.

\begin{table*}[t]\centering
\small
\begin{tabular}{lp{11cm}}\hline
Template &  [QUESTION] Please give the steps and the arabic numerals as the answer. \\
\hline
Prompt & Albert is wondering how much pizza he can eat in one day. He buys 2 large pizzas and 2 small pizzas. A large pizza has 16 slices and a small pizza has 8 slices. If he eats it all, how many pieces does he eat that day? 

Please give the steps and the arabic numerals as the answer. \\ \hline

ChatGPT Output & To solve the problem, we need to first find out how many slices are in 2 large pizzas and 2 small pizzas: 2 large pizzas x 16 slices per pizza = 32 slices

2 small pizzas x 8 slices per pizza = 16 slices. 

Next, we add the number of slices in each size pizza to find the total number of slices: 32 slices + 16 slices = 48 slices. 

Therefore, if Albert eats all 2 large pizzas and 2 small pizzas, he will eat 48 slices of pizza that day. 

Answer: 48 slices. \\

\hline
\end{tabular}
\caption{The prompt template used to collect math reasoning dataset for fine-tuning. An example from GSM8K is also included.}
\label{tab:p_math}
\end{table*}

\section{Commonsense Data Templates}
\label{sec:commonsense_templates}

As each dataset in the commonsense reasoning domain entails distinct tasks, we adopt a structured template by initially describing the task’s goal, followed by the corresponding content and answer. Table \ref{tab:p_commonsense} shows the templates used to collect commonsense reasoning data for fine-tuning. 

\begin{table*}[t]\centering
\small
\begin{tabular}{lp{11cm}}\hline
Dataset & Fine-tuning Data Template \\
\hline
BoolQ & Please answer the following question with true or false, question: [QUESTION]

Answer format: true/false

the correct answer is [ANSWER] \\ \hline

PIQA & Please choose the correct solution to the question: [QUESTION]

Solution1: [SOLUTION\_1]

Solution2: [SOLUTION\_2]

Answer format: solution1/solution2 

the correct answer is [ANSWER]\\ \hline
SIQA & Please choose the correct answer to the question: [QUESTION]

Answer1: [ANSWER\_1]

Answer2: [ANSWER\_2]

Answer3: [ANSWER\_3]

Answer format: answer1/answer2/answer3 

the correct answer is [ANSWER]\\ \hline
HellaSwag & Please choose the correct ending to complete the given sentence: [ACTIVITY\_lABEL]: [CONTEXT]

Ending1: [ENDING\_1] 

Ending2: [ENDING\_2] 

Ending3: [ENDING\_3] 

Ending4: [ENDING\_4] 

Answer format: ending1/ending2/ending3/ending4 

the correct answer is [ANSWER]\\ \hline
WinoGrande &  Please choose the correct answer to fill in the blank to complete the given sentence: [SENTENCE]

Option1: [OPTION\_1]

Option2: [OPTION\_2]

the correct answer is [ANSWER]\\ \hline
ARC-e\&ARC-c & Please choose the correct answer to the question: [QUESTION] 

Answer1: [ANSWER\_1]

Answer2: [ANSWER\_2]

Answer3: [ANSWER\_3]

Answer4: [ANSWER\_4]

Answer format: answer1/answer2/answer3/answer4

the correct answer is [ANSWER]
\\    \hline
OBQA &     Please choose the correct answer to the question: [QUESTION] 

Answer1: [ANSWER\_1]

Answer2: [ANSWER\_2]

Answer3: [ANSWER\_3]

Answer4: [ANSWER\_4]

Answer format: answer1/answer2/answer3/answer4

the correct answer is [ANSWER]  \\    \hline

\hline
\end{tabular}
\caption{The data template of each dataset used to create commonsense reasoning data for fine-tuning.}
\label{tab:p_commonsense}
\end{table*}

\section{Placement Analysis}
\label{placement_analysis}

Table \ref{tab:placement_results} shows the performance regarding the placement of adapters in various locations on math reasoning datasets. The fine-tuning dataset utilized for this study is Math10K. Meanwhile, the base models employed is LLaMA-7B. We can observe that for the Series Adapter, the best position is to place it after the MLP layers, achieving an average accuracy of $59.5\%$ on the math reasoning datasets. As for the Parallel Adapter, when we place it within the MLP layers, it achieves the best performance of $61.7\%$. Regarding LoRA, we need to insert it simultaneously into both the Multi-head Attention layers and MLP layers to achieve the best performance of $60\%$.

\begin{table*}[t]\centering
\small
\begin{tabular}{llccccccc}
\hline
Model & Loc & MultiArith  &GSM8K &AddSub &AQuA &SingleEq &SVAMP & Avg \\\hline


\multirow{3}{*}{Series}& Attn & 92.3 & 32.0 & 80.0 & 16.9 & 80.5 & 47.9 & 58.3 \\
& MLP  & 92.8 & 33.3 & 80.0 & 15.0 & 83.5 & 52.3 & 59.5 \\
& Both & 94 & 29.8 & 84.1 & 17.3 & 83.5 & 45.1 & 59.0\\
\hline

\multirow{3}{*}{Parallel}& Attn & 94.5 & 33.5 & 83.0 & 17.3 & 80.5 & 46.9 & 59.3 \\
& MLP  &94.5& 35.3& \textbf{86.6}& \textbf{18.1}& \textbf{86.0}& \textbf{49.6}& \textbf{61.7 }\\
& Both &94.3& 30.2& 84.8& 17.7& 84.3& 47.2& 59.8\\

\hline

\multirow{3}{*}{LoRA}& Attn &94.2& 35.3& 79.7& 16.9& 78.7& 45.9& 58.5\\
& MLP & 95.8& 35.0& 80.0& 15.7& 81.7& 47.0& 59.2 \\
& Both & \textbf{96.2} & \textbf{35.6} & 80.5 & 15.7 & 82.3 & \textbf{49.6} & 60.0\\

\hline
\end{tabular}
\caption{An evaluation of the accuracy regarding the placement of adapters in various locations is conducted on math reasoning datasets. The fine-tuning dataset used for this analysis is Math10K. In this context, "Attn" refers to the multi-head attention layer, while "MLP" denotes the MLP layer. The base model employed for this study is LLaMA-7B.
}
\label{tab:placement_results}

\end{table*}

In order to enhance the breadth of our research findings, we conducted additional experiments involving the placement of adapters on various LLMs such as GPT-J and BLOOMz. These experiments were conducted across different model sizes, specifically 7B and 13B parameters. Furthermore, we extended our investigation to encompass diverse tasks, including Commonsense tasks. This approach enabled us to generalize our observations across a wider spectrum of LLMs, sizes, and tasks, thus providing a more comprehensive understanding of the adapter placement strategies.

\subsection{Various LLMs}
Initially, our evaluation focused on comparing the placement of adapters in the context of GPT-J-6B and BLOOMz-7B models, specifically concerning Arithmetic Reasoning tasks. The subsequent Table \ref{tab:gptj_placement} and Table \ref{tab:bloomz_placement} displays the accuracy attained on Arithmetic Reasoning datasets.

\begin{table*}[t]\centering
\small
\begin{tabular}{llccccccc}
\hline
Model & Loc & MultiArith  &GSM8K &AddSub &AQuA &SingleEq &SVAMP & Avg \\\hline


\multirow{3}{*}{GPT-J$_\text{6B}$-Series}& Attn & 90.3&	16.8&	85.0&	16.9&	78.0&	43.3&	55.1\\
& MLP  & 91.7&	19.5&	85.8&	15.0&	81.7&	43.6&	\textbf{56.2} \\
& Both & 90.7&	13.2&	72.7&	16.0&	72.4&	32.8&	49.6\\
\hline

\multirow{3}{*}{GPT-J$_\text{6B}$-Parallel}& Attn & 83.3&	15.5&	84.3&	15.3&	80.3&	45.4&	54.0 \\
& MLP  &92.2&	18.9&	83.8&	17.9&	80.7&	41.1&	\textbf{55.8}\\
& Both &93.2&	17.2&	86.1&	13.0&	80.1&	40.7&	55.1\\

\hline

\multirow{3}{*}{GPT-J$_\text{6B}$-LoRA}& Attn & 87.2& 17.1&	79.4&	13.0&	74.4&	42.8&	52.3\\
& MLP & 91.7 & 22.8&	81.5&	15.7&	80.7&	47.1&	56.6 \\
& Both & 90.7&	23.0&	84.1&	16.1&	84.1&	46.0&	\textbf{57.3}\\

\hline
\end{tabular}
\caption{An evaluation of the accuracy regarding the placement of adapters in various locations is conducted on math reasoning datasets. The fine-tuning dataset used for this analysis is Math10K. In this context, "Attn" refers to the multi-head attention layer, while "MLP" denotes the MLP layer. The base model employed for this study is GPT-J-6B.
}
\label{tab:gptj_placement}

\end{table*}

\begin{table*}[t]\centering
\small
\begin{tabular}{llccccccc}
\hline
Model & Loc & MultiArith  &GSM8K &AddSub &AQuA &SingleEq &SVAMP & Avg \\\hline


\multirow{3}{*}{BLOOMz$_\text{7B}$-Series}& Attn & 84.3&	14.3&	66.1&	20.1&	63.0&	32.7&	46.8\\
& MLP  & 80.7&	14.3&	72.6&	20.5&	69.3&	38.1&	\textbf{49.3} \\
& Both & 77.8&	14.8&	76.2&	14.2&	67.5&	36.1&	47.8\\
\hline

\multirow{3}{*}{BLOOMz$_\text{7B}$-Parallel}& Attn & 83.7&	16.5&	68.1&	15.0&	64.0&	36.7&	47.3 \\
& MLP  &85.8&	18.5&	77.7&	18.9&	74.8&	36.4&	\textbf{52.0}\\
& Both &88.5&	15.2&	75.7&	16.1&	70.1&	34.0&	49.9\\

\hline

\multirow{3}{*}{BLOOMz$_\text{7B}$-LoRA}& Attn & 80.7&	15.8&	59.7&	15.7&	55.1&	29.7&	42.8\\
& MLP & 86.0&	16.4&	69.6&	17.7&	66.1&	40.3&	49.4 \\
& Both & 82.8&	17.4&	72.4&	21.3&	69.9&	41.0&	\textbf{50.8}\\

\hline
\end{tabular}
\caption{An evaluation of the accuracy regarding the placement of adapters in various locations is conducted on math reasoning datasets. The fine-tuning dataset used for this analysis is Math10K. In this context, "Attn" refers to the multi-head attention layer, while "MLP" denotes the MLP layer. The base model employed for this study is BLOOMz-7B.
}
\label{tab:bloomz_placement}

\end{table*}

The bold figures represent the optimal adapter placements for GPT-J-6B and BLOOMz-7B models. Specifically, for both GPT-J-6B and BLOOMz-7B, the series adapter demonstrated optimal performance when placed after the MLP layers. Similarly, the parallel adapter exhibited its best results when positioned in parallel with the MLP layers. Lastly, the LoRA adapter showed superior performance when located after both the Attention layers and the MLP layers concurrently. This observation aligns seamlessly with the assertions made in this paper.

\subsection{Various Size}
This evaluation focuses on assessing various adapter placements with LLaMA-13B in the context of Arithmetic Reasoning tasks, aiming to compare the outcomes with those obtained from LLaMA-7B. Table \ref{tab:llama13b_placement} presents the accuracy achieved on Arithmetic Reasoning datasets, highlighting the comparative analysis between the two model configurations.

\begin{table*}[t]\centering
\small
\begin{tabular}{llccccccc}
\hline
Model & Loc & MultiArith  &GSM8K &AddSub &AQuA &SingleEq &SVAMP & Avg \\\hline


\multirow{3}{*}{LLaMA$_\text{13B}$-Series}& Attn & 97.7&	33.3&	81.8&	15.7&	86.0&	50.7&	60.9 \\
& MLP  & 93.0&	44.0&	80.5&	22.0&	87.6&	50.8&	\textbf{63.0} \\
& Both & 93.8&	29.7&	81.5&	18.6&	84.8&	48.2&	59.4\\
\hline

\multirow{3}{*}{LLaMA$_\text{13B}$-Parallel}& Attn & 96.8&	34.0&	85.3&	17.0&	88.0&	52.4&	62.3 \\
& MLP  &94.3&	43.3&	83.0&	20.5&	89.6&	55.7&	\textbf{64.4}\\
& Both &95.2&	31.8&	84.1&	15.7&	89.2&	52.8&	61.5\\

\hline

\multirow{3}{*}{LLaMA$_\text{13B}$-LoRA}& Attn &94.0&	37.0&	86.0&	16.5&	87.4&	53.5&	62.4\\
& MLP &96.3&	42.1&	84.1&	18.1&	87.8&	55.5&	64.0\\
& Both &94.8&	47.5&	87.3&	18.5&	89.8&	54.6&	\textbf{65.4}\\

\hline
\end{tabular}
\caption{An evaluation of the accuracy regarding the placement of adapters in various locations is conducted on math reasoning datasets. The fine-tuning dataset used for this analysis is Math10K. In this context, "Attn" refers to the multi-head attention layer, while "MLP" denotes the MLP layer. The base model employed for this study is LLaMA-13B.
}
\label{tab:llama13b_placement}

\end{table*}

The optimal placement of series adapter, parallel adapter, and LoRA is also consistent with LLaMA-7B reported in the paper.

\subsection{Various Tasks}

In order to ascertain the generalizability of the placement strategies identified in the context of Arithmetic Reasoning tasks to other domains such as Commonsense reasoning, we conducted a comprehensive evaluation of various adapter placements with LLaMA-7B. Table \ref{tab:commonsense_placement} presents the accuracy scores achieved on Commonsense Reasoning datasets, providing valuable insights into the effectiveness of the adapter placements in this specific task.

\begin{table*}[t]\centering
\setlength{\tabcolsep}{3pt}
\small
\begin{tabular}{llccccccccc}
\hline
Model & Loc & BoolQ&	PIQA&	SIQA&	HellaSwag&	WinoGrande&	ARC-e&	ARC-c&	OBQA&	Avg \\\hline


\multirow{3}{*}{LLaMA$_\text{7B}$-Series}& Attn & 63.7&	76.3&	75.8&	51.7&	73.6&	69.4&	52.9&	69.8&	66.7 \\
& MLP  & 63.0&	79.2&	76.3&	67.9&	75.7&	74.5&	57.1&	72.4&	\textbf{70.8} \\
& Both & 62.4&	72.4&	72.4&	43.7&	70.6&	62.6&	48.6&	63.6&	62.0\\
\hline

\multirow{3}{*}{LLaMA$_\text{7B}$-Parallel}& Attn & 64.4&	75.8&	76.5&	71.1&	79.2&	72.7&	56.3&	71.0&	70.9 \\
& MLP  &67.9&	76.4&	78.8&	69.8&	78.9&	73.7&	57.3&	75.2&	\textbf{72.3}\\
& Both &64.9&	75.6&	75.2&	66.9&	74.6&	67.8&	53.7&	70.2&	68.6\\

\hline

\multirow{3}{*}{LLaMA$_\text{7B}$-LoRA}& Attn &	68.7&	78.3&	74.9&	68.1&	76.2&	78.1&	63.2&	73.6&	72.6\\
& MLP & 66.5&	81.5&	78.5&	73.3&	72.4&	80.0&	64.8&	72.2&	73.7 \\
& Both & 68.9&	80.7&	77.4&	78.1&	78.8&	77.8&	61.3&	74.8&	\textbf{74.7}\\

\hline
\end{tabular}
\caption{An evaluation of the accuracy regarding the placement of adapters in various locations is conducted on Commonsense reasoning datasets. The fine-tuning dataset used for this analysis is Commonsense170K. In this context, "Attn" refers to the multi-head attention layer, while "MLP" denotes the MLP layer. The base model employed for this study is LLaMA-7B.
}
\label{tab:commonsense_placement}

\end{table*}

Our observations reveal that the optimal placement of series adapters, parallel adapters, and LoRA with LLaMA-7B for Commonsense reasoning aligns with their placement in Arithmetic reasoning tasks. These findings demonstrate a consistent pattern in adapter placement across various models, sizes, and task types, emphasizing the stability and reliability of our results.

\section{Configuration Analysis}
\label{configuration_analysis}

Table \ref{tab:configuration_results} shows the accuracy comparison regarding different settings of variable for PEFT methods on math reasoning datasets. The fine-tuning dataset used for this study is Math10K. It can be noted that when the number of virtual tokens in Prefix-Tuning is set to 10, Prefix-Tuning attains an average accuracy of $42.0\%$ on math reasoning datasets. By configuring the bottleneck dimension to 256, Series and Parallel Adapter demonstrate the highest level of performance. However, when the bottleneck size is increased to 512, the accuracy of both Series and Parallel Adapter decreases. The typical setting for LoRA rank is set to 8, but we have discovered that a larger rank can enhance the performance of LoRA. Remarkably, when the rank is increased to 32, LoRA achieves an accuracy of $61.9\%$.

\begin{table*}[t]\centering
\small
\begin{tabular}{llccccccc}
\hline
Model & Variable & MultiArith  &GSM8K &AddSub &AQuA &SingleEq &SVAMP & Average \\\hline


\multirow{4}{*}{Prefix}& vt=10 & 63.2&	24.4&	57.0&	14.2&	55.3&	38.1&	42.0 \\
& vt=20  & 60.3&	22.9&	46.1&	16.1&	51.8&	30.6&	38.0\\
& vt=30 & 51.2&	16.9&	42.3&	15.4&	41.9&	31.1&	33.1\\
& vt=40 & 54.2&	17.8&	49.6&	\textbf{21.7}&	52.0&	33.2&	38.1 \\

\hline

\multirow{4}{*}{Series}& bn=64 & 93.0 &31.6&	80.5&	14.6&	79.9&	44.1&	57.3 \\
& bn=128  & 93.8&	33.4&	81.5&	18.5&	79.9&	48.1&	59.2 \\
& bn=256 & 92.8&	33.3&	80.0&	15.0&	83.5&	\textbf{52.3}&	59.5 \\
& bn=512 & 95.3&	33.7&	77.2&	14.9&	76.6&	42.0&	56.6 \\

\hline

\multirow{4}{*}{Parallel}& bn=64 & 92.7&	36.8&	77.5&	16.9&	81.9&	48.5&	59.1 \\
& bn=128  &95.8	&36.9&	83.0&	17.7&	83.3&	47.8&	60.8 \\
& bn=256 &94.5&	35.3&	86.6&	18.1&	\textbf{86.0}&	49.6&	61.7 \\
& bn=512 &95.0&	31.5&	81.8&	14.6&	81.5&	43.4&	58.0 \\
\hline

\multirow{4}{*}{LoRA}& r=4 & 95.2&	34.7&	78.5&	16.1&	77.8&	45.4&	57.9 \\
& r=8  &\textbf{96.2}&	35.6&	80.5&	15.7&	82.3&	49.6&	60.0 \\
& r=16 & 95.5&	36.2&	82.8&	13.8&	84.4&	50.9&	60.6\\
& r=32 &95.0&	\textbf{37.5}&	\textbf{83.3}&	18.9&	84.4&	52.1&	\textbf{61.9} \\

\hline
\end{tabular}
\caption{The accuracy comparison regarding different settings of variable for PEFT methods on math reasoning datasets. The fine-tuning dataset used for this analysis is Math10K. In this context, "vt" refers to the number of virtual tokens, "bn" denotes the bottleneck size, while "r" is the LoRA rank. The base model employed for this study is LLaMA-7B.
}
\label{tab:configuration_results}

\end{table*}

\section{Qualitative Examples}
\label{sec:examples}

We will show examples randomly sampled from math reasoning and commonsense reasoning datasets in this section. 

\begin{table*}[t]\centering
\small
\begin{tabular}{lp{11cm}}\hline
Question &  At the schools book fair Sam bought 13 adventure books and 17 mystery books. If 15 of the books were used, how many new books did he buy?  \\
\hline
ChatGPT & To calculate the number of mosaic tiles needed to cover two-thirds of Mr. Boarden's 36 square foot bathroom, follow these steps:

Find two-thirds of 36: Multiply 36 by 2/3.

(36 * 2/3) = 24 square feet

Determine the number of mosaic tiles needed per square foot: Given that Mr. Boarden needs 24 mosaic tiles per square foot, multiply the result from step 1 by 24.

(24 * 24) = 576 mosaic tiles

Therefore, Mr. Boarden would need 576 mosaic tiles to cover two-thirds of his 36 square foot bathroom.
\\ \hline

LLaMA$_\text{13B}$-Prefix & 1. Find the total number of books Sam bought: 

13 + 17 = 30 books 

2. Subtract the number of used books from the total number of books: 

30 - 15 = 15 new books 

Therefore, Sam bought 15 new books at the book fair. 

Steps: 

1. Total number of books = 13 + 17 = 30 

2. Number of used books = 15 

3. Number of new books = Total number of books - Number of used books 

4. Number of new books = 30 - 15 = 15 

5. Sam bought 15 new books at the book fair. 

Answer: Sam bought 15 new books at the book fair. (15)

\\ \hline

LLaMA$_\text{13B}$-Series & 1. Find the total number of books Sam bought: 13 + 17 = 30

2. Subtract the number of used books from the total number of books: 30 - 15 = 15

3. Therefore, Sam bought 15 new books. 

Answer: 15
\\ \hline

LLaMA$_\text{13B}$-Parallel & To find out how many new books Sam bought, we need to subtract the number of used books from the total number of books he bought.

Total books = 13 + 17 = 30

Used books = 15

New books = Total books - Used books

New books = 30 - 15

New books = 15

Therefore, Sam bought 15 new books.
\\\hline
LLaMA$_\text{13B}$-LoRA &   1. Find the total number of books Sam bought: 13 + 17 = 30

2. Subtract the number of used books from the total number of books: 30 - 15 = 15

3. Therefore, Sam bought 15 new books. 

Answer: 15
\\\hline

\hline
\end{tabular}
\caption{An example randomly sampled from MultiArith. The outputs of ChatGPT and LLaMA-13B with different PEFT methods.}
\end{table*}

\begin{table*}[t]\centering
\small
\begin{tabular}{lp{11cm}}\hline
Question &  Mr Boarden is remodeling his bathroom. For every square foot, he needs 24 mosaic tiles. How many mosaic tiles would Mr Boarden need to cover two thirds of his 36 sq ft bathroom? \\
\hline
ChatGPT & To calculate the number of mosaic tiles needed to cover two-thirds of Mr. Boarden's 36 square foot bathroom, follow these steps: 

1. Find two-thirds of 36: Multiply 36 by 2/3.

\textbf{(36 * 2/3) = 24 square feet}

2.Determine the number of mosaic tiles needed per square foot: Given that Mr. Boarden needs 24 mosaic tiles per square foot, multiply the result from step 1 by 24.

\textbf{(24 * 24) = 576 mosaic tiles}

Therefore, Mr. Boarden would need 576 mosaic tiles to cover two-thirds of his 36 square foot bathroom.
\\ \hline

LLaMA$_\text{13B}$-Prefix & 1 square foot = 4 square inches

1 square inch = 16 square centimeters

So, 1 sq ft = 16 sq cm

Therefore, 36 sq ft = 36 x 16 sq cm = 576 sq cm

To cover two-thirds of the bathroom, Mr Boarden needs to cover 576 sq cm x 2/3 = 384 sq cm

Since 1 sq cm = 1 mosaic tile, Mr Boarden needs 384 sq cm / 1 sq cm = 384 mosaic tiles

Therefore, Mr Boarden needs 384 mosaic tiles to cover two-thirds of his 36 sq ft bathroom. \\ \hline

LLaMA$_\text{13B}$-Series & 1. Find the area of the bathroom: 36 sq ft

2. Find two-thirds of the area: \textbf{2/3 x 36 sq ft = 24 sq ft}

3. Multiply the area by the number of tiles needed per square foot: \textbf{24 sq ft x 24 tiles/sq ft = 576 tiles}

4. Therefore, Mr Boarden would need 576 mosaic tiles to cover two-thirds of his 36 sq ft bathroom. \\ \hline

LLaMA$_\text{13B}$-Parallel &  1. Calculate the total area of the bathroom: 36 sq ft

2. Determine how much area Mr Boarden wants to cover with mosaic tiles: \textbf{2/3 x 36 sq ft = 24 sq ft}

3. Calculate the total number of mosaic tiles needed: \textbf{24 sq ft x 24} mosaic tiles per sq ft = 600 mosaic tiles

Therefore, Mr Boarden would need 600 mosaic tiles to cover two thirds of his 36 sq ft bathroom. The answer in Arabic numerals is 600.
\\\hline
LLaMA$_\text{13B}$-LoRA &  1. Calculate the total area of the bathroom: 36 sq ft

2. Determine how much area needs to be covered: \textbf{2/3 x 36 sq ft = 24 sq ft}

3. Calculate the number of mosaic tiles needed: \textbf{24 sq ft x 24 mosaic tiles per sq ft = 576 mosaic tiles}

Therefore, Mr Boarden would need 576 mosaic tiles to cover two thirds of his 36 sq ft bathroom. The answer in Arabic numerals is 576.  \\\hline

\hline
\end{tabular}
\caption{An example randomly sampled from GSM8K. The outputs of ChatGPT and LLaMA-13B with different PEFT methods.}
\end{table*}

\begin{table*}[t]\centering
\small
\begin{tabular}{lp{11cm}}\hline
Question & There are 7 crayons in the drawer. Mary took 3 crayons out of the drawer. How many crayons are there now?   \\
\hline
ChatGPT & To calculate the number of crayons remaining in the drawer after Mary took some out, follow these steps:

Subtract the number of crayons Mary took from the total number of crayons in the drawer: 7 - 3 = 4 crayons.

Therefore, there are now 4 crayons remaining in the drawer.
\\ \hline

LLaMA$_\text{13B}$-Prefix & 1. Start with the initial number of crayons in the drawer: 7.

2. Subtract the number of crayons Mary took out of the drawer: 3.

3. The final number of crayons in the drawer is 7 - 3 = 4.

Therefore, there are 4 crayons in the drawer now.
\\ \hline

LLaMA$_\text{13B}$-Series & 1. Start with the initial number of crayons in the drawer: 7

2. Subtract the number of crayons Mary took out: 3

3. Perform the subtraction: 7 - 3 = 4

4. Therefore, there are now 4 crayons in the drawer.

Answer: There are now 4 crayons in the drawer.
\\ \hline

LLaMA$_\text{13B}$-Parallel & 1. Start with the total number of crayons in the drawer: 7

2. Subtract the number of crayons Mary took out: 3

3. Perform the subtraction: 7 - 3 = 4

4. The final answer is 4 crayons.

Therefore, there are now 4 crayons in the drawer.
\\\hline
LLaMA$_\text{13B}$-LoRA &    1. Start with the initial number of crayons: 7

2. Subtract the number of crayons Mary took out: 7 - 3 = 4

3. The final answer is 4 crayons.

Therefore, there are now 4 crayons in the drawer.
\\\hline

\hline
\end{tabular}
\caption{An example randomly sampled from AddSub. The outputs of ChatGPT and LLaMA-13B with different PEFT methods.}
\end{table*}

\begin{table*}[t]\centering
\small
\begin{tabular}{lp{11cm}}\hline
Question &  Suppose you want to arrange your English, Hindi, Mathematics, History, Geography and Science books on a shelf. In how many ways can you do it ? Answer Choices: (A) 520 (B) 720 (C) 920 (D) None (E) Cannot be determined  \\
\hline
ChatGPT & To calculate the number of ways you can arrange the English, Hindi, Mathematics, History, Geography, and Science books on a shelf, we can use the concept of permutations. Since order matters when arranging the books, we need to find the number of permutations.

To determine the total number of arrangements, follow these steps:

Count the number of books: There are 6 books in total.

Use the formula for permutations of n objects, where n is the number of books: n! (n factorial).

n! = 6! = 6 x 5 x 4 x 3 x 2 x 1 = 720

Therefore, you can arrange the books on the shelf in 720 different ways.

The correct answer choice is (B) 720.
\\ \hline

LLaMA$_\text{13B}$-Prefix & 1. Let's start by assigning variables to the unknowns in the problem. Let a, b, and c be the digits in M, N, and 396c respectively.

2. We know that M and N are three-digit numbers, so a + b + c = 3.

3. We also know that M - N + 396c = 990, so we can substitute a, b, and c into this equation:

a(1000000 - 100000 + 396c) + b(1000000 - 100000 + 396c) + c(1000000 - 100000 + 396c) = 990

4. Simplifying this equation, we get:

a
\\ \hline

LLaMA$_\text{13B}$-Series & To solve this problem, we need to use the multiplication principle of counting, which states that if there are m ways to do one thing and n ways to do another thing, then there are m x n ways to do both things together.

In this case, there are 6 books to arrange (English, Hindi, Mathematics, History, Geography, and Science). We can arrange these books in 6! ways.

Therefore, the total number of ways to arrange the books on the shelf is 6! = 720.

Hence, the answer is option B) 720.
\\ \hline

LLaMA$_\text{13B}$-Parallel & The total number of books to be arranged on the shelf is 5 (English, Hindi, Mathematics, History, Geography, and Science).

To find the number of ways to arrange these books on the shelf, we can use the formula for permutations of n objects taken r at a time, which is given by:

P(n,r) = n! / (n-r)!

where n is the total number of objects, and r is the number of objects to be arranged at a time.

In this case, n = 5 (the total number of books), and r = 5 (since we want to arrange all 5 books on the shelf).

Substituting these values into the formula, we get:

P(5,5) = 5! / (5-5)!
P(5,5) = 5! / (5-5)!
P(5,5) = 5! / (5-5)!
P(5,5) = 5! / (5-5)!
P(5,5) = 5! / (5-5)!
\\\hline
LLaMA$_\text{13B}$-LoRA &  The total number of ways to arrange 6 books on a shelf is 6! = 6 x 5 x 4 x 3 x 2 x 1 = 720.

Therefore, the answer is option B) 720.  \\\hline

\hline
\end{tabular}
\caption{An example randomly sampled from AQuA. The outputs of ChatGPT and LLaMA-13B with different PEFT methods.}
\end{table*}

\begin{table*}[t]\centering
\small
\begin{tabular}{lp{11cm}}\hline
Question &   Oceanside Bike Rental Shop charges 17 dollars plus 7 dollars an hour for renting a bike. Tom paid 80 dollars to rent a bike. How many hours did he pay to have the bike checked out? \\
\hline
ChatGPT & To calculate the number of hours Tom paid to have the bike checked out, follow these steps:

1. Subtract the base charge from the total amount Tom paid: 80 - 17 = 63 dollars.

2. Divide the remaining amount by the hourly rate to find the number of hours: 63 / 7 = 9 hours.

Therefore, Tom paid to have the bike checked out for 9 hours.

Note: It's important to note that the result represents the whole number of hours. If there were any partial hours, they would not be reflected in the calculation.
\\ \hline

LLaMA$_\text{13B}$-Prefix & 1. Let x be the number of hours Tom paid to rent the bike.

2. The cost of renting the bike for x hours is 17 + 7x dollars.

3. We know that Tom paid 80 dollars, so we can set up an equation:

17 + 7x = 80

4. Solving for x, we divide both sides by 7:

x = 10

5. Therefore, Tom paid 10 hours to rent the bike. 

\\ \hline

LLaMA$_\text{13B}$-Series & 1. Find the total cost of renting the bike: 

\$17 + \$7/hour = \$24/hour 

2. Divide the total cost by the cost per hour to find the number of hours: 

\$80 ÷ \$24/hour = 3.33 hours 

3. Round up to the nearest whole number since you can't rent a bike for a fraction of an hour: 

3 hours 

Therefore, Tom paid to have the bike checked out for 3 hours.
\\ \hline

LLaMA$_\text{13B}$-Parallel & 1. Subtract the initial charge of \$17 from the total amount paid: \$80 - \$17 = \$63

2. Divide the remaining amount by the hourly rate: \$63 ÷ \$7 = 9

3. The answer is 9 hours.

Therefore, Tom paid to have the bike checked out for 9 hours.
\\\hline
LLaMA$_\text{13B}$-LoRA &  1. Let x be the number of hours Tom paid to have the bike checked out.

2. The cost of renting the bike for x hours is 7x dollars.

3. The total cost of renting the bike for x hours is 17 + 7x dollars.

4. We know that Tom paid 80 dollars to rent the bike, so we can set up an equation: 17 + 7x = 80

5. Solving for x, we can subtract 17 from both sides: 7x = 63

6. Dividing both sides by 7, we get x = 9

Therefore, Tom paid to have the bike checked out for 9 hours.  \\\hline

\hline
\end{tabular}
\caption{An example randomly sampled from SingleEq. The outputs of ChatGPT and LLaMA-13B with different PEFT methods.}
\end{table*}

\begin{table*}[t]\centering
\small
\begin{tabular}{lp{11cm}}\hline
Question &  Matthew gave equal numbers of crackers and cakes to his 4 friends. If he had 32 crackers and 98 cakes initially. How many crackers did each person eat?  \\
\hline
ChatGPT & To calculate the number of crackers each person ate, we need to find the number of crackers that each friend received initially and subtract it from the total number of crackers Matthew had.

Here are the steps:

Divide the total number of crackers by the number of friends (4) to find the number of crackers each friend received initially: 32 / 4 = 8 crackers per friend.

Since we want to find the number of crackers each person ate, we need to subtract the initial number of crackers per friend from the total number of crackers: 32 - 8 = 24 crackers.

Therefore, each person ate 24 crackers.
\\ \hline

LLaMA$_\text{13B}$-Prefix & 1. Find the total number of crackers and cakes that Matthew initially had: 32 + 98 = 130

2. Divide the total number of crackers and cakes by the number of friends to find the initial amount given to each friend: 130 ÷ 5 = 26

3. Subtract the initial amount given to each friend from the total number of crackers and cakes to find the remaining amount: 130 - 26 = 104

4. Divide the remaining amount by the number of friends to find the amount each person ate: 104 ÷ 5 = 21

Therefore, each person ate 21 crackers.
\\ \hline

LLaMA$_\text{13B}$-Series & 1. Divide the total number of crackers by the number of friends: 32 ÷ 4 = 8

2. Divide the total number of cakes by the number of friends: 98 ÷ 4 = 24

3. Add the number of crackers and cakes each friend received: 8 + 24 = 32

4. Therefore, each person ate 32 crackers and 24 cakes.

Answer: Each person ate 32 crackers.
\\ \hline

LLaMA$_\text{13B}$-Parallel & 1. Divide the total number of crackers by the number of friends: 32 ÷ 4 = 8

2. Divide the total number of cakes by the number of friends: 98 ÷ 4 = 24

3. Add the number of crackers and cakes each person ate: 8 + 24 = 32

Therefore, each person ate 32 crackers and cakes combined.
\\\hline
LLaMA$_\text{13B}$-LoRA & 1. Divide the total number of crackers by the number of friends: 32 ÷ 4 = 8

2. Divide the total number of cakes by the number of friends: 98 ÷ 4 = 24

3. Add the number of crackers and cakes each person received: 8 + 24 = 32

Therefore, each person ate 32 crackers and cakes in total.   \\\hline

\hline
\end{tabular}
\caption{An example randomly sampled from SVAMP. The outputs of ChatGPT and LLaMA-13B with different PEFT methods.}
\end{table*}

\begin{table*}[t]\centering
\small
\begin{tabular}{lp{11cm}}\hline
Question &  Does ethanol take more energy make that produces?  \\
\hline
ChatGPT & False.
\\ \hline

LLaMA$_\text{13B}$-Prefix & the correct answer is true
\\ \hline

LLaMA$_\text{13B}$-Series & the correct answer is true
\\ \hline

LLaMA$_\text{13B}$-Parallel & false
\\\hline
LLaMA$_\text{13B}$-LoRA &  the correct answer is false  \\\hline

\hline
\end{tabular}
\caption{An example randomly sampled from BoolQ. The outputs of ChatGPT and LLaMA-13B with different PEFT methods.}
\end{table*}

\begin{table*}[t]\centering
\small
\begin{tabular}{lp{11cm}}\hline
Question & How do I ready a guinea pig cage for it's new occupants?

Solution1: Provide the guinea pig with a cage full of a few inches of bedding made of ripped paper strips, you will also need to supply it with a water bottle and a food dish.

Solution2: Provide the guinea pig with a cage full of a few inches of bedding made of ripped jeans material, you will also need to supply it with a water bottle and a food dish.\\
\hline
ChatGPT & Solution1
\\ \hline

LLaMA$_\text{13B}$-Prefix & the correct answer is solution1
\\ \hline

LLaMA$_\text{13B}$-Series & the correct answer is solution2
\\ \hline

LLaMA$_\text{13B}$-Parallel & the correct answer is solution2
\\\hline
LLaMA$_\text{13B}$-LoRA & the correct answer is solution1   \\\hline

\hline
\end{tabular}
\caption{An example randomly sampled from PIQA. The outputs of ChatGPT and LLaMA-13B with different PEFT methods.}
\end{table*}

\begin{table*}[t]\centering
\small
\begin{tabular}{lp{11cm}}\hline
Question & Tracy didn't go home that evening and resisted Riley's attacks. What does Tracy need to do before this?

Answer1: make a new plan Answer2: Go home and see Riley Answer3: Find somewhere to go\\
\hline
ChatGPT & Answer1/make a new plan
\\ \hline

LLaMA$_\text{13B}$-Prefix & the correct answer is answer1
\\ \hline

LLaMA$_\text{13B}$-Series & the correct answer is answer3
\\ \hline

LLaMA$_\text{13B}$-Parallel & the correct answer is answer3
\\\hline
LLaMA$_\text{13B}$-LoRA &  the correct answer is answer3  \\\hline

\hline
\end{tabular}
\caption{An example randomly sampled from SIQA. The outputs of ChatGPT and LLaMA-13B with different PEFT methods.}
\end{table*}

\begin{table*}[t]\centering
\small
\begin{tabular}{lp{11cm}}\hline
Question & Roof shingle removal: A man is sitting on a roof. he

Ending1: is using wrap to wrap a pair of skis. Ending2: is ripping level tiles off. Ending3: is holding a rubik's cube. Ending4: starts pulling up roofing on a roof.   \\
\hline
ChatGPT & Ending4: starts pulling up roofing on a roof.
\\ \hline

LLaMA$_\text{13B}$-Prefix & the correct answer is ending4
\\ \hline

LLaMA$_\text{13B}$-Series & the correct answer is ending4
\\ \hline

LLaMA$_\text{13B}$-Parallel & the correct answer is ending4
\\\hline
LLaMA$_\text{13B}$-LoRA &  the correct answer is ending4  \\\hline

\hline
\end{tabular}
\caption{An example randomly sampled from . The outputs of ChatGPT and LLaMA-13B with different PEFT methods.}
\end{table*}

\begin{table*}[t]\centering
\small
\begin{tabular}{lp{11cm}}\hline
Question &  Sarah was a much better surgeon than Maria so \_ always got the easier cases.

Option1: Sarah Option2: Maria
\\ \hline
ChatGPT & option1
\\ \hline

LLaMA$_\text{13B}$-Prefix & the correct answer is option1
\\ \hline

LLaMA$_\text{13B}$-Series & the correct answer is option1
\\ \hline

LLaMA$_\text{13B}$-Parallel & the correct answer is option1
\\\hline
LLaMA$_\text{13B}$-LoRA &  the correct answer is option1  \\\hline

\hline
\end{tabular}
\caption{An example randomly sampled from WinoGrande. The outputs of ChatGPT and LLaMA-13B with different PEFT methods.}
\end{table*}

\begin{table*}[t]\centering
\small
\begin{tabular}{lp{11cm}}\hline
Question &  Which statement best explains why photosynthesis is the foundation of most food webs?

Answer1: Sunlight is the source of energy for nearly all ecosystems. Answer2: Most ecosystems are found on land instead of in water. Answer3: Carbon dioxide is more available than other gases. Answer4: The producers in all ecosystems are plants. \\
\hline
ChatGPT & 
\\ \hline

LLaMA$_\text{13B}$-Prefix & Answer1. Sunlight is the source of energy for nearly all ecosystems.
\\ \hline

LLaMA$_\text{13B}$-Series & the correct answer is answer1
\\ \hline

LLaMA$_\text{13B}$-Parallel & the correct answer is answer4
\\\hline
LLaMA$_\text{13B}$-LoRA &  the correct answer is answer1  \\\hline

\hline
\end{tabular}
\caption{An example randomly sampled from ARC-e. The outputs of ChatGPT and LLaMA-13B with different PEFT methods.}
\end{table*}

\begin{table*}[t]\centering
\small
\begin{tabular}{lp{11cm}}\hline
Question & An astronomer observes that a planet rotates faster after a meteorite impact. Which is the most likely effect of this increase in rotation?

Answer1: Planetary density will decrease. Answer2: Planetary years will become longer. Answer3: Planetary days will become shorter. Answer4: Planetary gravity will become stronger.   \\
\hline
ChatGPT & Answer3
\\ \hline

LLaMA$_\text{13B}$-Prefix & the correct answer is answer3
\\ \hline

LLaMA$_\text{13B}$-Series & the correct answer is answer2
\\ \hline

LLaMA$_\text{13B}$-Parallel & the correct answer is answer3
\\\hline
LLaMA$_\text{13B}$-LoRA &  the correct answer is answer3  \\\hline

\hline
\end{tabular}
\caption{An example randomly sampled from ARC-c. The outputs of ChatGPT and LLaMA-13B with different PEFT methods.}
\end{table*}

\begin{table*}[t]\centering
\small
\begin{tabular}{lp{11cm}}\hline
Question & A person wants to start saving money so that they can afford a nice vacation at the end of the year. After looking over their budget and expenses, they decide the best way to save money is to

Answer1: make more phone calls Answer2: quit eating lunch out Answer3: buy less with monopoly money Answer4: have lunch with friends   \\
\hline
ChatGPT & Answer2: quit eating lunch out.
\\ \hline

LLaMA$_\text{13B}$-Prefix & the correct answer is answer2
\\ \hline

LLaMA$_\text{13B}$-Series & the correct answer is answer2
\\ \hline

LLaMA$_\text{13B}$-Parallel & the correct answer is answer2
\\\hline
LLaMA$_\text{13B}$-LoRA &  the correct answer is answer2  \\\hline

\hline
\end{tabular}
\caption{An example randomly sampled from OBQA. The outputs of ChatGPT and LLaMA-13B with different PEFT methods.}
\end{table*}
\chapter{Appendix of InstructAV}
\label{appendixb} 

\section{Case Study for \textsf{InstructAV}}
\label{case_study}
We have included randomly selected examples to showcase the capabilities of \textsf{InstructAV} in generating both classification predictions and language feature-based explanations. One such example from the IMDB dataset is detailed in Table~\ref{tab:instructav_example}. This example demonstrates how \textsf{InstructAV} first provides a classification prediction and then follows it with detailed explanations anchored in specific language features pertinent to the classification.

These examples highlight \textsf{InstructAV}'s proficiency in delivering accurate classifications paired with clear and credible explanations. Notably, the explanations are tailored to each specific sample, rather than being generic or broad. This customization enhances the explanations' effectiveness, as they are directly correlated with the content of the original text. For instance, the model explicates features such as writing style and use of expressions by incorporating words and phrases directly from the text in question. This approach not only bolsters the relevance of the explanations but also their trustworthiness, providing users with insightful and contextually grounded rationale behind each classification decision made by \textsf{InstructAV}.

\begin{table*}[htb]
    \centering
    \small
    \begin{tabular}{lp{11cm}}
        \toprule
        \begin{minipage}{0.95\textwidth}
        \textbf{Instruction}: \textit{Please decide if the following Text1 and Text2 are written by the same authors. 'yes' means from the same author, 'no' means not from the same author. Then, provide an analysis based on writing styles. }
        \\\\
        \textbf{Text 1:} 
        \textit{ The British Soap Awards 2005 I am a fan of only two soaps , Coronation Street and EastEnders , but I am always going to watch this programme to see what the other soaps are like , if they deserve to win something , and what I may be missing out on . I always want to agree with the awards 
        ...} [Truncated due to length restriction]
        \\\\
        \textbf{Text 2:}
        \textit{Father of the Bride As far as I can work out , it is after this film that the career of Steve Martin has had very few laughs , which is a shame , he just needs to find the right film . Anyway , this is the remake of the 1950 film , and it sees middle aged father George Banks ( Martin ) with wife Nina ( Diane Keaton ) having to prepare the wedding for their 21-year-old daughter Annie Banks ( Kimberly Williams-Paisley )
        ...} [Truncated due to length restriction] \\
        \hrule
        \end{minipage} \\

        \begin{minipage}{0.95\textwidth}
        \hrule
            \textbf{ Generated Output} 
        \hrule
\textbf{Classification:} The correct answer is yes. \\
\textbf{Explanation:} Upon analyzing Text 1 and Text 2 based on the listed writing style characteristics, we find the following similarities that could suggest that both texts were written by the same author:

\textbf{Writing Style:} Both texts provide personal opinions and evaluations of the subject matter. Text 1 focuses on the British Soap Awards, while Text 2 provides a review of the film \textbackslash``Father of the Bride.\textbackslash``"

\textbf{Expressions and Idioms:} Both texts use colloquial expressions and idioms, such as \textbackslash``if they deserve to win something\textbackslash`` in Text 1 and \textbackslash``he just needs to find the right film\textbackslash`` in Text 2.

\textbf{Tone and Mood:} Both texts have an informal and conversational tone. The author expresses their personal preferences and opinions in both texts.

\textbf{Sentence Structure and Syntax:} Both texts use a mix of compound and simple sentences. For example, \textbackslash``I always want to agree with the awards that are given according to the clips/photos shown\textbackslash`` in Text 1 and \textbackslash``Martin does do well, Keaton offers a good support, and Short adds giggles as the foreign wedding organiser\textbackslash`` in Text 2.

\textbf{Punctuation Style:} Both texts use consistent punctuation, including commas and full stops.

\textbf{Special Characters Style, Capitalization Style:} Both texts capitalize proper nouns and show titles. However, Text 2 uses unconventional capitalization for emphasis, such as \textbackslash``Very good!\textbackslash``

\textbf{Compound and Separate Spelling:} No significant differences in spelling patterns are observed between the two texts.

\textbf{Acronyms and Abbreviations:} Both texts use acronyms and abbreviations, such as \textbackslash``EE\textbackslash`` for \textbackslash``EastEnders\textbackslash`` in Text 1 and \textbackslash``B.D.Wong\textbackslash`` for \textbackslash``B.D. Wong\textbackslash`` in Text 2.

\textbf{Characters Style:} Both texts mention specific actors and their roles, indicating a familiarity with the subject matter.

\textbf{Diatopic Variations and Foreign Languages:} Both texts appear to be written in standard English without diatopic variations or foreign language references.

\textbf{Other Relevant Aspects:} Both texts provide specific details about the subject matter, indicating a personal familiarity or interest in the topic. The author's opinions and evaluations are evident in both texts. 
        \hrule
        \end{minipage} \\
    \end{tabular}
    \caption{A text pair example with the generated responses from \textsf{InstructAV}}
    \label{tab:instructav_example}
\end{table*}

\section{Prompt Templates}
\label{app:prompt}
\begin{table*}[htb]
    \centering
    \small
    \begin{tabular}{lp{11cm}}
        \toprule
        \begin{minipage}{\textwidth}
        \textbf{Task}: Text1 and Text2 are written by the same author. Please analyze their writing styles and explain why they are written by the same author. You can refer to the following characteristics of writing style.    1. writing style.   2. expressions and Idioms.   3. tone and mood.   4. sentence structure and syntax.   5. punctuation style.   6. special characters style, capitalization style.   7. compound and separate spelling.   8. acronyms and abbreviations.   9. characters style.   10. Diatopic variations and foreign languages.   11. any other relevant aspect. \\\\
        \textbf{Text 1:} 
        \textit{The British Soap Awards 2005 I am a fan of only two soaps , Coronation Street and EastEnders , but I am always going to watch this programme to see what the other soaps are like , if they deserve to win something , and what I may be missing out on.  
       ...} [Truncated due to length restriction]
        \\\\
        \textbf{Text 2:}
        \textit{Father of the Bride As far as I can work out , it is after this film that the career of Steve Martin has had very few laughs , which is a shame , he just needs to find the right film. Anyway , this is the remake of the 1950 film , and ....} [Truncated due to length restriction] \\
        Please follow the format of the analysis method in the demostrations.
        You will be given 2 demostrations.\\
        \#\#\# \textbf{Demostration Start:} \\
  Text 1: Mama's homemade remedy for laughs! Mama Harper develops her mother's concoction to help her family get over an illness like the flu. After her son and bimbo-in-law Naomi who is now expecting their child realized that they can't afford it......[Truncated due to length restriction] \\
  Text 2:Not so Bad! Remember that this show was one of the first sitcoms in syndication in the 1980s. It was a new thing. When a sitcom or drama could not get a network , they went with syndication. I thought this show was alright in syndication because it reached ......[Truncated due to length restriction] \\\\
  Upon analyzing Text 1 and Text 2 based on the listed writing style characteristics, we find the following similarities that could suggest that both texts were written by the same author:\\
  Writing Style: Both texts seem to be reminiscing about older television shows. The author uses a mixture of narrative and opinion in both texts, indicating a personal connection or memory attached to the subject matter.\\
  Expressions and Idioms: The expressions "I miss this show" and "I don't recall ever not watching it" showcase a personal touch and sentimentality in both texts.\\
  Tone and Mood: Both texts have a nostalgic and somewhat informal tone. The author shares memories and impressions about shows from the past, using phrases like "I was a kid so I didn't know better" and "Still, I miss this show."\\
  Sentence Structure and Syntax: Both texts employ a mix of compound and simple sentences. For instance, "When a sitcom or drama could not get a network, they went with syndication" and "Mama Harper develops her mother's concoction to help her family get over an illness like the flu."\\
  Punctuation Style: Both texts employ spaces before and after commas inconsistently, such as in "\$180 , 000" and "Marla Pennington , Edie McClurg."\\
  Special Characters Style, Capitalization Style: Both texts capitalize show titles like "Mama's Family" and show proper nouns. However, both texts sometimes lack proper capitalization at the start of sentences, such as "Still" in Text 1.\\
  Compound and Separate Spelling: No significant differences in spelling patterns are observed between the two texts. 
  Acronyms and Abbreviations: The second text uses "1980s" to denote the decade, showcasing a preference for this kind of abbreviation.\\
  Characters Style: The character portrayal in both texts has a casual undertone. For example, "bimbo-in-law Naomi" and "the guy who played her father."\\
  Diatopic Variations and Foreign Languages: Both texts appear to be written in American English and reference American television shows.\\
  Other Relevant Aspects:Both texts contain anecdotal or subjective experiences like "I was a kid so I didn't know better" and "Still, I miss this show." The author provides specific details about show characters, indicating a personal familiarity with the subject. There's a recurring theme of syndication and older television programming in both texts, further suggesting the same author might have written them.\\
  In conclusion, based on the aforementioned similarities in writing style, expressions, tone, and other characteristics, it is plausible that Text 1 and Text 2 were written by the same author.\\
         \hrule
        \end{minipage} \\
    \end{tabular}
    \caption{Prompt used for generating explanation labels}
    \label{tab:prompt_example}
\end{table*}

\section{Collected Datasets}
\label{app:datasets}
\begin{table*}[htb]
    \centering
    \small
    \begin{tabular}{lp{11cm}}
        \toprule
        \begin{minipage}{\textwidth}
        \textbf{Classification Datasets} \\
        \textbf{Instruction}: Please decide if the following Text 1 and Text 2 are written by the same authors. 'yes' means from the same author, 'no' means not from the same author. \\\\
        \textbf{Text 1}: \textit{Send in the Clowns!!! I don't know why we haven't heard of Judy Cohen since this episode. She can break your heart singing Stephen Sondheim's Send in the Clowns. She plays herself as an aspiring singer who beats herself down very much because she is overweight and not the classic beauty......}[Truncated due to length restriction] \\ 
        \textbf{Text 2}: \textit{Could have been better! Columbo's Most Dangerous Match is about chess. An unlikely subject or motive for murder in my opinion. The star is Manchurian Candidate's Lawrence Harvey who plays a deaf chess champion......}[Truncated due to length restriction]\\\\
        \textbf{Output}: The correct answer is yes. \\
        \hrule
        \textbf{Classification \& Explanation Datasets} \\
        \hrule
        \textbf{Instruction}: \textit{Please decide if the following Text 1 and Text 2 are written by the same authors. 'yes' means from the same author, 'no' means not from the same author. Then, provide an analysis based on writing styles.}    \\\\
        \textbf{Text 1}: \textit{Send in the Clowns!!! I don't know why we haven't heard of Judy Cohen since this episode. She can break your heart singing Stephen Sondheim's Send in the Clowns. She plays herself as an aspiring singer who beats herself down very much because she is overweight and not the classic beauty......}[Truncated due to length restriction] \\ 
        \textbf{Text 2}: \textit{Could have been better! Columbo's Most Dangerous Match is about chess. An unlikely subject or motive for murder in my opinion. The star is Manchurian Candidate's Lawrence Harvey who plays a deaf chess champion......}[Truncated due to length restriction]\\\\
        \textbf{Output}: The correct answer is yes. Analyzing Text 1 and Text 2 based on the listed writing style characteristics, we find the following similarities that could suggest that both texts were written by the same author:\\
            Writing Style: Both texts provide detailed commentary and analysis of specific episodes of a TV show. The author shares personal opinions and emotional responses to the episodes.\\
            Expressions and Idioms: Both texts use phrases like ``break your heart'' and ``a hell of a game'' to evoke emotional reactions and convey personal opinions.\\
            Tone and Mood: Both texts exhibit enthusiasm and engagement with the subject matter. The author expresses admiration for certain performances and emphasizes their impact.\\
            Sentence Structure and Syntax: Both texts use a mix of simple and compound sentences. The sentences focus on describing specific events and interactions between characters.\\
            Punctuation Style: Both texts use consistent punctuation, including proper use of commas and ellipses. They enhance the flow and expressiveness of the narrative.\\
            Special Characters Style, Capitalization Style: Both texts consistently capitalize proper nouns and show titles. The author pays attention to the details and presents them accurately.\\
            Compound and Separate Spelling: Both texts use standard spelling patterns without significant variations.\\
            Acronyms and Abbreviations: Neither text employs acronyms or abbreviations.\\
            Characters Style: Both texts describe specific characters and their roles within the TV show episodes. The author demonstrates familiarity with the characters' interactions and motivations.\\
            Diatopic Variations and Foreign Languages: Both texts are written in English without reference to specific geographical locations or foreign languages.\\
            Other Relevant Aspects: Both texts contain personal appeals to specific individuals, such as ``Where are you Judy Cohen?'' and ``You almost feel sorry for him.'' The author shows a deep connection to the subject matter and a desire for continued engagement.\\
            In conclusion, based on the aforementioned similarities in writing style, expressions, tone, and other characteristics, it is likely that Text 1 and Text 2 were written by the same author.\\
         \hrule
        \end{minipage} \\
    \end{tabular}
    \caption{Example of collected \textbf{Classification} Datasets and \textbf{Classification \& Explanation} Datasets}
    \label{tab:dataset_example}
\end{table*}


\printbibliography[heading=bibintoc]


\end{document}